\documentclass[sigconf]{acmart}
\AtBeginDocument{%
  \providecommand\BibTeX{{%
    \normalfont B\kern-0.5em{\scshape i\kern-0.25em b}\kern-0.8em\TeX}}}

\usepackage{bm}
\usepackage{enumitem}
\usepackage{graphicx}
\usepackage{amsthm}
\usepackage{hyperref}
\usepackage{balance}
\usepackage{multirow}
\usepackage{appendix}
\usepackage{tabularx}
\usepackage{tcolorbox}

\newcommand{\header}[1]{\noindent \textbf{#1}}
\newcommand{\modelname}{EHRBench}
\newcommand{\best}[1]{\underline{#1}$^{\textbf{\#1}}$}
\newcommand{\second}[1]{\underline{#1}$^{\textbf{\#2}}$}
\newcommand{\third}[1]{\underline{#1}$^{\textbf{\#3}}$}
\newcommand{\fourth}[1]{\underline{#1}$^{\textbf{\#4}}$}
\newcommand{\fifth}[1]{\underline{#1}$^{\textbf{\#5}}$}
\newcommand{\sixth}[1]{\underline{#1}$^{\textbf{\#6}}$}
\newcommand{\seventh}[1]{\underline{#1}$^{\textbf{\#7}}$}
\newcommand{\eighth}[1]{\underline{#1}$^{\textbf{\#8}}$}

\copyrightyear{2026}
\acmYear{2026}
\setcopyright{cc}
\setcctype{by-nc-nd}
\acmConference[KDD 2026] {Proceedings of the 32nd ACM SIGKDD Conference on Knowledge Discovery and Data Mining V.2}{August 9--13, 2026}{Jeju Island, Republic of Korea.}
\acmBooktitle{Proceedings of the 32nd ACM SIGKDD Conference on Knowledge Discovery and Data Mining V.2 (KDD 2026), August 9--13, 2026, Jeju Island, Republic of Korea}
\acmISBN{979-8-4007-2259-2/2026/08}
\acmDOI{10.1145/3770855.3817571}

\author{Yuzhang Xie}
\affiliation{\institution{Emory University}\city{Atlanta}\state{GA}\country{USA}}
\email{yuzhang.xie@emory.edu}

\author{Keqi Han}
\affiliation{\institution{Emory University}\city{Atlanta}\state{GA}\country{USA}}
\email{keqi.han@emory.edu}

\author{Yunpeng Xiao}
\affiliation{\institution{Emory University}\city{Atlanta}\state{GA}\country{USA}}
\email{yunpeng.xiao@emory.edu}

\author{Hejie Cui}
\affiliation{\institution{Stanford University}\city{Stanford}\state{CA}\country{USA}}
\email{hejie.cui@stanford.edu}

\author{Guanchen Wu}
\affiliation{\institution{Emory University}\city{Atlanta}\state{GA}\country{USA}}
\email{guanchen.wu@emory.edu}

\author{Ziyang Zhang}
\affiliation{\institution{Emory University}\city{Atlanta}\state{GA}\country{USA}}
\email{ziyang.zhang2@emory.edu}

\author{Kai Shu}
\affiliation{\institution{Emory University}\city{Atlanta}\state{GA}\country{USA}}
\email{kai.shu@emory.edu}

\author{Jiaying Lu}
\affiliation{\institution{Emory University}\city{Atlanta}\state{GA}\country{USA}}
\email{jiaying.lu@emory.edu}

\author{Xiao Hu}
\affiliation{\institution{Emory University}\city{Atlanta}\state{GA}\country{USA}}
\email{xiao.hu@emory.edu}

\author{Carl Yang}
\authornote{Carl Yang is the corresponding author.}
\affiliation{\institution{Emory University}\city{Atlanta}\state{GA}\country{USA}}
\email{j.carlyang@emory.edu}

\renewcommand{\shortauthors}{Yuzhang Xie et al.}

\begin{document}
\begin{abstract}
Clinical decision-making (CDM) is central to real-world clinical workflows, where clinicians infer diagnoses, select treatments, or anticipate future health outcomes under incomplete evidence. LLMs are increasingly used to support these decisions due to strong language capabilities, broad biomedical knowledge, and efficiency, yet the reliability of LLMs on real-world clinical decision tasks remains insufficiently understood. To evaluate CDM models, especially LLM-based models, an ideal and practical medical decision benchmark should be constructed via an automated yet reliable pipeline to ensure both scale and quality. Moreover, the grounding of a CDM benchmark in real patient EHRs can better support evaluation on practical CDM tasks that require substantive biomedical knowledge and clinical inference. To fill the gaps, we introduce \modelname{}, an automated and reliable EHR-grounded benchmark for evaluating LLM-based clinical decision-making at scale. To ensure scalability and reliability, \modelname{} is constructed through an EHR--LLM--knowledge-base (KB) interaction pipeline. For efficiency, we use a specialized LLM to automatically convert encounter-level EHR trajectories into structured templates and deterministically instantiate the templates into QA items. In parallel, we apply systematic KB-based verification and enrichment to filter hallucinated or ambiguous relations and to improve reliability. Using this pipeline, we construct nearly 1M (960{,}067) QA items spanning three core inference-required clinical decision tasks: diagnosis, treatment, and prognosis. We benchmark more than 30 representative LLMs on \modelname{} and provide detailed analyses of performance and robustness. The results show consistent capability trends across settings, further validating the reliability of \modelname{} and highlighting actionable gaps toward clinically reliable LLM systems \footnote{The guideline for source code and data of \modelname{} is available at the GitHub link \href{https://github.com/constantjxyz/EHRBench}{https://github.com/constantjxyz/EHRBench}}.
\end{abstract}



\title{\modelname{}: An Automated and Reliable EHR-based Benchmark for Clinical Decision Making with LLMs}
\maketitle

\section{Introduction}
\label{sec:introduction}

Clinical decision-making (CDM) is a fundamental part of real-world clinical workflows, where clinicians must infer diagnoses, determine treatments, or forecast future clinical states from incomplete evidence \cite{subbiah2023next,harish2021artificial,masic2022medical,pelaccia2017novel}. 
For instance, given the observed diagnoses of an encounter, an \emph{in-encounter diagnosis completion} decision requires inferring concurrent conditions, such as identifying chronic kidney disease when type 2 diabetes and diabetic nephropathy are present. Similarly, an \emph{in-encounter treatment selection} decision involves selecting appropriate treatments, such as identifying the necessary anticoagulation for a patient with atrial fibrillation. Furthermore, a \emph{next-encounter prognosis prediction} decision requires anticipating potential downstream outcomes or diagnoses in subsequent encounters, such as forecasting ischemic stroke risk in patients with hypertension and hyperlipidemia. These decisions directly affect patient care and outcomes, carrying substantial clinical significance for patient safety and well-being \cite{panagioti2019prevalence,vasey2021association}.

Large language models (LLMs) are increasingly deployed to support these clinical decisions \cite{zhou2025ehr,molinet2024explanatory,jia2025medikal,oniani2024enhancing,xie2025kerap,hunik2025diagnostic}, owing to their robust language understanding capabilities, broad biomedical knowledge acquired during pre-training, and superior efficiency relative to traditional manual workflows \cite{singhal2023LLM,kumar2023AIforDiagnosis,bhasuran2025preliminary,lu2025crossagentie,wang2025biomedjimpact}.
This rapid progress raises a central question: how reliably do LLMs perform on core clinical decision tasks when the evidence reflects patient-specific, real-world clinical data? Benchmarks are essential for addressing this question, as they enable controlled, reproducible comparisons and provide guidance for the development of safer CDM systems.

Building these benchmarks requires an automated and reliable construction pipeline.
Historically, many medical QA resources have achieved high quality through substantial domain expertise and meticulous manual curation \cite{malaviya2024expertqa,singhal2023LLM,zhou2025automating,wornow2023ehrshot}. However, the high cost associated with manual effort typically limits these benchmarks to a small number of patient records, which restricts the scale and diversity of evaluation \cite{yan2024large,bosma2025dragon}. Since CDM is inherently complex and multi-faceted, large-scale benchmarks are essential for comprehensive evaluation, which in turn necessitates the transition toward automated construction pipelines.

Recent studies have explored the use of LLMs themselves to scale up benchmark creation by generating questions under specific constraints \cite{long2024llms,artsi2024large,sileo2024generating}. While this produces a large volume of data, it raises quality issues since LLMs can hallucinate. Consequently, ensuring that LLM-generated benchmarks are clinically realistic and unambiguous requires more than formatting constraints alone; it necessitates systematic validation (e.g., via external knowledge bases) to mitigate hallucinated clinical relations and ambiguous answers \cite{huang2025survey,niu2024ragtruth}.

Beyond constructing an automated and reliable pipeline, the data source of the benchmark is also important. Grounding a CDM benchmark in patients' real electronic health records (EHRs) facilitates more authentic evaluations of practical CDM tasks.
Currently, most existing medical benchmarks are derived from well-formed narrative sources such as exams, textbooks, clinical guidelines, and clinical notes \cite{jin2021medqa,pal2022medmcqa,liu2024clinicbench,kweon2024ehrnoteqa,mehandru2025erreason,liu2024medchain,kim2024medexqa,dada2025medisumqa,zhang2025llmeval,zuo2025medxpertqa,wang2025trialpanorama}. These sources often make clinical reasoning explicit—for example, by directly stating the rationale for diagnoses or treatments—thereby reducing the need for inference. In contrast, clinicians routinely reason over longitudinal EHRs, where the underlying clinical logic is not pre-digested but must be inferred from patterns of structured events. Unlike general medical sources that emphasize idealized and broadly applicable knowledge, EHRs capture personalized, longitudinal, real-world clinical events and care patterns at scale \cite{knevel2023real,xie2026enhanced,zhang2025type}. Furthermore, compared with free-text clinical notes, which are costly to curate and typically focus on a limited set of salient details, the structured tabular components of EHRs have far higher volume, cover a broader range of clinical concepts, and reflect substantially greater variability in real-world practice  \cite{kim2023ehrchallenges}.

Despite this potential, directly leveraging raw structured EHRs for benchmark construction remains challenging. Clinical relationships in EHRs are largely implicit and must be inferred from temporally ordered events, while fragmentation across coding systems complicates faithful transformation into natural-language prompts without introducing artifacts or label leakage \cite{wu2025utilizing,xie2024promptlink}. In addition, EHR trajectories are often extremely long, making it difficult to convert raw records into LLM-feasible inputs while preserving data fidelity \cite{zhang2024tacco,shao2026llm}. As a result, existing EHR-based benchmarks often emphasize reading-comprehension or information-retrieval tasks (e.g., ``what treatment did the patient receive during this visit'' \cite{lee2022ehrsql,xu2025medagentgym}), rather than core CDM tasks that require substantive biomedical knowledge and clinical inference, such as deciding what should be prescribed given a diagnosis.

\begin{figure}[htbp]
\centering
\includegraphics[width=0.9\columnwidth]{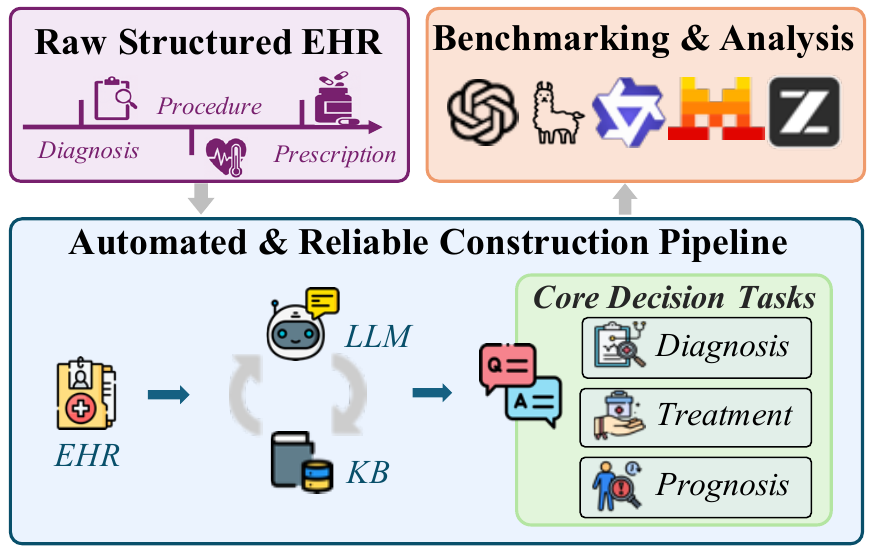}
\caption{Overview of \modelname{}. \modelname{} automatically and reliably transforms raw structured EHR trajectories into QA benchmarks via an EHR-LLM-KB interaction pipeline and evaluates representative LLMs on three core clinical decision tasks: diagnosis, treatment, and prognosis.}
\label{fig:intro}
\end{figure}

To bridge these gaps, we introduce \modelname{}, an automated and reliable benchmark grounded in real-world electronic health records (EHRs) for evaluating the clinical decision-making capabilities of LLMs. 
As illustrated in Figure~\ref{fig:intro}, our framework systematically transforms raw structured EHR trajectories into a benchmark that is both large-scale and high-quality through a multi-stage pipeline that integrates EHR data, LLMs, and external biomedical knowledge bases (KBs).
Specifically, we use LLMs to generate question templates (including clinical relations, questions, and answers) from EHR trajectories, which are concurrently validated (for clinical relations) and enriched (with entity definitions and retrieved evidence) using external KBs to ensure clinical reliability. These generated templates are deterministically instantiated into multiple types of QA items to ensure diversity. 
Using \modelname{}, we evaluate representative LLMs on three core CDM tasks that require substantive biomedical knowledge and clinical inference, covering in-encounter diagnosis completion (\textit{diagnosis}), in-encounter treatment selection (\textit{treatment}), and next-encounter outcome prediction (\textit{prognosis}). We further analyze model performance in terms of accuracy, efficiency, and robustness.

Our contributions are summarized as follows:
\begin{itemize}[itemsep=1pt, topsep=0pt, parsep=0pt, partopsep=0pt]
    \item We construct \modelname{}, a large-scale, EHR-grounded QA benchmark for evaluating LLMs’ clinical decision-making capabilities, comprising nearly 1 million QA items (960,067). To the best of our knowledge, \modelname{} is the first benchmark built directly from raw structured EHR trajectories that leverages LLMs for question template generation while enforcing systematic verification for clinical reliability.
    \item We propose an automated and reliable benchmark construction framework based on EHR--LLM--KB interactions, where LLMs enable scalable template generation, KBs provide principled validation and enrichment, and EHR trajectories supply realistic longitudinal clinical evidence.
    \item We formulate clinical decision making as conditional inference over partially observed EHR data and design three representative tasks: diagnosis completion, treatment selection, and next-encounter prognosis, which require substantive biomedical knowledge and clinical inference over implicit clinical relations and longitudinal patient trajectories.
    \item We systematically benchmark more than 30 representative LLMs on \modelname{} and conduct comprehensive analyses of their accuracy, efficiency, and robustness, providing actionable insights for developing and evaluating clinically reliable LLM systems.
\end{itemize}

\section{Related Work}

\header{Medical QA Benchmarks.}
Medical QA benchmarks are essential for measuring the biomedical knowledge and reasoning capabilities of clinical decision-supporting models, including LLMs \cite{xiao2025beyond}. A large body of work constructs high-quality QA resources through expert curation or carefully designed evaluation protocols. These approaches typically improve correctness and reduce ambiguity, but they often limit dataset scale due to annotation cost and the need for domain expertise, including MedAlign \cite{fleming2024medalign}, SD-Bench \cite{nori2025sequential}, ExpertQA \cite{malaviya2024expertqa}, and MedThink-Bench \cite{zhou2025automating}, which typically contain several hundred expert-annotated QA pairs. Most existing large-scale medical benchmarks are derived from general narrative sources such as exams, textbooks, and clinical guidelines, including MedQA \cite{jin2021medqa}, MedMCQA \cite{pal2022medmcqa}, ClinicBench \cite{liu2024clinicbench}, MedXpertQA \cite{zuo2025medxpertqa}, MedChain \cite{liu2024medchain}, MedExQA \cite{kim2024medexqa}, LLM-Eval-Med \cite{zhang2025llmeval}, TrialPanorama \cite{wang2025trialpanorama}, CHBench \cite{guo2024chbench}, CMB \cite{wang2024cmb}, MedOdyssey \cite{fan2025medodyssey}, MedS-Bench \cite{wu2025towards}, MultiFacetEval \cite{zhou2024multifaceteval}, ReasonMed \cite{sun2025reasonmed}, XMedBench \cite{wang2024apollo}, and related evaluations of medical reasoning and generalization. In addition, several benchmarks are grounded in healthcare practice-generated clinical notes, case reports, or dialogue-style clinical interactions, such as MediSumQA \cite{dada2025medisumqa}, EHRNoteQA \cite{kweon2024ehrnoteqa}, ER-REASON \cite{mehandru2025erreason}, CPUCase \cite{perets2025cupcase}, LongHealth \cite{adams2025longhealth}, MedR-Bench \cite{qiu2025quantifying}, MMMU \cite{yue2024mmmu}, HealthBench \cite{arora2025healthbench}, DiagnosisArena \cite{zhu2025diagnosisarena}, and CRAFT-MD \cite{johri2024craft}. Safety-centered medical benchmarks further evaluate risk, harmfulness, and reliability in clinical contexts, such as MedSafetyBench \cite{han2024medsafetybench} and MedRisk or related risk-oriented agents \cite{liu2025riskagent}. Beyond general QA, some benchmarks focus on specialized competencies such as medical calculation \cite{khandekar2024medcalc}, concept-centric QA \cite{shoham2024medconceptsqa}, or epidemiological question answering \cite{wei2026epiqal}. A separate but related direction builds agentic or interactive environments for sequential diagnosis and decision support, including MEDIQ \cite{li2024mediq}, AI Hospital \cite{fan2025ai}, AgentClinic \cite{schmidgall2024agentclinic}, MAQUE \cite{gong2025dialogue}, VivaBench \cite{chiu2025simulating}, AgentHospital \cite{li2024agent}, MMD-Eval \cite{liu2025interactive}, and AMIE \cite{tu2025towards}. Moreover, multimodal information, including ECG, genomics, imaging, and other medical data, is becoming increasingly important for CDM by providing complementary evidence about patient physiology and disease status \cite{wang2026se,xie2024improving,xie2022survival,wang2026position,han2026towards}. This trend has motivated multimodal medical benchmarks such as Asclepius \cite{liu2025asclepius}, CLIMB \cite{dai2025climb}, EHRXQA \cite{bae2023ehrxqa}, GMAI-MMBench \cite{ye2024gmai}, OmniMedVQA \cite{hu2024omnimedvqa}, and PMC-VQA \cite{zhang2023pmc}. Despite this breadth, few benchmarks are built directly from raw structured EHR trajectories in a way that preserves real-world patterns required for CDM.

\header{\textbf{EHR QA Benchmarks}.}
A growing body of work leverages raw EHR data to construct QA datasets and benchmarks~\cite{bardhan2022drugehrqa}. However, most existing resources primarily assess the ability of a model to retrieve explicit facts from large, redundant tabular records, rather than to infer clinical decisions from longitudinal context \cite{xie2025hypkg}. A representative line of work frames EHR QA as text-to-SQL parsing or database querying, including benchmarks such as EHRSQL~\cite{lee2022ehrsql} and emrQA~\cite{pampari2018emrqa}, as well as systems that map questions to executable queries (e.g., emrKBQA~\cite{raghavan2021emrkbqa}, MIMICSQL~\cite{wang2020mimicsql}) or agentic coding workflows~\cite{xu2025medagentgym}. Complementary knowledge-graph-based approaches, such as ClinicalKBQA~\cite{wang2022attention} and MIMIC-SPARQL~\cite{park2021knowledge}, and temporal-reasoning benchmarks like TIMER~\cite{cui2025timer}, further enable relational and time-aware querying. Finally, concurrent efforts increasingly evaluate LLMs on clinical decision tasks (e.g., EHR-R1~\cite{liao2025ehr}), underscoring the importance and urgency of our work. In contrast to these efforts, prior work does not emphasize an automated and reliable EHR-LLM-KB pipeline that explicitly extracts clinical relations from raw structured EHRs and then systematically verifies and filters them using large-scale knowledge bases.

\header{\textbf{Positioning of Our Work.}}
Complementary to prior benchmarks that rely on curated narratives or emphasize retrieval-oriented EHR QA, our work targets realistic CDM evaluation by (i) grounding the benchmark in raw structured EHR trajectories, (ii) formulating three core CDM tasks that require substantive biomedical knowledge and clinical inference beyond information access, (iii) using LLMs to extract implicit clinical logic from raw EHR data for efficiency, and (iv) enforcing systematic verification and enrichment via biomedical KBs to maintain reliability.

\section{\modelname\ Construction Methodology}
\label{sec:method}

\begin{figure*}[htbp]
\centering
\includegraphics[width=0.87\textwidth]{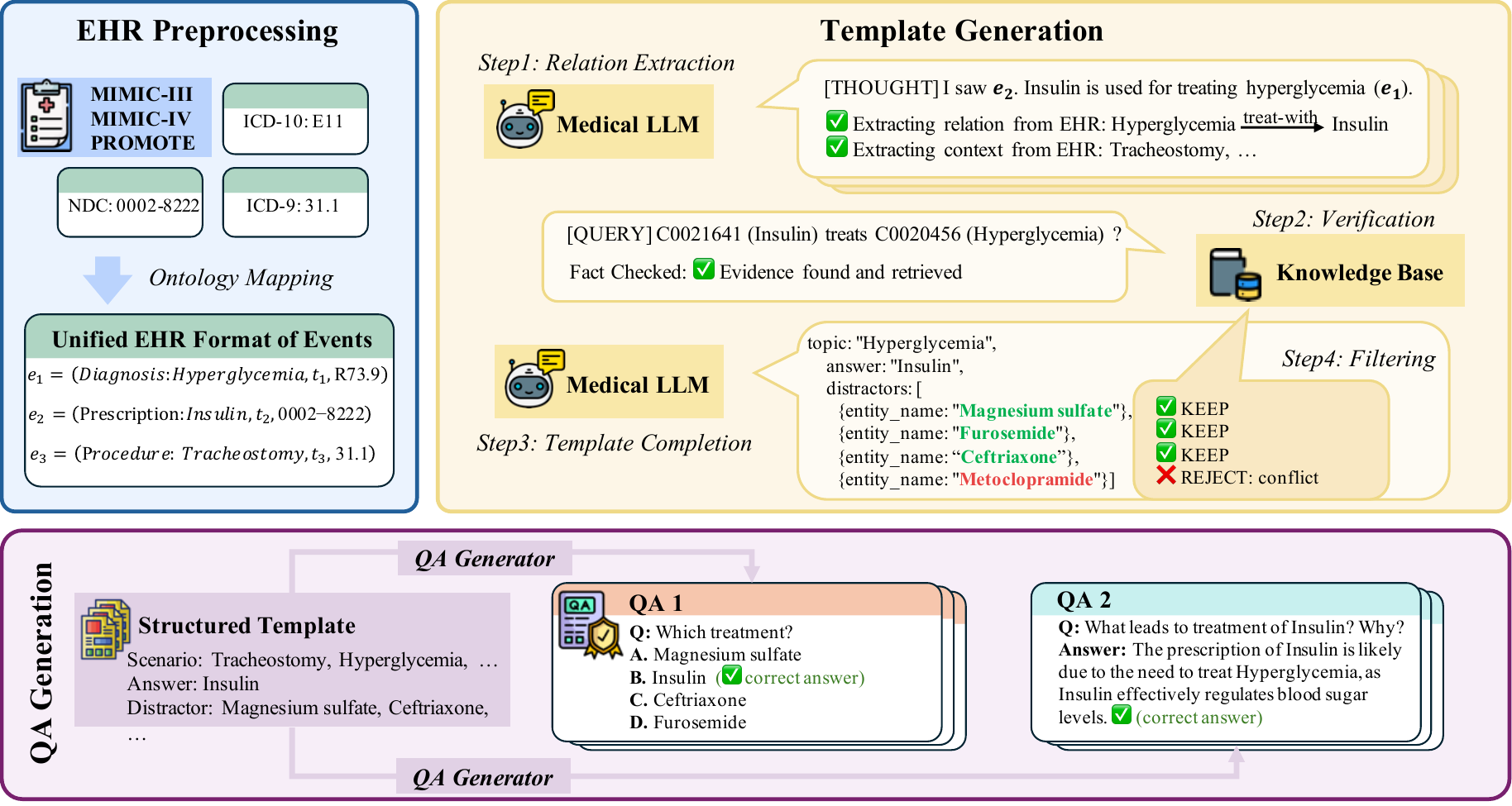}
\caption{Construction pipeline of \modelname{}. Starting from raw structured EHRs, we preprocess and normalize encounter-level clinical events into a standardized representation. We then generate structured templates integrating EHR signals, LLM-based extraction, and KB verification and enrichment. Finally, the QA generation module deterministically instantiates each template into multiple QA items for downstream evaluation. Overall, the pipeline is LLM-driven for scalability, KB-verified for reliability, and EHR-grounded for clinical relevance.}
\label{fig:framework}
\end{figure*}

\subsection{Problem Definition \& Framework}
\label{sec:problem_definition_framework}

Our goal is to transform structured EHRs into a clinically grounded QA benchmark for evaluating LLMs through an automated and reliable pipeline. Figure~\ref{fig:framework} presents an overview of the construction framework. The pipeline first preprocesses raw structured EHRs and normalizes clinical events into a standardized encounter-level representation. It then constructs templates through an automated EHR-LLM-KB interaction pipeline that extracts clinically meaningful signals and validates and enriches them through KB evidence. Finally, the pipeline instantiates each template into multiple QA variants and question formats, yielding task-specific QA items for evaluation. In summary, the pipeline is LLM-powered for scale, KB-checked for reliability, and grounded in real EHR data for clinical relevance. We describe each step in detail below.

\header{\textbf{EHR data collection \& representation.}}
Let
\begin{equation}
\mathcal{E} = \left\{ \mathcal{E}^{(1)}, \mathcal{E}^{(2)}, \dots, \mathcal{E}^{(N)} \right\}
\end{equation}
denote a cohort of EHRs from $N$ encounters, where $\mathcal{E}^{(n)}$ is the structured record associated with encounter $n$. We assume each encounter $\mathcal{E}^{(n)}$ is associated with a patient identifier $\pi(n)$, and encounters are ordered chronologically within each patient.

Each encounter-level EHR $\mathcal{E}^{(n)}$ is represented as a set of recorded clinical events:
\begin{equation}
\mathcal{E}^{(n)} = \left\{ e^{(n)}_{1}, e^{(n)}_{2}, \dots, e^{(n)}_{M_{n}} \right\},
\end{equation}
where $M_n$ is the number of events observed in encounter $n$.

Each event $e^{(n)}_m$ is represented as
\begin{equation}
e^{(n)}_m = \left( d^{(n)}_m, t^{(n)}_m, a^{(n)}_m \right),
\end{equation}
where $d^{(n)}_m$ is a textual description of the clinical event (e.g., diagnosis or prescription), $t^{(n)}_m$ is a timestamp, and $a^{(n)}_m$ denotes additional attributes such as medical codes or numerical values.

In practice, we follow common conventions by treating encounters as the basic temporal unit, aggregating clinical events at the encounter level rather than collapsing them to the patient level or relying on fine-grained timestamps. Purely patient-level aggregation is overly coarse because it ignores temporal structure and distinctions across encounters, while fine-grained timestamps are fragmented and reflect administrative logging rather than clinical onset (e.g., many diagnosis events are summarized as billing codes at discharge). Using encounters as the primary temporal unit aligns representations with documentation and CDM, enabling consistent aggregation over meaningful windows while allowing downstream tasks to focus on within-encounter evidence or longitudinal history.

Additional details regarding data sources and cohort construction are provided in Section~\ref{sec:data_collection}.

\header{\textbf{Template generation.}}
A template generation function is
\begin{equation}
g : \mathcal{E} \rightarrow \mathcal{P},
\end{equation}
which maps the encounter cohort $\mathcal{E}$ to $K$ structured QA templates:
\begin{equation}
\mathcal{P} = \left\{ P_k \right\}_{k=1}^{K}.
\end{equation}

The construction of $\mathcal{P}$ is a multi-stage interaction among EHR data, LLMs, and biomedical knowledge bases (KBs). Each template $P_k$ defines a clinically grounded blueprint that can be deterministically instantiated into one or more QA items. Each $P_k$ comprises:
\begin{itemize}[itemsep=1pt, topsep=0pt, parsep=0pt, partopsep=0pt]
  \item A template context $C_k \subset \mathcal{E}^{(n)}$, constructed by an LLM by selecting relevant events from an encounter record $\mathcal{E}^{(n)}$.
  \item A clinical relation $R_k=(x_k, r_k, y_k)$, where $x_k$ and $y_k$ are the subject and object entities and $r_k$ is the relation predicate that links them, such as $(\textit{Hypertension}, \textit{Cause}, \textit{Stroke})$. Each relation is extracted from EHR events using an LLM and verified against KBs to ensure clinical validity.
  \item A set of latent attributes $A_k$, generated by the LLM or retrieved from KBs, including entity definitions, evidence or rationale, candidate distractors, and the clinical topic associated with the relation.
\end{itemize}
All attributes in $P_k$ are exposed to the LLM during the subsequent QA generation stage to provide guidance. Additional details of the template generation procedure are provided in Section~\ref{sec:template_generation}.

\header{\textbf{QA generation.}}
A transformation function is defined as
\begin{equation}
    f : \mathcal{P} \rightarrow \mathcal{I}, \quad \text{where } \mathcal{I} = \left\{ \left( S_j, Q_j, B_j \right) \right\}_{j=1}^{J},
\end{equation}
which maps the constructed templates to a collection of $J$ constructed QA items. Each QA item consists of:
\begin{itemize}[itemsep=1pt, topsep=0pt, parsep=0pt, partopsep=0pt]
  \item a textual scenario $S_j$, which is a natural-language paragraph that verbalizes some background clinical events from a patient encounter;
  \item a natural-language question $Q_j$ constructed from the template,
  \item a metadata bundle $B_j$, including the choices, correct answer, clinical rationale, associated medical topic, and underlying clinical relations.
\end{itemize}
More details of the QA generation are provided in Section \ref{sec:qa_generation}.

\header{\textbf{Clinical decision tasks.}}
Using the formulation above, a collection of QA items $\mathcal{I}$ is constructed to target three core clinical decision tasks that require medical knowledge and inference. Each task corresponds to a conditional inference objective grounded in encounter-level EHR data $\mathcal{E}$, where the model receives a scenario $S^{(n)}$ composed of a subset of observed events.

\textbf{(I) Diagnosis decision (in-encounter diagnosis completion).}
This task evaluates intra-encounter diagnostic inference by predicting a missing diagnosis from other diagnoses recorded in the same encounter (referred to as ``diagnosis decision'' in this study for brevity). Given an encounter $n$ with diagnosis set $\mathcal{D}^{(n)}$, we withhold a target diagnosis $d^{(n)}_{\mathrm{tgt}} \in \mathcal{D}^{(n)}$ and create a scenario diagnosis subset
\begin{equation}
\mathcal{S}^{(n)} \subseteq \mathcal{D}^{(n)} \setminus \left\{ d^{(n)}_{\mathrm{tgt}} \right\}.
\end{equation}
The model is asked to infer the missing diagnosis:
\begin{equation}
d^{(n)}_{\mathrm{tgt}} \;\sim\;
p\!\left( d \;\middle|\; \mathcal{S}^{(n)} \right).
\end{equation}
Accordingly, the scenario description $S_j$ verbalizes $\mathcal{S}^{(n)}$, and the question asks for the most likely co-occurring diagnosis. This task measures whether the model captures clinically plausible co-morbidity patterns and diagnostic co-occurrence structure within a single encounter. 

\textbf{(II) Treatment decision (in-encounter treatment selection).}
This task models encounter-level treatment selection (referred to as ``treatment decision'' in this study for brevity). Given encounter $n$, a scenario diagnosis set is constructed as
\begin{equation}
\mathcal{S}^{(n)} \subseteq \mathcal{D}^{(n)}
\end{equation}
and the model is required to infer a target treatment $t^{(n)}_{\mathrm{tgt}}$ prescribed or performed during the same encounter:
\begin{equation}
t^{(n)}_{\mathrm{tgt}} \;\sim\;
p\!\left( t \;\middle|\; \mathcal{S}^{(n)} \right).
\end{equation}
Here, the scenario description $S_j$ verbalizes $\mathcal{S}^{(n)}$, and the question asks the model to select an appropriate treatment from $\mathcal{T}^{(n)}$ (i.e., a prescription or a procedure).

\textbf{(III) Prognosis decision (next-encounter outcome prediction).}
This task evaluates longitudinal reasoning over consecutive encounters to anticipate future diagnoses (referred to as ``prognosis decision'' in this study for brevity). Given two consecutive encounters $n$ and $n{+}1$ for the same patient, a scenario event set is constructed as
\begin{equation}
\mathcal{S}^{(n)} \subseteq \mathcal{D}^{(n)} \cup \mathcal{T}^{(n)}
\end{equation}
from encounter $n$, and the model is required to predict a target diagnosis $d^{(n+1)}_{\mathrm{tgt}}$ in the subsequent encounter:
\begin{equation}
d^{(n+1)}_{\mathrm{tgt}} \;\sim\;
p\!\left( d \;\middle|\; \mathcal{S}^{(n)} \right).
\end{equation}
Here $\mathcal{D}^{(n)}$ and $\mathcal{T}^{(n)}$ denote the diagnoses and treatments (including procedures and prescriptions) observed at encounter $n$. For each QA item $j$, the scenario description $S_j$ is a natural-language rendering of $\mathcal{S}^{(n)}$, and the question asks for a diagnosis that appears at encounter $n{+}1$. This task evaluates whether disease progression and treatment-related effects over time under partial observation are captured.

Overall, the resulting benchmark is designed to systematically evaluate LLMs’ ability to perform clinically grounded reasoning and decision-making over structured, longitudinal EHR data under partial observation. Details of the evaluation protocol are summarized in Appendix~\ref{sec:evalution_protocol}.

\subsection{Data Collection \& Preprocessing}
\label{sec:data_collection}

Our benchmark utilizes structured EHR trajectories from three real-world sources: MIMIC-III, MIMIC-IV, and PROMOTE. MIMIC-III (Version 1.4) is a widely-used, publicly available critical-care dataset from intensive care units at Beth Israel Deaconess Medical Center between 2001 and 2012 \cite{johnson2016mimic-iii}. MIMIC-IV (Version 3.1) is a newer release that extends MIMIC-III with updated hospital data from the same institution (2008--2022) \cite{johnson2023mimic-iv}. To further reduce potential contamination from public corpora and evaluate language models in a setting less prone to data leakage, we additionally include PROMOTE, a private dataset from Emory Healthcare spanning 2012--2021 \cite{xie2025promote1,wu2025promote2}.

Across all sources, we treat an inpatient stay as the basic encounter unit. For each encounter, we extract billing-code-derived clinical events, including diagnoses and treatments, where treatments cover both medical procedures and medication prescriptions. During preprocessing, we normalize heterogeneous source schemas into a unified event representation with (i) a standardized event type in ``\{\textit{diagnosis}, \textit{procedure}, \textit{prescription}\}'', (ii) a human-readable event description mapped from raw clinical codes (e.g., ICD), and (iii) a consistent encounter timeline ordering. 
We mainly use \textit{PyHealth} \cite{pyhealth2023yang} to extract information from raw EHRs and perform preprocessing.

From the preprocessed EHRs, we define an \emph{EHR instance} as the minimal input unit presented to the LLM. For the diagnosis and treatment tasks, each instance $u^{(i)}$ corresponds to a single encounter $\mathcal{E}^{(n)}$. For the prognosis task, each instance $u^{(i)}$ corresponds to a pair of consecutive encounters $(\mathcal{E}^{(n)}, \mathcal{E}^{(n+1)})$ from the same patient. Detailed cohort statistics are summarized in Appendix \ref{sec:qa_statistics}.

\subsection{Template Generation}
\label{sec:template_generation}
After preprocessing the EHR cohort, structured templates $\mathcal{P}$ are constructed from $\mathcal{E}$. Each template $P_k$ specifies a clinically grounded blueprint that can be deterministically instantiated into QA items, following the formulation in Section~\ref{sec:problem_definition_framework}. Specifically, $P_k$ includes a template context $C_k$, a target clinical relation $R_k=(x_k,r_k,y_k)$, and latent attributes $A_k$ that support downstream generation. Template construction is implemented through a multi-stage interaction among EHR data, a medical LLM, and a biomedical knowledge base.

\header{\textbf{Stage 1 (Relation extraction: EHR $\rightarrow$ LLM $\rightarrow$ KB).}}
For each EHR input instance $u^{(i)}$, an instruction-fine-tuned medical LLM (specifically, HuatuoGPT-o1-8B~\cite{chen2024huatuogpto1}) is prompted to extract clinically salient relations from the patient record under strict JSON output constraints. The objective is to capture implicit clinical logic encoded in structured EHR data. Extracted relations are deduplicated and assigned unique identifiers, producing candidate relation triplets $R_k=(x_k,r_k,y_k)$ that later QA items target, such as \textit{(Hyperglycemia, Treat-with, Insulin)}, together with an associated rationale. The LLM is also prompted to extract a small set of auxiliary context events, such as \textit{Tracheostomy} and \textit{Hemothorax}, which are aggregated into $C_k$. These auxiliary events are constrained to have no lexical or semantic overlap with any entity in $R_k$, meaning they cannot directly repeat or paraphrase entities appearing in the target relation triplet. Outputs of this stage are passed to KB for verification. More details about the extraction of clinical relations and context from raw EHR data are presented in Appendix \ref{sec:appendix_relation_extraction}. 

\header{\textbf{Stage 2 (Relation verification and enrichment: KB $\rightarrow$ LLM).}}
In this stage, relations extracted in Stage~1 are validated and enriched through external biomedical evidence by querying a composite KB that integrates UMLS \cite{bodenreider2004umls}, SemMedDB \cite{kilicoglu2012semmeddb}, DrugBank \cite{wishart2008drugbank}, PubMed \cite{canese2013pubmed}, and ICD \cite{o2005icd}. UMLS provides standardized concept identifiers (CUIs) and textual definitions across biomedical vocabularies \cite{bodenreider2004umls}. SemMedDB provides semantic relations (e.g., \textit{Cause} and \textit{Treat-with}) extracted from PubMed abstracts, thereby offering literature-supported evidence for relations between biomedical concepts \cite{kilicoglu2012semmeddb}. Event strings extracted from EHRs are first resolved to standardized concepts through the UMLS API, mapping each entity to a UMLS CUI with source vocabularies such as ICD and DrugBank. After concept linking, evidence for relations is retrieved through SemMedDB.

Given an LLM-extracted clinical relation from a patient EHR record $R_k = (x_k, r_k, y_k)$, we verify its validity by checking for supporting evidence in SemMedDB, which contains KB relation triplets automatically extracted from PubMed abstracts. A candidate relation \( R_k \) is retained only if it satisfies all three criteria:
\begin{itemize}[itemsep=1pt, topsep=0pt, parsep=0pt, partopsep=0pt]
    \item Positive support: SemMedDB contains evidence supporting the relation, such as $(x_k, \textit{Cause}, y_k)$ or $(x_k, \textit{Treat-with}, y_k)$.
    \item No negative evidence: SemMedDB does not contain contradictory relations, such as $(x_k, \textit{Neg-cause}, y_k)$.
    \item No conflicting background evidence: No contradictory relation is found with respect to a predefined set of background concepts $C_k$, such as $(C_k, \textit{Neg-cause}, y_k)$.
\end{itemize}
These checks ensure that each retained relation reflects a clinically valid association supported by biomedical knowledge, thereby reducing the risk of hallucinated relations introduced by LLMs.

To support downstream QA generation, entity definitions are retrieved from UMLS, and evidence sentences are retrieved from PubMed through SemMedDB. The verified relations, together with their associated definitions and evidence, are stored in structured templates $P_k$ and used in subsequent QA construction. More details about how to use the KB are provided in Appendix \ref{sec:appendix_umls_linking}.

\header{\textbf{Stage 3 (Template completion: LLM $\rightarrow$ KB).}}
In this stage, each verified template $P_k$ is completed by prompting the LLM to generate additional structured attributes under a strict JSON schema. For each verified relation $R_k=(x_k,r_k,y_k)$, the LLM produces: (i) a set of distractor candidates that compete with the object entity $y_k$ in the relation, e.g., \textit{Magnesium sulfate} and \textit{Furosemide} as distractors for \textit{Insulin}; these candidates are generated by the LLM or sampled from EHR data of other patients to capture both model knowledge and real-world patterns; (ii) a high-level clinical condition topic associated with the relation, e.g., \textit{Hyperglycemia}; and (iii) a concise rationale that summarizes the relation together with KB-retrieved evidence, e.g., \textit{``The prescription of Insulin is likely due to the need to treat Hyperglycemia, as Insulin effectively regulates blood sugar levels.''} The resulting attributes are stored in the updated template $P_k$ and used for downstream QA generation. Additional details of the generated templates are provided in Appendix~\ref{sec:qa_templates}.

\header{\textbf{Stage 4 (Template filtering: KB $\rightarrow$ Template Output).}}
In the final stage, unqualified or misleading distractors within templates are removed via KB verification to preserve an unambiguous answer set. Following the procedure from Stage 2, each distractor term is resolved to a UMLS CUI, and SemMedDB is queried for supporting predicate evidence. A distractor is filtered out if it forms any clinically supported relation that would render it a plausible correct answer given the question context. Specifically, a distractor is removed if: 
\begin{itemize}[itemsep=1pt, topsep=0pt, parsep=0pt, partopsep=0pt]
\item SemMedDB provides positive evidence linking the distractor to the subject entity $x_k$ under a compatible predicate, such as $(x_k, \text{Cause}, \text{distractor})$. 
\item SemMedDB provides positive evidence linking the distractor to any auxiliary context event in $C_k$ under a compatible predicate, such as $(C_k, \text{Cause}, \text{distractor})$. 
\end{itemize}
After filtering, between three and five distractors are retained per template; templates failing to meet the minimum distractor count are discarded to ensure QA quality.

\subsection{QA Generation}
\label{sec:qa_generation}

Each template $P_k$ provides (i) a context $C_k$, (ii) a verified clinical relation $R_k=(x_k,r_k,y_k)$, and (iii) latent attributes $A_k$, including entity definitions, supporting evidence or rationale, candidate distractors, and an associated clinical topic. The templates are provided to an LLM to instantiate multiple types of QA items.

For each QA item $I_j$, an event-complete scenario $S_j$ is constructed by augmenting the template context with the relation subject entity, i.e., $S_j \leftarrow C_k \cup \{x_k\}$. This design grounds $S_j$ in observed encounter-level clinical events while explicitly tying the scenario to the verified relation, thereby providing faithful background information for question construction.

Within each template, multiple-choice questions (MCQs) are instantiated with an option count $c \in \{4, 5, 6\}$. For each task, a task-specific question skeleton is used to construct $Q_j$; for example, in the prognosis task, the question is phrased as \emph{``Given the prior clinical history summarized above, what diagnosis may occur at the next encounter?''} During evaluation, the tested LLM receives $(S_j, Q_j)$ as input. For a given $c$, one correct answer is designated and the remaining $c-1$ options are filled with distractors retrieved from the template. The explanation is taken from the template rationale, generated by integrating real-world EHR patterns, internal LLM knowledge, and KB-retrieved evidence. To increase diversity, each question is paraphrased and each choice set is permuted to create multiple MCQ versions. Additional details of the question skeleton and question paraphrasing are provided in Appendix~\ref{sec:qa_generation}.

Open-ended questions (OEQs) are also constructed to elicit a free-text response and a corresponding explanation aligned with the target clinical relation $R_k=(x_k,r_k,y_k)$. These questions follow the same three clinical decision tasks and task-specific skeletons to form $Q_j$; for example, an OEQ is phrased as \emph{``Given the prior clinical history summarized above, what event may lead to acute kidney failure at the next encounter? Why?''} The gold-standard answer is defined as the rationale generated during template completion by integrating real-world EHR patterns, internal model knowledge, and KB-retrieved evidence.

For each template $P_k$ and option count $c$, MCQs are instantiated for 4-choice MCQ (4 paraphrased versions), 5-choice MCQ (5 versions), 6-choice MCQ (6 versions), and 1 OEQ. As a result, each template yields at most 16 QA items across four question types, enabling evaluation under different response constraints. Finally, the generated benchmark contains 960{,}067 QA items. QA statistics are provided in Appendix~\ref{sec:qa_statistics}.

\section{Experiments}
\label{sec:experiments}

\subsection{Benchmarking LLMs on \modelname{} Across Clinical Decision Tasks}
\label{sec:main_results}

In our main experiments, we evaluate a comprehensive set of 31 representative LLMs on the constructed benchmark dataset. Details of all utilized LLMs in this study are provided in Appendix \ref{sec:llm_modelcard}. Specifically, the LLM used for \modelname{} generation (HuatuoGPT-o1-8B) is not evaluated here to avoid bias. These evaluated models are categorized into three primary groups:
\begin{enumerate}[itemsep=1pt, topsep=0pt, parsep=0pt, partopsep=0pt, label=(\alph*)]
    \item \textbf{Open source general LLMs} that serve as critical performance baselines and widely accessible tools for comparative analysis: (a.1) glm4-9b and glm4-32b~\cite{glm2024chatglm}  (a.2) llama3-8b, llama3-70b, llama3.1-8b, llama3.2-3b, llama3.3-70b~\cite{grattafiori2024llama3}; (a.3) mistral-7b, mistral-small3-24b~\cite{jiang2023mistral}, and ministral-8b~\cite{liu2026ministral}; (a.4) qwen2.5-3b, qwen2.5-7b and qwen2.5-32b~\cite{qwen2.5}; (a.5) qwen3-4b, qwen3-8b, and qwen3-32b~\cite{yang2025qwen3}; (a.6) smollm3-3b~\cite{bakouch2025smollm3}; (a.7) yi-1.5-9b and yi-1.5-34b \cite{young2024yi}.
    \item \textbf{Medical LLMs} that are pretrained on healthcare corpora and specialized for tackling medical tasks: (b.1) doctor-r1-8b~\cite{lai2025doctor}; (b.2) med42-8b ~\cite{med42v2}; (b.3) ultramedical-8b~\cite{zhang2024ultramedical}; (b.4) m1-7b-23k and m1-32b-1k~\cite{huang2025m1}.
    \item \textbf{HIPAA compliant API-based LLMs} that ensure the secure processing of protected health information: (c.1) gpt-4.1-nano, gpt-4.1-mini, and gpt-4.1~\cite{achiam2023gpt4}; (c.2) gpt-5-nano, gpt-5-mini, gpt-5, gpt-5.2~\cite{singh2025gpt5}.
\end{enumerate}

\begin{table*}[htbp]
\centering
\caption{Benchmarking LLMs on \modelname{} across tasks, data sources, and question types. We use abbreviations Dx/Tx/Px for diagnosis/treatment/prognosis decision task, MIII/MIV/PRO for MIMIC-III/MIMIC-IV/PROMOTE, and 4C/5C/6C for 4/5/6-choice MCQs. Within each column, we mark the top-8 results (ranked first 25\%) using underlines and rank superscripts: \best{} \second{} \third{} \fourth{} \fifth{} \sixth{} \seventh{} \eighth{}.}
\label{tab:main_benchmark_overview}
\resizebox{0.93\linewidth}{!}{
\begin{tabular}{l|c|cc|ccc|ccc|ccc}
\toprule
\multirow{2}{*}{\textbf{Model}}
&\textbf{Overall}
&\multicolumn{2}{c|}{\textbf{Rank}}
&\multicolumn{3}{c|}{\textbf{Task Acc}}
&\multicolumn{3}{c|}{\textbf{Source Acc}}
&\multicolumn{3}{c}{\textbf{Type Acc}}\\
&\textbf{Acc (\%)$\uparrow$}&\textbf{Avg $\downarrow$}&\textbf{SD.}
&\textbf{Dx (\%)}&\textbf{Tx (\%)}&\textbf{Px (\%)}
&\textbf{MIII (\%)}&\textbf{MIV (\%)}&\textbf{PRO (\%)}
&\textbf{4C (\%)}&\textbf{5C (\%)}&\textbf{6C (\%)}\\
\midrule
\multicolumn{13}{l}{\textit{Open source general LLMs}}\\
\midrule
glm4-9b & 59.62 & 16.70 & 2.52 & 58.36 & 72.61 & 47.90 & 61.92 & 61.11 & 56.41 & 64.76 & 59.47 & 54.64 \\
glm4-32b & \seventh{66.12} & \seventh{7.05} & 2.44 & \sixth{67.09} & \sixth{77.90} & 53.36 & 66.27 & \eighth{66.45} & \fourth{65.85} & \eighth{70.81} & \eighth{65.89} & \seventh{61.66} \\
llama3-8b & 48.90 & 24.23 & 1.13 & 44.61 & 63.44 & 38.63 & 51.19 & 48.60 & 47.74 & 54.87 & 48.00 & 43.82 \\
llama3-70b & 63.35 & 10.72 & 3.38 & 62.21 & \seventh{77.63} & 50.20 & 65.39 & 63.72 & 61.20 & 68.41 & 63.19 & 58.45 \\
llama3.1-8b & 56.76 & 19.41 & 3.74 & 53.82 & 73.45 & 43.02 & 57.50 & 56.77 & 55.79 & 62.67 & 56.66 & 50.97 \\
llama3.2-3b & 49.85 & 23.67 & 1.22 & 43.18 & 65.41 & 40.96 & 50.16 & 51.20 & 48.46 & 55.79 & 48.78 & 44.99 \\
llama3.3-70b & \fourth{67.28} & \fourth{5.23} & 2.08 & \third{68.35} & \fourth{79.05} & \seventh{54.44} & \fifth{68.74} & \fourth{67.94} & \sixth{65.35} & \fourth{71.98} & \fourth{67.07} & \fourth{62.79} \\
mistral-7b & 38.23 & 28.04 & 2.13 & 36.59 & 41.90 & 36.21 & 38.54 & 38.25 & 37.85 & 40.16 & 38.56 & 35.98 \\
ministral-8b & 56.48 & 20.32 & 1.64 & 53.11 & 71.97 & 44.37 & 58.54 & 57.39 & 54.31 & 61.84 & 55.79 & 51.82 \\
mistral-small3-24b & 65.01 & 9.52 & 4.49 & 66.20 & 75.02 & \eighth{53.81} & \eighth{67.00} & 65.36 & 63.38 & 69.41 & 64.19 & 61.42 \\
qwen2.5-3b & 37.87 & 28.98 & 1.07 & 34.94 & 47.75 & 30.93 & 39.72 & 39.21 & 35.23 & 47.93 & 36.09 & 29.59 \\
qwen2.5-7b & 57.74 & 18.99 & 2.30 & 56.27 & 72.04 & 44.91 & 59.76 & 59.25 & 54.98 & 62.22 & 57.47 & 53.52 \\
qwen2.5-32b & 64.97 & 8.81 & 2.95 & \seventh{66.48} & 76.87 & 51.54 & 65.22 & \seventh{67.00} & 63.05 & 69.80 & 64.50 & 60.59 \\
qwen3-4b & 60.63 & 14.99 & 2.77 & 59.67 & 73.46 & 48.76 & 62.01 & 61.73 & 58.39 & 66.37 & 60.24 & 55.27 \\
qwen3-8b & 60.87 & 14.27 & 2.84 & 58.19 & 74.49 & 49.93 & 62.26 & 62.11 & 58.74 & 67.09 & 60.21 & 55.31 \\
qwen3-32b & \sixth{66.78} & \sixth{6.54} & 2.14 & \fifth{67.97} & \eighth{77.34} & \sixth{55.04} & \seventh{68.48} & \sixth{67.18} & \fifth{65.55} & \sixth{71.33} & \sixth{66.47} & \fifth{62.55} \\
smollm3-3b & 45.82 & 25.84 & 1.34 & 40.79 & 58.29 & 38.39 & 46.31 & 46.08 & 45.04 & 51.01 & 45.09 & 41.37 \\
yi-1.5-9b & 45.51 & 25.88 & 1.37 & 41.97 & 57.52 & 37.05 & 46.87 & 45.50 & 44.84 & 53.15 & 45.64 & 37.74 \\
yi-1.5-34b & 58.94 & 17.65 & 2.65 & 56.70 & 72.25 & 47.86 & 60.72 & 60.07 & 56.56 & 64.48 & 58.61 & 53.71 \\
\midrule
\multicolumn{13}{l}{\textit{Medical LLMs}}\\
\midrule
doctor-r1-8b & 61.07 & 14.06 & 2.32 & 58.74 & 74.49 & 49.98 & 61.94 & 62.26 & 59.01 & 67.07 & 60.56 & 55.57 \\
med42-8b & 36.48 & 29.31 & 1.04 & 33.48 & 45.54 & 30.41 & 36.88 & 35.00 & 37.45 & 38.69 & 38.41 & 32.34 \\
ultramedical-8b & 29.02 & 30.60 & 0.69 & 19.14 & 43.09 & 24.83 & 31.17 & 29.56 & 27.99 & 38.76 & 28.26 & 20.05 \\
m1-7b-23k & 46.08 & 26.01 & 1.82 & 38.42 & 63.07 & 36.74 & 46.50 & 46.99 & 45.14 & 50.03 & 45.75 & 42.45 \\
m1-32b-1k & 63.21 & 11.47 & 3.64 & 63.07 & 74.84 & 51.73 & 62.87 & 65.49 & 61.46 & 68.49 & 62.80 & 58.35 \\
\midrule
\multicolumn{13}{l}{\textit{HIPAA compliant API-based LLMs}}\\
\midrule
gpt-4.1-nano & 60.48 & 15.09 & 2.28 & 58.02 & 74.03 & 49.39 & 61.59 & 61.89 & 58.28 & 65.42 & 60.39 & 55.63 \\
gpt-4.1-mini & \fifth{66.79} & \fifth{6.28} & 1.91 & \eighth{66.41} & \fifth{77.90} & \fifth{56.05} & \sixth{68.58} & \fifth{67.36} & \seventh{64.45} & \fifth{71.34} & \fifth{66.76} & \sixth{62.26} \\
gpt-4.1 & \second{69.43} & \second{2.51} & 1.32 & \second{69.87} & \third{80.10} & \third{58.33} & \second{70.59} & \second{69.77} & \third{67.87} & \second{73.97} & \second{69.21} & \second{65.11} \\
gpt-5-nano & 57.80 & 19.31 & 1.93 & 56.39 & 70.86 & 46.16 & 58.64 & 58.72 & 55.68 & 63.27 & 57.86 & 52.29 \\
gpt-5-mini & \eighth{66.12} & \eighth{7.84} & 3.48 & 65.40 & 76.17 & \fourth{56.79} & \fourth{69.08} & 66.05 & \eighth{63.88} & \seventh{70.87} & \seventh{65.98} & \eighth{61.51} \\
gpt-5 & \third{69.06} & \third{3.21} & 1.86 & \fourth{68.26} & \best{80.45} & \second{58.46} & \third{70.16} & \third{69.18} & \second{68.26} & \third{73.64} & \third{68.92} & \third{64.61} \\
gpt-5.2 & \best{70.91} & \best{1.69} & 1.10 & \best{72.02} & \second{80.13} & \best{60.59} & \best{71.50} & \best{71.06} & \best{70.70} & \best{75.40} & \best{70.53} & \best{66.81} \\
\bottomrule
\end{tabular}}
\end{table*}

We benchmark 31 representative LLMs on \modelname{} to assess their clinical decision-making capability under a unified inference protocol. Specifically, we evaluate three core tasks (Diagnosis/Treatment/Prognosis), three data sources (MIMIC-III/MIMIC-IV/PROMOTE), and three multiple-choice settings (4-choice/5-choice/ 6-choice). We report accuracy at multiple granularities (task-level, source-level, and type-level), and additionally compute each model's overall accuracy and its rank statistics (mean and standard deviation) across settings, where per-setting ranks are obtained by sorting models by accuracy within each setting and then aggregating ranks across all settings. Further details regarding the experiment protocol, including batching, deterministic decoding, hardware, and the fixed evaluation subset, are provided in Appendix~\ref{sec:implementation_details}. The aggregated accuracy results are summarized in Table~\ref{tab:main_benchmark_overview}. Additional results for cost analysis and error analysis are provided in Appendix~\ref{sec:cost_analysis} and \ref{sec:error_analysis}.

Overall, the benchmark results are broadly consistent with established model capability trends, with the highest-ranked systems corresponding to the most capable and recently released models in our evaluation set, validating the construction pipeline of \modelname{}. Specifically, gpt-5.2, gpt-4.1, gpt-5, llama3.3-70b, gpt-4.1-mini, qwen3-32b, glm4-32b, and gpt-5-mini emerge as the strongest performers. Among them, gpt-5.2 achieves the highest overall accuracy of 70.91\% with the best average rank of 1.69 and a low rank standard deviation of 1.10, indicating stable performance across tasks, sources, and question types. Meanwhile, the leading open-source models remain highly competitive: llama3.3-70b attains 67.28\% and qwen3-32b attains 66.78\%, narrowing the gap to API-based models to only 3-4 absolute points. Beyond the leaderboard, the relative ordering within model families follows expected scaling and generation trends. For example, within the Qwen series, qwen3-32b substantially outperforms smaller counterparts such as qwen3-8b (60.87\%) and qwen3-4b (60.63\%), and also improves over the previous-generation qwen2.5-32b (64.97\%). Collectively, these patterns suggest that \modelname{} reliably captures meaningful capability differences consistent with model capacity.

Performance varies substantially across clinical decision tasks: treatment selection consistently yields the highest accuracy, whereas prognosis prediction is the most challenging. Overall, the average accuracy across all models and all questions follows the ordering as Tx $>$ Dx $>$ Px (69.33\% $>$ 55.02\% $>$ 46.67\% ). This pattern is clinically intuitive. Treatment selection often depends on relatively direct, well-documented associations between medications and their indications, which are explicitly described in drug labels and consolidated in clinical practice guidelines. In contrast, diagnosis and prognosis tasks emphasize disease-to-disease causal and progression relations, which are typically less explicit, more confounded by comorbidities, and harder to infer from limited encounter evidence. Prognosis further requires anticipating conditions beyond the current visit by integrating longitudinal trajectories and subtle risk factors, making it inherently more difficult than in-encounter decision-making. Despite this difficulty, diagnosis completion and especially prognosis are critical for real-world care, underscoring the need to strengthen LLMs for longitudinal reasoning and forward-looking clinical prediction to support clinicians.

Dataset-source-level accuracies exhibit only moderate variation compared with the stronger task- and type-level effects. When aggregating across all tasks and all evaluated models, the accuracies on MIMIC-III/MIMIC-IV/PROMOTE are 58.26\%/57.69\%/55.45\%, suggesting that performance on the two public datasets and the private dataset is broadly comparable. The consistent trends across both public and private data support the robustness of our pipeline, indicating that \modelname{} can be instantiated on heterogeneous EHR sources while yielding benchmarks, capturing CDM behaviors with similar difficulty and discriminative power.

Increasing the number of multiple-choice options leads to a clear and consistent decline in accuracy. The average accuracy across all models and all questions decreases from 62.29\% (4C) to 56.69\% (5C) to 52.04\% (6C). Such monotonic trends are expected under a well-constructed multiple-choice benchmark, as additional options increase confusability and reduce the probability of correct selection under uncertainty. The consistent degradation across models therefore provides evidence that the \modelname{} pipeline produces valid question instances whose difficulty is appropriately controlled by the number of answer choices.

When comparing medical LLMs to the general-purpose base models they are adapted from, we do not observe consistent gains from medical fine-tuning on \modelname{}. For example, m1-32b-1k achieves 63.21\% overall accuracy, close to its base model qwen2.5-32b at 64.97\%. Appendix~\ref{sec:medical_vs_base} provides a more detailed breakdown. Similar observations have also been reported in prior work~\cite{xu2025medagentgym,dorfner2025evaluating}. These results suggest that current medical-domain specialization still leaves important gaps for EHR-grounded clinical decision-making. In \modelname{}, strong performance requires reasoning over patients' real longitudinal EHR context and answering questions that demand both biomedical knowledge and nontrivial inference, including disentangling confounded relations such as disease--disease progression and disease--treatment associations. Improving these capabilities likely requires training signals beyond domain text exposure, such as large-scale clinical case supervision and decision-focused objectives, for which EHR-grounded resources like \modelname{} may provide a useful foundation.

We also conduct additional analyses to examine whether the main experiment results are sensitive to benchmark construction choices or can be explained by shallow matching heuristics. Specifically, Appendix~\ref{sec:source_model_bias} evaluates whether the benchmark results are affected by the LLM used for QA generation, and Appendix~\ref{sec:context_event_size} studies whether changing the number of local EHR context events alters the observed performance trends. In addition, Appendix~\ref{sec:retrieval_baselines} compares LLMs with embedding-based non-LLM retrieval baselines under the same zero-shot QA setting. These analyses show that the main findings are robust to key construction choices and that strong performance on \modelname{} cannot be reduced to simple question-option semantic similarity matching.

\subsection{Additional Analyses and Validation}
\label{sec:additional_analyses}

Beyond the main benchmark results in Section~\ref{sec:main_results}, which focus on comparing representative LLMs under a unified zero-shot multiple-choice setting, we conduct several experiments with different settings in the Appendix to validate the reliability of the evaluation protocol and the stability of the main conclusions. 

\begin{itemize}[itemsep=1pt, topsep=0pt, parsep=0pt, partopsep=0pt]
    \item In Appendix~\ref{sec:reasoning_models}, we separately benchmark reasoning-oriented LLM configurations because explicit intermediate reasoning substantially increases token usage and would otherwise confound the direct model comparison in Section~\ref{sec:main_results}. This controlled study characterizes the accuracy-efficiency trade-off under different reasoning-effort settings, showing that additional reasoning generally improves performance while incurring higher token cost. This trend is consistent with expected scaling behavior and further supports the validity of the \modelname{} pipeline.

    \item In Appendix~\ref{sec:multi_version}, we evaluate multiple paraphrased question versions to test whether model performance is sensitive to surface-level wording. The consistently low variability and high prediction consistency across versions indicate that single-version evaluation in the main experiment provides a stable proxy for underlying CDM capability.

    \item In Appendix~\ref{sec:full_eval}, we run an additional evaluation on the extended question set covering all verified QA templates to examine whether the fixed-subset protocol introduces sampling artifacts. The near-identical model ranking and trend patterns suggest that the selected subset in the main experiment provides sufficient coverage for fair comparison at scale, further supporting the reliability of the generation pipeline.

    \item In Appendix~\ref{sec:open_questions}, we evaluate paraphrased open-ended questions to probe free-form clinical reasoning beyond multiple-choice selection. The performance trends remain consistent with the main benchmark, supporting the reliability of the open-ended question pipeline.
\end{itemize}

Collectively, these analyses show that the conclusions drawn from the main benchmark are robust to key evaluation design choices. They also provide additional evidence that \modelname{} offers a stable and reliable framework for evaluating EHR-grounded clinical decision-making capabilities.
\section{Conclusion}

In this work, we develop \modelname{} via an automated and reliable pipeline based on \emph{EHR--LLM--KB} interaction. The pipeline (i) converts encounter-level EHR trajectories into structured templates, (ii) deterministically instantiates these templates into large-scale QA items with controlled variants for robust evaluation, and (iii) applies KB-based verification and enrichment to improve reliability.

Under a unified inference protocol, the benchmarking of more than 30 representative LLMs on \modelname{} yields consistent trends that validate the benchmark. Recently released high-capability models achieve the strongest performance, treatment selection is consistently easier than the other two tasks, dataset effects remain modest, and current medical fine-tuning does not deliver consistent gains over the corresponding general-purpose base models.  Additional analyses further confirm the reliability of the \modelname{} construction pipeline and the evaluation protocol used in the main experiments. Collectively, these results provide actionable insights that can inform the design and evaluation of clinically reliable LLM systems in EHR-grounded medical decision making.

Overall, \modelname{} provides an automated and reliable benchmark with 960{,}067 QA items for evaluating LLM-based clinical decision making grounded in real-world structured EHR trajectories. We hope that \modelname{} will serve as a practical testbed to accelerate the development of clinically reliable LLM systems and to facilitate transparent and reproducible progress in EHR-grounded medical decision making.

\section*{Ethical Statement}
\label{sec:ethical_statement}

This study was conducted in full compliance with established ethical and data governance standards. MIMIC-III and MIMIC-IV are publicly available credentialed datasets accessed under the PhysioNet Credentialed Data Use Agreement and all relevant data usage policies. PROMOTE is a private dataset that was fully de-identified prior to use, and its use was approved by the Emory Institutional Review Board (IRB Protocol 2025P010425).

All raw EHR data were processed locally on HIPAA-compliant systems into structured templates and QA pairs and were not directly exposed to the evaluated LLMs. The released benchmark contains only de-identified QA items rather than raw patient records. We additionally designed the benchmark construction pipeline to reduce information leakage through controlled context generation, KB verification, and filtering procedures. No patient re-identification was attempted at any stage of the study.

\section*{Acknowledgement}
This research was partially supported by internal funds and GPU resources provided by Emory University, the U.S. National Science Foundation (Awards 2442172, 2312502, and 2319449), and the U.S. National Institutes of Health (Awards K25DK135913, RF1NS139325, R01DK143456, U18DP006922, and R01HL166233).

\newpage
\bibliographystyle{ACM-Reference-Format}
\balance
\bibliography{main}

\newpage
\appendix
\onecolumn
\section*{Appendices}

\section{Notation Table}
\label{app:notation_table}

\begin{table}[htbp!]
\centering
\caption{Notations used in the paper.}
\label{tab:notations}
\resizebox{0.9\linewidth}{!}{
\begin{tabular}{cl}
\toprule
\textbf{Notation} & \textbf{Description} \\
\midrule
$\mathcal{E}$ & Cohort of structured EHR encounter records. \\
$N$ & Number of encounters in the cohort $\mathcal{E}$. \\
$\mathcal{E}^{(n)}$ & Structured EHR record for encounter $n$. \\
$\pi(n)$ & Patient identifier associated with encounter $n$. \\
$e^{(n)}_{m}$ & The $m$-th clinical event in encounter $n$. \\
$M_n$ & Number of events observed in encounter $n$. \\
$d^{(n)}_{m}$ & Textual description of event $e^{(n)}_{m}$ (e.g., diagnosis/prescription). \\
$t^{(n)}_{m}$ & Timestamp of event $e^{(n)}_{m}$. \\
$a^{(n)}_{m}$ & Additional attributes of event $e^{(n)}_{m}$ (e.g., codes / numeric values). \\
$g$ & Template generation function mapping $\mathcal{E}$ to template set $\mathcal{P}$. \\
$\mathcal{P}$ & Set of constructed QA templates. \\
$K$ & Number of templates in $\mathcal{P}=\{P_k\}_{k=1}^{K}$. \\
$P_k$ & The $k$-th QA template. \\
$C_k$ & Template context set (selected events) for template $P_k$. \\
$R_k=(x_k,r_k,y_k)$ & Verified clinical relation in template $P_k$. \\
$x_k$ & Subject entity of relation $R_k$. \\
$r_k$ & Relation predicate of $R_k$. \\
$y_k$ & Object entity of relation $R_k$. \\
$A_k$ & Latent attributes associated with template $P_k$ (definitions/evidence/distractors/topic). \\
$f$ & QA instantiation function mapping templates $\mathcal{P}$ to QA items $\mathcal{I}$. \\
$\mathcal{I}$ & Set of constructed QA items. \\
$J$ & Number of QA items in $\mathcal{I}=\{(S_j,Q_j,B_j)\}_{j=1}^{J}$. \\
$S_j$ & Scenario (natural-language background) of QA item $j$. \\
$Q_j$ & Question text of QA item $j$. \\
$B_j$ & Metadata bundle of QA item $j$ (choices/answer/rationale/topic/relations). \\
$\mathcal{D}^{(n)}$ & Set of diagnoses observed in encounter $n$. \\
$\mathcal{T}^{(n)}$ & Set of treatments (prescriptions/procedures) observed in encounter $n$. \\
$\mathcal{S}^{(n)}$ & Scenario event subset used to form the QA context for encounter $n$. \\
$d^{(n)}_{\mathrm{tgt}}$ & Target (withheld) diagnosis in encounter $n$ for diagnosis decision. \\
$t^{(n)}_{\mathrm{tgt}}$ & Target treatment in encounter $n$ for treatment decision. \\
$d^{(n+1)}_{\mathrm{tgt}}$ & Target diagnosis in encounter $n{+}1$ for prognosis decision. \\
$p(\cdot\mid\mathcal{S}^{(n)})$ & Conditional distribution of the target event given scenario $\mathcal{S}^{(n)}$. \\
$|C_k|$ & Cardinality of the context set $C_k$. \\
\textit{CAUSE\_REL} & Positive SemMedDB predicate group used to verify ``cause'' relations. \\
\textit{AFFECT\_REL} & Positive SemMedDB predicate group used to verify ``affect'' relations. \\
\textit{ASSOC\_REL} & Positive SemMedDB predicate group used to verify ``associate-with'' relations. \\
\textit{USAGE\_REL} & Positive SemMedDB predicate group used to verify treatment-usage relations. \\
\textit{NEG\_CAUSE\_REL} & Negative SemMedDB predicate group used for contradiction checks of ``cause'' relations. \\
\textit{NEG\_AFFECT\_REL} & Negative SemMedDB predicate group used for contradiction checks of ``affect'' relations. \\
\textit{NEG\_ASSOC\_REL} & Negative SemMedDB predicate group used for contradiction checks of ``associate-with'' relations. \\
\textit{NEG\_USAGE\_REL} & Negative SemMedDB predicate group used for contradiction checks of treatment-usage relations. \\
\bottomrule
\end{tabular}
}
\end{table}

\newpage
\section{\modelname{} QA Statistics}
\label{sec:qa_statistics}

For \modelname{}, raw EHR records are drawn from three real-world sources: MIMIC-III, MIMIC-IV, and PROMOTE. MIMIC-III (Version 1.4) is a publicly available critical-care dataset containing 38{,}597 distinct patients and 53{,}423 hospital admissions from intensive care units at Beth Israel Deaconess Medical Center between 2001 and 2012 \cite{johnson2016mimic-iii}. MIMIC-IV (Version 3.1) is a newer release that extends MIMIC-III with updated hospital data from the same institution (2008--2022), including 364{,}627 patients and 546{,}028 hospital admissions \cite{johnson2023mimic-iv}. To further reduce potential contamination from public corpora and to evaluate language models in a setting less prone to data leakage, PROMOTE is additionally included as a private dataset from Emory Healthcare spanning 2012--2021, with records for 18{,}561 patients and 912{,}706 clinical records \cite{xie2025promote1,wu2025promote2}.

To ensure that each constructed QA item is grounded in sufficiently informative clinical documentation, we retain only encounters with rich structured signals. Concretely, an encounter is included only if it contains at least five diagnosis events and at least three treatment events (where treatments include both prescriptions and procedures). This filtering reduces the risk of constructing questions from sparse or under-specified visits and improves the reliability of the downstream relation extraction and QA instantiation. 

For each data source and task, up to the first 10{,}000 EHR instances are extracted and processed by the construction pipeline. End-to-end construction requires approximately 560 H100 GPU-hours. Across all sources and tasks, 465{,}748 candidate clinical relations are first extracted from EHR trajectories, and 62{,}786 knowledge-base-verified QA templates are retained (13.5\% retention). These templates are then instantiated into 960{,}067 QA items spanning three decision-making tasks---diagnosis (Dx), treatment (Tx), and prognosis (Px)---and four question formats: 4/5/6-choice multiple-choice questions (MCQs) and open-ended questions (OEQs). Table~\ref{tab:qa_stats} summarizes the breakdown by data source and task. Treatment accounts for the largest share of questions (450{,}501), followed by prognosis (259{,}123) and diagnosis (250{,}443). Aggregated by data source, \modelname{} contains 323{,}193 questions from MIMIC-III, 259{,}089 from MIMIC-IV, and 377{,}785 from PROMOTE, totaling \textbf{960{,}067} questions.

\begin{table*}[htbp]
\centering
\caption{\modelname{} construction statistics by task, data source, and question format. ``Candidate Relations'' counts relation candidates extracted from EHR trajectories before knowledge-base (KB) verification, and ``Verified Templates'' counts KB-supported templates retained for instantiation. ``MCQ-4/5/6'' report the numbers of 4/5/6-choice MCQs, and ``OEQ'' reports the number of open-ended questions.}
\label{tab:qa_stats}
\resizebox{0.95\linewidth}{!}{
\begin{tabular}{ccccccccc}
\toprule
\textbf{Task} & \textbf{Data Source} & \textbf{Candidate Relations} & \textbf{Verified Templates} & \textbf{MCQ-4} & \textbf{MCQ-5} & \textbf{MCQ-6} & \textbf{OEQ} & \textbf{Total} \\
\midrule
\multirow{4}{*}{\textbf{Diagnosis}}
& MIMIC-III & 32{,}338 & 4{,}406 & 17{,}624 & 21{,}315 & 24{,}264 & 4{,}329 & 67{,}532 \\
& MIMIC-IV  & 36{,}869 & 4{,}323 & 17{,}288 & 21{,}085 & 24{,}438 & 4{,}291 & 67{,}102 \\
& PROMOTE   & 56{,}699 & 7{,}259 & 29{,}024 & 36{,}250 & 43{,}278 & 7{,}257 & 115{,}809 \\
& Task-Total & 125{,}906 & 15{,}988 & 63{,}936 & 78{,}650 & 91{,}980 & 15{,}877 & 250{,}443 \\
\midrule
\multirow{4}{*}{\textbf{Treatment}}
& MIMIC-III & 53{,}637 & 9{,}878 & 39{,}428 & 49{,}230 & 58{,}926 & 9{,}873 & 157{,}457 \\
& MIMIC-IV  & 38{,}140 & 8{,}178 & 32{,}672 & 40{,}775 & 48{,}762 & 8{,}170 & 130{,}379 \\
& PROMOTE   & 45{,}899 & 10{,}173 & 40{,}672 & 50{,}830 & 60{,}990 & 10{,}173 & 162{,}665 \\
& Task-Total & 137{,}676 & 28{,}229 & 112{,}772 & 140{,}835 & 168{,}678 & 28{,}216 & 450{,}501 \\
\midrule
\multirow{4}{*}{\textbf{Prognosis}}
& MIMIC-III & 71{,}896 & 7{,}207 & 28{,}784 & 30{,}820 & 31{,}404 & 7{,}196 & 98{,}204 \\
& MIMIC-IV  & 60{,}816 & 4{,}452 & 17{,}800 & 19{,}470 & 19{,}890 & 4{,}448 & 61{,}608 \\
& PROMOTE   & 69{,}454 & 6{,}910 & 27{,}632 & 31{,}325 & 33{,}444 & 6{,}910 & 99{,}311 \\
& Task-Total & 202{,}166 & 18{,}569 & 74{,}216 & 81{,}615 & 84{,}738 & 18{,}554 & 259{,}123 \\
\midrule
\textbf{Full} & \textbf{\modelname{}}
& 465{,}748 & 62{,}786 & 250{,}924 & 301{,}100 & 345{,}396 & 62{,}647 & \textbf{960{,}067} \\
\bottomrule
\end{tabular}
}
\end{table*}

From the length statistics, each question stem is relatively detailed, averaging 332.4 characters (about 46.8 words). The answer options are much more concise: each option averages 28.6 characters (about 3.5 words), indicating that choices are typically short concept-level phrases. The reason field is also substantial, averaging 251.1 characters (about 36.4 words), suggesting there is enough space for coherent justifications rather than single-sentence fragments.

\section{Details of Methodology}
\label{sec:appendix_method_details}

This appendix provides additional implementation details for the benchmark construction pipeline described in Section~\ref{sec:method}, including encounter filtering, relation/context constraints, concept linking via UMLS, evidence retrieval via SemMedDB, task-specific relation definitions, and distractor generation and verification.

\subsection{Clinical Relations and Context Extraction from EHR}
\label{sec:appendix_relation_extraction}

\paragraph{\textbf{Relation cap.}}
For each EHR instance, the prompting stage may yield multiple candidate relations. To control template complexity and limit noise from overly long candidate lists, we retain at most $15$ candidate relations per instance after deduplication.

\paragraph{\textbf{Context construction and size.}}
For each retained relation template $P_k$, we generate a small auxiliary context set $C_k$ consisting of exactly two events per instance, as described in Section~\ref{sec:template_generation}. During QA instantiation, each question scenario incorporates three events in total: the two context events plus the relation-subject entity, i.e.,
\begin{equation}
S_j \leftarrow C_k \cup \{x_k\}, \quad \text{with } |C_k|=2.
\end{equation}
This design yields concise, controlled scenarios while preserving grounding in observed encounter events.

\paragraph{\textbf{Information leakage prevention.}}
When extracting clinical relations and contextual events from raw EHR data, we enforce strict non-overlap constraints between context events and clinical relations to prevent information leakage, since they may be developed into question stems and choices. This design ensures that each generated question is clinically meaningful rather than surface-level pattern matching.

\paragraph{\textbf{Task-Specific Relation Definitions}}

Each template contains a verified relation $R_k=(x_k,r_k,y_k)$ and is associated with one of three clinical decision tasks. We define the permissible entity roles and relation label spaces as follows.

\textbf{(I) Diagnosis decision.}
For the diagnosis task, both $x_k$ and $y_k$ are diagnoses recorded in the \emph{same} encounter. LLMs will focus on relations reflecting shared pathophysiologic mechanisms, or common comorbidity patterns, supporting questions that ask which additional diagnosis is likely to be identified given other diagnoses already present in the encounter.
The relation label is constrained to the same set:
\[
r_k \in \{\textit{cause},\ \textit{affect},\ \textit{associate-with}\}.
\]

\textbf{(II) Treatment decision.}
For the treatment task, $x_k$ is a diagnosis from the encounter and $y_k$ is a treatment action (prescription or procedure) from the same encounter. The relation label is constrained to:
\[
r_k \in \{\textit{treat-with-drug},\ \textit{treat-with-procedure}\},
\]
with evidence matched through \textit{USAGE\_REL} predicates and entity typing determined by the event category (prescription vs.\ procedure) after preprocessing.

\textbf{(III) Prognosis decision.}
For the prognosis task, $x_k$ is a diagnosis or treatment from the \emph{prior} encounter, and $y_k$ is a diagnosis in the \emph{subsequent} encounter. LLMs will focus on direct complications, risk-modifying effects, or clinically plausible associations that link earlier conditions or interventions to later outcomes. The relation label is constrained to
\[
r_k \in \{\textit{cause},\ \textit{affect},\ \textit{associate-with}\},
\]
where these labels map to SemMedDB predicate groups via
\textit{CAUSE\_REL}, \textit{AFFECT\_REL}, and \textit{ASSOC\_REL} (and their negative counterparts for contradiction checks).

\subsection{KB Usage}
\label{sec:appendix_umls_linking}

\paragraph{\textbf{Concept linking.}}
Concept linking is performed using the UMLS RESTful web services (UMLS web search) provided by the UMLS API.\footnote{\href{https://documentation.uts.nlm.nih.gov/rest/home.html}{UMLS API Link}} The current UMLS web version at query time is used. Each entity mention extracted from EHRs or produced during template completion (e.g., relation entities and distractor candidates) is resolved to a UMLS Concept Unique Identifier (CUI) using UMLS search.

To improve mapping precision, the search space is constrained to source vocabularies when appropriate:
\begin{itemize}[itemsep=1pt, topsep=0pt, parsep=0pt, partopsep=0pt]
    \item For \textbf{prescriptions}, source vocabularies aligned with \textbf{DrugBank} are prioritized.
    \item For \textbf{procedures} and \textbf{diagnoses}, source vocabularies aligned with \textbf{ICD} are prioritized.
\end{itemize}
The resolved CUIs serve as canonical identifiers for downstream KB retrieval and predicate matching in SemMedDB.

\paragraph{\textbf{SemMedDB Dataset.}}
The SemMedDB dataset is downloaded\footnote{\href{https://lhncbc.nlm.nih.gov/temp/SemRep_SemMedDB_SKR/SemMedDB_download.html}{SemMedDB Link}}, and the \textit{PREDICATION} and \textit{SENTENCE} files are used for relation verification and enrichment. In the \textit{PREDICATION} file, duplicate records are removed based on the \textit{SUBJECT}, \textit{OBJECT}, and \textit{PREDICATE} columns to obtain unique KB relations. Predicates are then grouped into positive and negative sets that correspond to the target labels for verification and filtering:
\begin{align*}
\textit{CAUSE\_REL} &= \left\{
\begin{aligned}[t]
&\textit{CAUSES},\ \textit{PRODUCES},\ \textit{CONVERTS\_TO}
\end{aligned}
\right\},\\[2pt]
\textit{AFFECT\_REL} &= \left\{
\begin{aligned}[t]
&\textit{PREDISPOSES},\ \textit{COMPLICATES},\ \textit{STIMULATES},\\
&\textit{AUGMENTS},\ \textit{AFFECTS}
\end{aligned}
\right\},\\[2pt]
\textit{ASSOC\_REL} &= \left\{
\begin{aligned}[t]
&\textit{ASSOCIATED\_WITH},\ \textit{COEXISTS\_WITH},\ \textit{INTERACTS\_WITH}
\end{aligned}
\right\},\\[2pt]
\textit{USAGE\_REL} &= \left\{
\begin{aligned}[t]
&\textit{TREATS},\ \textit{DIAGNOSES},\ \textit{ADMINISTERED\_TO},\ \textit{USED\_FOR},\\
&\textit{MEASUREMENT\_OF},\ \textit{METHOD\_OF},\ \textit{PREVENTS},\ \textit{INHIBITS},\ \textit{DISRUPTS}
\end{aligned}
\right\},
\end{align*}
and their corresponding negative forms:
\begin{align*}
\textit{NEG\_CAUSE\_REL} &= \left\{
\begin{aligned}[t]
&\textit{NEG\_CAUSES},\ \textit{NEG\_PRODUCES},\ \textit{NEG\_CONVERTS\_TO}
\end{aligned}
\right\},\\[2pt]
\textit{NEG\_AFFECT\_REL} &= \left\{
\begin{aligned}[t]
&\textit{NEG\_PREDISPOSES},\ \textit{NEG\_COMPLICATES},\ \textit{NEG\_STIMULATES},\\
&\textit{NEG\_AUGMENTS},\ \textit{NEG\_AFFECTS}
\end{aligned}
\right\},\\[2pt]
\textit{NEG\_ASSOC\_REL} &= \left\{
\begin{aligned}[t]
&\textit{NEG\_ASSOCIATED\_WITH},\ \textit{NEG\_COEXISTS\_WITH},\ \textit{NEG\_INTERACTS\_WITH}
\end{aligned}
\right\},\\[2pt]
\textit{NEG\_USAGE\_REL} &= \left\{
\begin{aligned}[t]
&\textit{NEG\_TREATS},\ \textit{NEG\_DIAGNOSES},\ \textit{NEG\_ADMINISTERED\_TO},\ \textit{NEG\_USED\_FOR},\\
&\textit{NEG\_MEASUREMENT\_OF},\ \textit{NEG\_METHOD\_OF},\ \textit{NEG\_PREVENTS},\ \textit{NEG\_INHIBITS},\ \textit{NEG\_DISRUPTS}
\end{aligned}
\right\}.
\end{align*}
After evidence is identified, the corresponding \textit{SENTENCE\_ID} is used to retrieve supporting PubMed sentences from the \textit{SENTENCE} file.

\subsection{QA Template Completion}
\label{sec:qa_templates}

\paragraph{\textbf{Distractor generation.}}
For each verified relation template, an LLM generates $10$ distractor candidates based on internal knowledge and structured template attributes. In addition, observed clinical events are sampled to expand the candidate pool, yielding up to $25$ total distractor candidates per template.

\paragraph{\textbf{Topic generation.}}
For each relation template, a high-level clinical condition topic is generated to summarize the clinical focus of the relation. For diagnosis and prognosis tasks, the topic is derived from $y_k$ (the object entity). For treatment tasks, the topic is derived from $x_k$ (the subject entity), because the object entity is a treatment rather than a diagnosis. Statistics of the generated topics are reported in Appendix~\ref{sec:topic_analysis}.

\paragraph{\textbf{Rationale generation.}}
The LLM is prompted to review the extracted clinical relations and the retrieved KB evidence (e.g., entity definitions, positive evidence, and supporting sentences) to generate a concise clinical rationale. The resulting rationales are stored in the templates and can be directly used as explanations for the MCQs and OEQs derived from the corresponding templates.

\paragraph{\textbf{Template diversity constraint per patient.}}
To avoid over-representing a single patient trajectory and to promote diversity across conditions and care patterns, the number of verified templates per patient is restricted. Each patient contributes at most three templates across all eligible encounters, selected to maximize coverage of distinct relations and reduce redundancy.

\paragraph{\textbf{Minimum distractor threshold.}}
To ensure that MCQs include meaningful alternatives and reduce ambiguity, templates with fewer than three verified distractors after filtering are discarded. As a result, each instantiated MCQ contains 4, 5, or 6 choices. This constraint is applied before the final QA instantiation stage.

\subsection{QA Generation}
\label{sec:qa_generation}

\paragraph{\textbf{Question Skeleton}}
\begin{itemize}[itemsep=1pt, topsep=0pt, parsep=0pt, partopsep=0pt]
    \item Diagnosis MCQ: Based on the clinical context summarized above, which additional diagnosis is most likely to be present or identified during this visit?
    \item Treatment MCQ: Given the clinical context summarized above, which treatment is most likely to be prescribed during this visit?
    \item Prognosis MCQ: Given the prior clinical history summarized above, which diagnosis is most likely to be present or identified at the next visit?
    \item Diagnosis OEQ: Given the diagnoses of the patient, what may lead to this target diagnosis during the same visit? Why?
    \item Treatment OEQ: Given the diagnoses of the patient, what may lead to this target treatment during the same visit? Why?
    \item Prognosis OEQ: Given the prior visit of the patient, what may lead to this target diagnosis at the next visit? Why?
\end{itemize}

\paragraph{\textbf{Question Paraphrasing}}
To improve robustness and linguistic diversity, an instruction-following LLM is used to generate $V$ paraphrased versions of the same scenario while preserving clinical intent. Concretely, $V{=}6$ paraphrased scenarios $\{S^{(v)}_j\}_{v=1}^{V}$ are generated, and each scenario is paired with a fixed ask-only question to form the corresponding question instances. Prompts are designed to produce surface-level paraphrases of the same clinical context by rephrasing wording and sentence structure without introducing new information, clinical interpretation, or emphasis across events. Furthermore, the LLM is prohibited from using abstraction or causal language, and similar length and structure are maintained across versions to ensure that variation is limited to linguistic form rather than semantic content. This controlled paraphrasing strategy enables robust evaluation under deterministic rewording while preventing information leakage and preserving clinical meaning.

\paragraph{\textbf{Choice Permutation}}
To mitigate sensitivity to answer-position bias, $c$ permutation variants are generated for each instance using a fixed random seed to ensure reproducibility. For a $c$-choice MCQ, each variant permutes the order of the $c$ answer options while updating the ground-truth label to match the new position of the correct answer. As a result, across the $c$ variants, each option appears exactly once in each position, eliminating systematic advantages associated with any specific answer index. This design prevents models from exploiting positional heuristics, enforces position-invariant evaluation, and enables a more faithful assessment of model reasoning ability rather than sensitivity to answer ordering.

\subsection{Trade-off in Knowledge-base Verification}
\label{sec:kb_tradeoff}

The benchmark construction pipeline adopts a precision-first verification strategy: uncertain or weakly supported relations are discarded rather than retained. This design choice is intended to improve the reliability and clinical validity of retained QA items at scale.
Among candidate relations not retained, approximately 81\% lacked supporting KB evidence, 12\% were associated with explicit negative evidence, 5\% were removed by the per-patient diversity cap, and 2\% failed distractor-quality filtering. These statistics indicate that most rejected candidates arise from insufficient supporting evidence rather than direct contradiction.

To estimate potential false negatives introduced by the conservative filtering strategy, we manually audited 225 rejected candidate relations (25 sampled items across 3 tasks and 3 data sources). Approximately 24\% were judged clinically plausible despite not being retained. These misses likely arise from incomplete KB coverage for medically reasonable relations, near-synonymous CUI granularity mismatch, and entity-linking loss introduced during API-based normalization.

This behavior reflects the intended trade-off of the current pipeline. \modelname{} prioritizes higher precision and stronger grounding of retained relations over maximizing recall, acknowledging that some clinically plausible but weakly supported relations may be excluded during verification.

\subsection{Quality Control}
\label{sec:quality_control}

LLM-based generation enables scalable and automated benchmark construction; however, reliability in the clinical domain requires systematic safeguards. We therefore implement multi-stage quality control throughout the construction of \modelname{} to improve correctness, clarity, and fairness. These safeguards are tightly integrated into the \emph{EHR--LLM--KB} pipeline and operate at multiple stages of template generation, instantiation, and validation. The quality control pipeline consists of the following components:

\begin{itemize}[itemsep=1pt, topsep=0pt, parsep=0pt, partopsep=0pt]

\item \textbf{Terminology normalization.}
Raw EHR events are mapped to normalized, human-readable descriptions and standardized biomedical codes (UMLS CUIs). This step bridges heterogeneity across EHR systems and external knowledge bases, reducing spurious variation caused by synonyms, abbreviations, and data-source-specific naming conventions. Normalization ensures consistent grounding of clinical concepts across data sources and enables reliable downstream KB verification.

\item \textbf{Knowledge-base verification of clinical relations.}
To reduce hallucinations and improve clinical validity, all LLM-extracted relations are verified and enriched using external biomedical knowledge bases, including UMLS, SemMedDB, PubMed, ICD, and DrugBank. Relations are retained only if they are supported by positive evidence and are not contradicted by known negative or conflicting relations. Unsupported or contradictory relations are discarded during Stage~2 of template generation, ensuring that retained relations reflect clinically plausible associations grounded in biomedical knowledge.

\item \textbf{Knowledge-base filtering of templates.}
To preserve an unambiguous choice structure in multiple-choice questions, candidate distractors are further pruned through KB-based filtering. Clinically plausible alternatives that are also supported by knowledge bases given the anchor concept and auxiliary context are removed, reducing the risk of multiple correct answers. This filtering step, applied in Stage~4 of template generation, enforces single-answer correctness while maintaining clinical realism.

\item \textbf{Leakage prevention.}
To minimize information leakage from the context to the answer options, we enforce forbidden-term constraints and non-overlap rules between question stems and choices. In particular, extracted clinical relations are required not to trivially overlap with events explicitly mentioned in the EHR context. These constraints prevent models from exploiting surface cues or lexical overlap instead of performing genuine clinical reasoning.

\item \textbf{Format and structural validation.}
All LLM outputs are required to conform to a predefined JSON schema and are parsed using robust validation routines. At the QA-item level, each record is checked to ensure that required fields are present and non-empty, that option sets are well-formed, and that duplicate or malformed items are removed. This validation step guarantees structural consistency across the entire benchmark.

\item \textbf{Diversity and fairness controls.}
\modelname{} is constructed at large scale, spanning multiple clinical decision tasks, sources, and question types, resulting in a total of 960{,}067 QA items. To mitigate evaluation bias, multiple deterministic variants are generated through paraphrasing and answer-option permutations, ensuring that model performance reflects robust reasoning rather than sensitivity to surface form. This design promotes fair and stable comparison across models.

\end{itemize}

Together, these safeguards ensure that the generated QA items fully leverage the scalability of LLM-based generation and the richness of real-world EHR data, while maintaining high standards of clinical correctness, structural clarity, and evaluation fairness.


\subsection{\modelname{} Usage Protocol}
\label{sec:evalution_protocol}

The generated QA items are used to evaluate the CDM capability of LLMs. For each QA item $I_j=(S_j,Q_j,B_j)$, the model receives only the natural-language input $(S_j,Q_j)$. All metadata $B_j$, including the answer, choices, evidence, and clinical relations, are withheld from the model and used exclusively for scoring and post-hoc analysis. The model is required to produce an answer in the specified format for the corresponding question type.

Because the benchmark contains parallel variants and multiple question types, flexible evaluation settings are supported. For efficiency and broader coverage of clinical concepts, evaluation can be conducted on a fixed subset of variants, such as a single version of the 4-choice questions (e.g., Version~1), which provides a controlled and non-redundant evaluation set. For robustness, evaluation can be conducted across multiple variants and aggregated, for example, by combining all four variants of the 4-choice MCQs, all five variants of the 5-choice MCQs, and all six variants of the 6-choice MCQs, thereby reducing sensitivity to wording and answer-position effects. For MCQs, \emph{accuracy (ACC)} is used as the primary metric. For OEQs, metrics include \emph{Coverage} (whether the answer covers the target clinical relation), \emph{ROUGE} (comparison with the reference rationale), and \emph{BERTScore}. 

\section{Topic Analysis}
\label{sec:topic_analysis}

We also extract question ``topics'' and analyze their distribution. Topics are generated during the template generation step (see Section~\ref{sec:template_generation} and Appendix~\ref{sec:qa_templates}). Each topic represents the primary clinical condition that a question targets, i.e., a ``diagnosis'' in structured EHR data. After extracting topics from \modelname{}, each topic is linked to an ICD-10-CM code, truncated to the 3-digit level (three characters), and used to aggregate the number of questions (Count) and data source share (Percent) together with ICD-10-CM category descriptions. In total, 3{,}808 unique ICD codes are mapped, and 752 unique codes remain after 3-digit truncation, indicating broad coverage across ICD-10-CM categories and diverse clinical relations in \modelname{}. Table~\ref{tab:top20_icd3_topics} summarizes the top-20 ICD-10-CM 3-character categories in the full data source.

\begin{table*}[htbp]
\centering
\caption{Top-20 most frequent ICD-10-CM categories in \modelname{}. We report ICD-10 codes truncated to three characters, along with the total number of questions (Count), the corresponding data source share (Percent), and code descriptions.}
\label{tab:top20_icd3_topics}
\resizebox{0.7\linewidth}{!}{
\begin{tabular}{cccl}
\toprule
\textbf{ICD} &\textbf{ Count} & \textbf{Percent (\%) }& \textbf{Description} \\
\midrule
I50 & 67486 & 7.11 & Heart failure \\
E87 & 58869 & 6.20 & Other disorders of fluid, electrolyte and acid-base balance \\
K59 & 49915 & 5.26 & Other functional disorders of intestine \\
I10 & 43732 & 4.61 & Essential (primary) hypertension \\
N17 & 30485 & 3.21 & Kidney failure, acute \\
I48 & 26060 & 2.75 & Atrial fibrillation and flutter \\
I49 & 22835 & 2.41 & Other cardiac arrhythmias \\
J96 & 20986 & 2.21 & Respiratory failure, not elsewhere classified \\
N18 & 20966 & 2.21 & Chronic kidney diseases \\
E86 & 18484 & 1.95 & Volume depletion \\
F41 & 18013 & 1.90 & Other anxiety disorders \\
F32 & 17910 & 1.89 & Depressive episode \\
E78 & 16597 & 1.75 & Disorder of lipoprotein metabolism and other lipidemias \\
I20 & 13811 & 1.45 & Angina pectoris \\
E03 & 12535 & 1.32 & Other hypothyroidism \\
K76 & 11721 & 1.23 & Other diseases of liver \\
T81 & 11167 & 1.18 & Complications of procedures, not elsewhere classified \\
G89 & 11001 & 1.16 & Pain, not elsewhere classified \\
J44 & 10854 & 1.14 & Other chronic obstructive pulmonary disease \\
D64 & 10310 & 1.09 & Other anemia \\
\bottomrule
\end{tabular}
}
\end{table*}

From Table~\ref{tab:top20_icd3_topics}, the most frequent ICD-10-CM 3-character categories correspond to clinically common conditions and acute decompensation syndromes that routinely drive EHR-based decision making, which supports that the \modelname{} construction pipeline yields clinically sensible topics and code mappings. In particular, the head of the distribution is dominated by high-prevalence cardiometabolic and acute-care presentations, such as heart failure (I50, 7.11\%), fluid/electrolyte and acid--base disorders (E87, 6.20\%), functional intestinal disorders (K59, 5.26\%), and essential (primary) hypertension (I10, 4.61\%). Failure-state phenotypes are also prominent, including acute kidney failure (N17, 3.21\%), chronic kidney disease (N18, 2.21\%), and respiratory failure (J96, 2.21\%), consistent with the fact that many inpatient encounters center on physiologic instability and organ dysfunction. Meanwhile, the presence of comorbidity- and symptom-burden categories---such as volume depletion (E86, 1.95\%), pain (G89, 1.16\%), anemia (D64, 1.09\%), and mental health conditions (F41/F32, 1.90\%/1.89\%)---suggests that the extracted topics cover both primary diagnoses and common accompanying problems reflected in longitudinal records. At the same time, the distribution exhibits a long-tail pattern: the top-20 ICD-10-CM 3-character categories cover 51\% of all questions, implying that nearly half of the benchmark is distributed across a wide range of lower-frequency categories. This long tail is desirable for evaluation because it reduces over-reliance on a small set of frequent conditions and encourages models to generalize beyond the most common clinical scenarios.

\begin{table*}[htbp]
\centering
\caption{Top-20 most frequent ICD-10-CM categories in \modelname{} (from MIMIC-III). We report ICD-10 codes truncated to three characters, along with the total number of questions (Count), the corresponding data source share (Percent), and code descriptions.}
\label{tab:top20_icd3_mimic3}
\resizebox{0.7\linewidth}{!}{
\begin{tabular}{cccl}
\toprule
\textbf{ICD} &\textbf{ Count} & \textbf{Percent (\%) }& \textbf{Description} \\
\midrule
I50 & 31464 & 9.85 & Heart failure \\
E87 & 25992 & 8.14 & Other disorders of fluid, electrolyte and acid-base balance \\
J96 & 15236 & 4.77 & Respiratory failure, not elsewhere classified \\
N17 & 13565 & 4.25 & Kidney failure, acute \\
I48 & 10837 & 3.39 & Atrial fibrillation and flutter \\
K59 & 10459 & 3.27 & Other functional disorders of intestine \\
T81 & 8989 & 2.81 & Complications of procedures, not elsewhere classified \\
I10 & 7967 & 2.49 & Essential (primary) hypertension \\
N18 & 6926 & 2.17 & Chronic kidney diseases \\
E86 & 5857 & 1.83 & Volume depletion \\
I49 & 5320 & 1.67 & Other cardiac arrhythmias \\
K76 & 4699 & 1.47 & Other diseases of liver \\
G40 & 4696 & 1.47 & Epilepsy and recurrent seizures \\
I20 & 4672 & 1.46 & Angina pectoris \\
E83 & 4612 & 1.44 & Disorder of mineral metabolism \\
E03 & 4145 & 1.30 & Other hypothyroidism \\
A41 & 4008 & 1.25 & Other sepsis \\
E08--E13 & 3761 & 1.18 & Diabetes mellitus \\
D64 & 3716 & 1.16 & Other anemia \\
J44 & 3672 & 1.15 & Other chronic obstructive pulmonary disease \\
\bottomrule
\end{tabular}
}
\end{table*}

\begin{table*}[htbp]
\centering
\caption{Top-20 most frequent ICD-10-CM categories in \modelname{} (from MIMIC-IV). We report ICD-10 codes truncated to three characters, along with the total number of questions (Count), the corresponding data source share (Percent), and code descriptions.}
\label{tab:top20_icd3_mimic4}
\resizebox{0.7\linewidth}{!}{
\begin{tabular}{cccl}
\toprule
\textbf{ICD} &\textbf{ Count} & \textbf{Percent (\%) }& \textbf{Description} \\
\midrule
K59 & 17895 & 6.97 & Other functional disorders of intestine \\
E87 & 13242 & 5.16 & Other disorders of fluid, electrolyte and acid-base balance \\
I10 & 10482 & 4.08 & Essential (primary) hypertension \\
E86 & 10210 & 3.98 & Volume depletion \\
I50 & 9186 & 3.58 & Heart failure \\
N17 & 8837 & 3.44 & Kidney failure, acute \\
F32 & 8666 & 3.37 & Depressive episode \\
F41 & 8533 & 3.32 & Other anxiety disorders \\
E78 & 6624 & 2.58 & Disorder of lipoprotein metabolism and other lipidemias \\
E03 & 6321 & 2.46 & Other hypothyroidism \\
N18 & 5754 & 2.24 & Chronic kidney diseases \\
I49 & 5152 & 2.01 & Other cardiac arrhythmias \\
G89 & 4935 & 1.92 & Pain, not elsewhere classified \\
I48 & 4858 & 1.89 & Atrial fibrillation and flutter \\
K76 & 4246 & 1.65 & Other diseases of liver \\
J44 & 3996 & 1.56 & Other chronic obstructive pulmonary disease \\
D64 & 3850 & 1.50 & Other anemia \\
R11 & 3290 & 1.28 & Nausea and vomiting \\
I20 & 2712 & 1.06 & Angina pectoris \\
R10 & 2244 & 0.87 & Abdominal and pelvic pain \\
\bottomrule
\end{tabular}
}
\end{table*}

\begin{table*}[htbp]
\centering
\caption{Top-20 most frequent ICD-10-CM categories in \modelname{} (from PROMOTE). We report ICD-10 codes truncated to three characters, along with the total number of questions (Count), the corresponding data source share (Percent), and code descriptions.}
\label{tab:top20_icd3_promote}
\resizebox{0.7\linewidth}{!}{
\begin{tabular}{cccl}
\toprule
\textbf{ICD} &\textbf{ Count} & \textbf{Percent (\%) }& \textbf{Description} \\
\midrule
I50 & 26836 & 7.20 & Heart failure \\
I10 & 25283 & 6.79 & Essential (primary) hypertension \\
K59 & 21561 & 5.79 & Other functional disorders of intestine \\
E87 & 19635 & 5.27 & Other disorders of fluid, electrolyte and acid-base balance \\
I49 & 12363 & 3.32 & Other cardiac arrhythmias \\
I48 & 10365 & 2.78 & Atrial fibrillation and flutter \\
N18 & 8286 & 2.22 & Chronic kidney diseases \\
N17 & 8083 & 2.17 & Kidney failure, acute \\
E78 & 8005 & 2.15 & Disorder of lipoprotein metabolism and other lipidemias \\
F32 & 6469 & 1.74 & Depressive episode \\
I20 & 6427 & 1.72 & Angina pectoris \\
F41 & 6275 & 1.68 & Other anxiety disorders \\
M62 & 5653 & 1.52 & Other disorders of muscle \\
I63 & 5409 & 1.45 & Cerebral infarction \\
I95 & 5392 & 1.45 & Hypotension \\
E55 & 5253 & 1.41 & Vitamin D deficiency \\
R11 & 4676 & 1.25 & Nausea and vomiting \\
I51 & 4321 & 1.16 & Complications and ill-defined descriptions of heart disease \\
T87 & 3985 & 1.07 & Complications peculiar to reattachment and amputation \\
K21 & 3764 & 1.01 & Gastro-esophageal reflux disease \\
\bottomrule
\end{tabular}
}
\end{table*}

We also analyze question topics from each data source (MIMIC-III, MIMIC-IV, and PROMOTE). Tables~\ref{tab:top20_icd3_mimic3}--\ref{tab:top20_icd3_promote} report the top-20 ICD-10-CM categories (truncated to the ICD-10-CM 3-character level) within each data source, including aggregated counts, data source shares, and code descriptions. Across all three sources, the head of the distribution is broadly consistent with the full-dataset pattern: common cardiometabolic and decompensation-related categories repeatedly appear among the most frequent topics, such as heart failure (I50), essential (primary) hypertension (I10), electrolyte and acid--base disorders (E87), functional intestinal disorders (K59), acute and chronic kidney disease (N17/N18), atrial fibrillation and related arrhythmias (I48/I49), and angina pectoris (I20). This cross-source agreement suggests that the extracted topics capture clinically prevalent conditions that are routinely documented in structured EHRs, and further supports that the \modelname{} topic extraction and ICD mapping procedures are clinically sensible rather than being driven by data source-specific artifacts. At the same time, each data source exhibits distinct secondary emphases consistent with its underlying cohort and documentation patterns, including higher prevalence of respiratory failure (J96) and sepsis (A41) in MIMIC-III, symptom-oriented categories such as nausea/vomiting (R11) and abdominal pain (R10) in MIMIC-IV, and additional cardiovascular and peri-procedural complications (e.g., I63, I95, I51, T87) in PROMOTE.

\begin{table*}[htbp]
\centering
\caption{Top-20 most frequent ICD-10-CM categories in \modelname{} (from the diagnosis subset). We report ICD-10 codes truncated to three characters, along with the total number of questions (Count), the corresponding data source share (Percent), and code descriptions.}
\label{tab:top20_icd3_dx}
\resizebox{0.9\linewidth}{!}{
\begin{tabular}{cccl}
\toprule
\textbf{ICD} & \textbf{Count} & \textbf{Percent (\%)} & \textbf{Description} \\
\midrule
I50 & 19914 & 8.08 & Heart failure \\
E87 & 13627 & 5.53 & Other disorders of fluid, electrolyte and acid-base balance \\
I49 & 13062 & 5.30 & Other cardiac arrhythmias \\
N17 & 9880 & 4.01 & Kidney failure, acute \\
I10 & 8584 & 3.48 & Essential (primary) hypertension \\
I48 & 8165 & 3.31 & Atrial fibrillation and flutter \\
J96 & 6427 & 2.61 & Respiratory failure, not elsewhere classified \\
N18 & 6284 & 2.55 & Chronic kidney diseases \\
E78 & 5632 & 2.29 & Disorder of lipoprotein metabolism and other lipidemias \\
D64 & 5449 & 2.21 & Other anemia \\
F41 & 5321 & 2.16 & Other anxiety disorders \\
I51 & 4554 & 1.85 & Complications and ill-defined descriptions of heart disease \\
M62 & 3869 & 1.57 & Other disorders of muscle \\
E86 & 3620 & 1.47 & Volume depletion \\
R53 & 3485 & 1.41 & Malaise and fatigue \\
G47 & 3155 & 1.28 & Sleep disorders \\
T81 & 2856 & 1.16 & Complications of procedures, not elsewhere classified \\
J81 & 2848 & 1.16 & Pulmonary edema \\
K76 & 2781 & 1.13 & Other diseases of liver \\
R65 & 2701 & 1.10 & Symptoms and signs specifically associated with systemic inflammation and infection \\
\bottomrule
\end{tabular}
}
\end{table*}

\begin{table*}[htbp]
\centering
\caption{Top-20 most frequent ICD-10-CM categories in \modelname{} (from the treatment subset). We report ICD-10 codes truncated to three characters, along with the total number of questions (Count), the corresponding data source share (Percent), and code descriptions.}
\label{tab:top20_icd3_tx}
\resizebox{0.7\linewidth}{!}{
\begin{tabular}{cccl}
\toprule
\textbf{ICD} &\textbf{ Count} & \textbf{Percent (\%) }& \textbf{Description} \\
\midrule
K59 & 48763 & 10.91 & Other functional disorders of intestine \\
E87 & 36589 & 8.19 & Other disorders of fluid, electrolyte and acid-base balance \\
I10 & 32175 & 7.20 & Essential (primary) hypertension \\
I50 & 29370 & 6.57 & Heart failure \\
F32 & 13329 & 2.98 & Depressive episode \\
E86 & 12054 & 2.70 & Volume depletion \\
I20 & 11336 & 2.54 & Angina pectoris \\
I48 & 9717 & 2.17 & Atrial fibrillation and flutter \\
G89 & 9674 & 2.17 & Pain, not elsewhere classified \\
F41 & 9236 & 2.07 & Other anxiety disorders \\
R11 & 9211 & 2.06 & Nausea and vomiting \\
E03 & 8380 & 1.88 & Other hypothyroidism \\
E78 & 8224 & 1.84 & Disorder of lipoprotein metabolism and other lipidemias \\
E08--E13 & 7921 & 1.77 & Diabetes mellitus \\
E83 & 7546 & 1.69 & Disorder of mineral metabolism \\
J44 & 6529 & 1.46 & Other chronic obstructive pulmonary disease \\
A49 & 6138 & 1.37 & Bacterial infection of unspecified site \\
G40 & 5760 & 1.29 & Epilepsy and recurrent seizures \\
K21 & 5573 & 1.25 & Gastro-esophageal reflux disease \\
N17 & 5131 & 1.15 & Kidney failure, acute \\
\bottomrule
\end{tabular}
}
\end{table*}

\begin{table*}[htbp]
\centering
\caption{Top-20 most frequent ICD-10-CM categories in \modelname{} (from the prognosis subset). We report ICD-10 codes truncated to three characters, along with the total number of questions (Count), the corresponding data source share (Percent), and code descriptions.}
\label{tab:top20_icd3_px}
\resizebox{0.7\linewidth}{!}{
\begin{tabular}{cccl}
\toprule
\textbf{ICD} &\textbf{ Count} & \textbf{Percent (\%) }& \textbf{Description} \\
\midrule
I50 & 18202 & 7.12 & Heart failure \\
N17 & 15474 & 6.05 & Kidney failure, acute \\
N18 & 11690 & 4.57 & Chronic kidney diseases \\
J96 & 9983 & 3.90 & Respiratory failure, not elsewhere classified \\
E87 & 8653 & 3.38 & Other disorders of fluid, electrolyte and acid-base balance \\
I48 & 8178 & 3.20 & Atrial fibrillation and flutter \\
I49 & 8115 & 3.17 & Other cardiac arrhythmias \\
T87 & 6089 & 2.38 & Complications peculiar to reattachment and amputation \\
T81 & 5175 & 2.02 & Complications of procedures, not elsewhere classified \\
I95 & 4597 & 1.80 & Hypotension \\
P39 & 4047 & 1.58 & Other infections specific to the perinatal period \\
K76 & 3938 & 1.54 & Other diseases of liver \\
N95 & 3879 & 1.52 & Menopausal and other perimenopausal disorders \\
F41 & 3456 & 1.35 & Other anxiety disorders \\
D64 & 3277 & 1.28 & Other anemia \\
I10 & 2973 & 1.16 & Essential (primary) hypertension \\
E86 & 2810 & 1.10 & Volume depletion \\
E78 & 2741 & 1.07 & Disorder of lipoprotein metabolism and other lipidemias \\
J44 & 2686 & 1.05 & Other chronic obstructive pulmonary disease \\
R00 & 2664 & 1.04 & Abnormal heart beat \\
\bottomrule
\end{tabular}
}
\end{table*}

We also analyze question topics separately for the three clinical decision tasks (diagnosis, treatment, and prognosis). Tables~\ref{tab:top20_icd3_dx}--\ref{tab:top20_icd3_px} summarize the top-20 ICD-10-CM categories (ICD-10-CM 3-character level) within each task subset. Overall, the three subsets share a consistent head dominated by common cardiometabolic and decompensation-related conditions, such as heart failure (I50), fluid/electrolyte and acid--base disorders (E87), essential (primary) hypertension (I10), atrial fibrillation and related arrhythmias (I48/I49), acute kidney failure (N17), chronic kidney disease (N18), and respiratory failure (J96). These patterns indicate that the extracted topics reflect clinically plausible task-specific distributions while remaining broadly consistent across tasks.

\newpage
\section{Extended Contents of the Main Experiment: Benchmarking LLMs on \modelname{} Across Clinical Decision Tasks}
\label{sec:extended_main_experiment}

\subsection{Implementation Details}
\label{sec:implementation_details}
When running experiments, we process questions in batches of ten using instruction prompts to improve efficiency and ensure stable outputs. Inputs are typically around 8{,}000 tokens, with a maximum context length of 10{,}240 tokens. To further reduce latency, we enable early stopping once the model produces a complete JSON-formatted answer. All experiments are implemented in Python~3.11.5 and executed on NVIDIA H200 GPUs with CUDA~12.4. For Azure OpenAI Service, we use API version \textit{``2025-03-01-preview''}. We adopt deterministic decoding to ensure reproducibility; for example, when running the open-source general LLM \textit{``llama3-8b''}, we set \textit{``do-sample=False, temperature=0, top-k=1''}. All models are evaluated using the same prompt template.

To ensure a fair comparison across tasks, sources, and choice types with potentially different numbers of generated QA instances, we construct a fixed evaluation subset for every setting: for each source-task-type combination, we take the first 3{,}000 questions from the first version. Therefore, each LLM is evaluated with 81{,}000 questions in total, which provides a very comprehensive evaluation.

\subsection{Cost Details}
\label{sec:cost_analysis}

We analyze the computational and monetary costs of the main experiment in Section~\ref{sec:main_results}. Table~\ref{tab:cost_efficiency_overview} reports total token usage, end-to-end runtime, API cost (when applicable), throughput defined as tokens per hour, and overall accuracy. Total token usage is tightly clustered around 10--11M tokens across most models, with only a few higher-token runs (e.g., mistral-7b at 12.87M and yi-1.5-34b at 13.04M), indicating that the main comparison is conducted under a largely matched token budget. In contrast, runtime varies widely from 1.01h (smollm3-3b) to 20.87h (llama3.3-70b), so throughput is largely time-determined in this setting. As a result, throughput spans an order of magnitude, from 0.50 M Tokens/h (llama3.3-70b) to 10.21 M Tokens/h (smollm3-3b), suggesting that system-level efficiency differences dominate over token-count differences under the same evaluation protocol. API-based models often achieve higher throughput than similarly sized self-hosted models, but this efficiency comes with non-negligible monetary cost.

The most accurate API-based models are also among the most expensive and slower in practice. For example, gpt-5.2 achieves the highest overall accuracy (70.91\%) but incurs the largest API cost (\$71.03) and runs at 1.00 M Tokens/h with a 12.17h runtime. The next tier, gpt-4.1 (69.43\%) and gpt-5 (69.06\%), reduces the cost relative to gpt-5.2 but still requires \$40.67--\$42.37, with 4.02--6.45h runtime and 1.57--2.52 M Tokens/h throughput. These results indicate a practical trade-off: improvements at the top end of accuracy can require disproportionately larger time and monetary budgets.

\begin{figure*}[htbp]
  \centering
  \includegraphics[width=\linewidth]{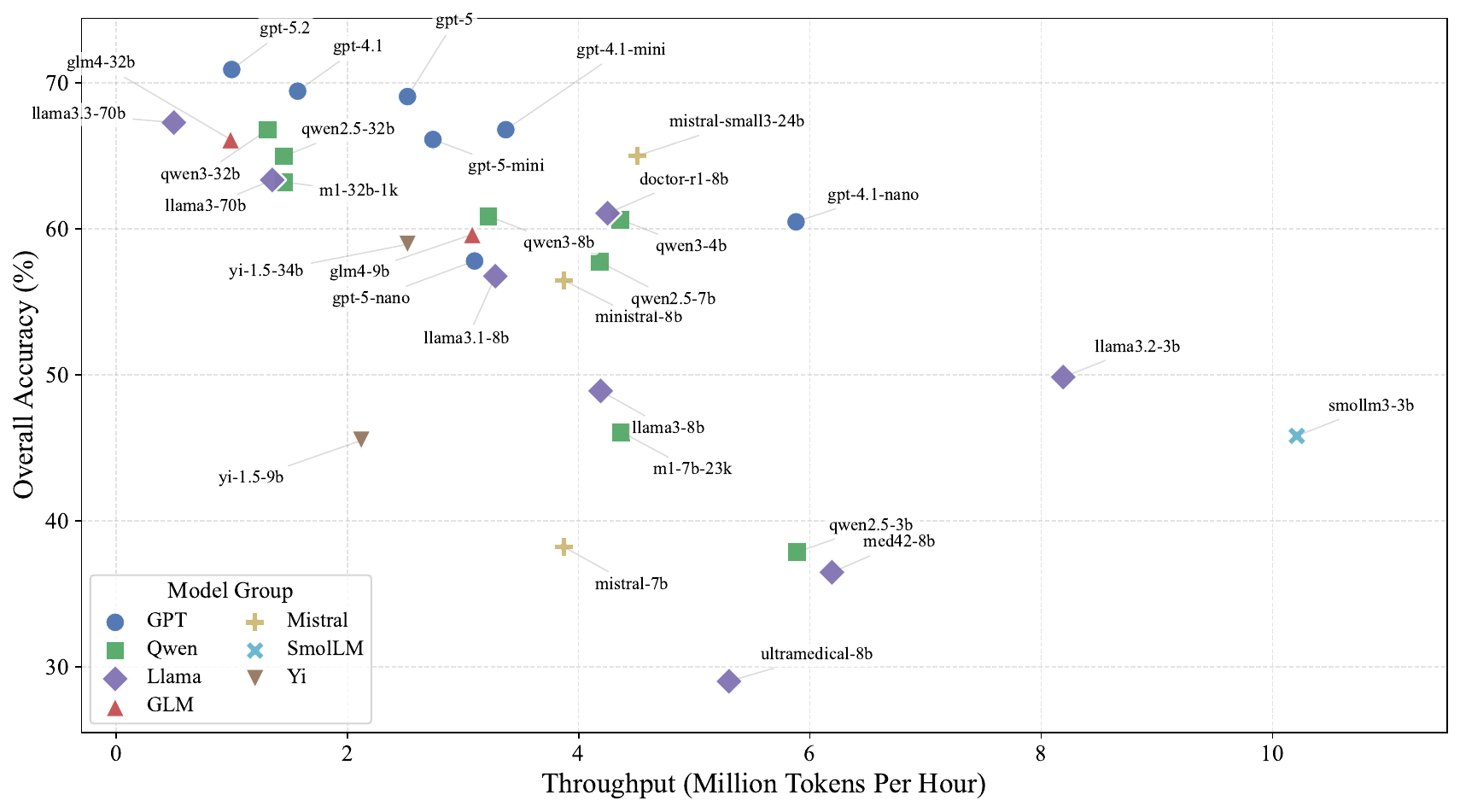}
  \caption{Overall accuracy versus throughput (M Tokens/h) for all evaluated models. Throughput is computed as total tokens divided by end-to-end runtime under the same evaluation protocol. For visualization, medical LLMs are grouped by the family of their corresponding base models: med42, ultramedical, and doctor-r1 are grouped under Llama, while m1 is grouped under Qwen.}
  \label{fig:cost}
\end{figure*}

To further illustrate this trade-off, Figure~\ref{fig:cost} plots overall accuracy against throughput. The highest accuracies concentrate at relatively low throughput, and the upper envelope is formed by comparatively slow models in this benchmark setup. Across all models, higher throughput does not imply higher accuracy, and the high-throughput region is primarily populated by weaker models, indicating that efficiency and effectiveness can be decoupled in practice. A mid-throughput band around 4--6 M Tokens/h provides a pragmatic operating point when both quality and runtime matter. For example, mistral-small3-24b reaches 65.01\% at 4.51 M Tokens/h, and doctor-r1-8b attains 61.07\% at 4.25 M Tokens/h. On the API side, gpt-4.1-nano runs at 5.88 M Tokens/h with 60.48\% accuracy, offering materially higher throughput than frontier models but with a noticeable accuracy gap to gpt-4.1 and gpt-5.2. Within model families, scaling typically increases accuracy while reducing throughput. In the Llama series, accuracy rises from 48.90\% (llama3-8b) to 63.35\% (llama3-70b) and 67.28\% (llama3.3-70b), while throughput drops from 4.19 to 1.35 and further to 0.50 M Tokens/h. A similar pattern appears in Qwen: qwen3-4b and qwen3-8b achieve 60.63\% and 60.87\% at 4.36 and 3.22 M Tokens/h, whereas qwen3-32b improves to 66.78\% but slows to 1.31 M Tokens/h. These within-family trends reinforce that accuracy gains from scaling are accompanied by systematically lower throughput under a fixed evaluation protocol.

\begin{table*}[htbp]
\centering
\caption{Cost and efficiency of benchmarking LLMs on \modelname{}. We report total token usage (Tokens (M)), end-to-end runtime (Time (h)), API cost (Money (\$); not applicable to self-hosted models), throughput defined as total tokens divided by total time (Tokens (M)/h), and overall accuracy (Overall Acc (\%)$\uparrow$).}
\label{tab:cost_efficiency_overview}
\resizebox{0.7\linewidth}{!}{
\begin{tabular}{l|ccc|cc}
\toprule
\multirow{2}{*}{\textbf{Model}}
&\multicolumn{3}{c|}{\textbf{Total Cost}} &\textbf{Throughput}
&\textbf{Overall}\\
&\textbf{Tokens (M)}&\textbf{Time (h)}&\textbf{Money (\$)}&\textbf{(M Tokens/h)$\uparrow$}
&\textbf{Acc (\%)$\uparrow$}\\
\midrule
\multicolumn{6}{l}{\textit{Open source general LLMs}}\\
\midrule
glm4-9b & 10.46 & 3.40 & -- & 3.08 & 59.62 \\
glm4-32b & 10.42 & 10.53 & -- & 0.99 & 66.12 \\
llama3-8b & 10.47 & 2.50 & -- & 4.19 & 48.90 \\
llama3-70b & 10.45 & 7.73 & -- & 1.35 & 63.35 \\
llama3.1-8b & 10.51 & 3.20 & -- & 3.28 & 56.76 \\
llama3.2-3b & 10.41 & 1.27 & -- & 8.19 & 49.85 \\
llama3.3-70b & 10.45 & 20.87 & -- & 0.50 & 67.28 \\
mistral-7b & 12.87 & 3.33 & -- & 3.87 & 38.23 \\
ministral-8b & 10.51 & 2.72 & -- & 3.87 & 56.48 \\
mistral-small3-24b & 10.41 & 2.31 & -- & 4.51 & 65.01 \\
qwen2.5-3b & 10.54 & 1.79 & -- & 5.89 & 37.87 \\
qwen2.5-7b & 10.54 & 2.52 & -- & 4.18 & 57.74 \\
qwen2.5-32b & 10.37 & 7.14 & -- & 1.45 & 64.97 \\
qwen3-4b & 10.48 & 2.40 & -- & 4.36 & 60.63 \\
qwen3-8b & 10.54 & 3.27 & -- & 3.22 & 60.87 \\
qwen3-32b & 10.39 & 7.92 & -- & 1.31 & 66.78 \\
smollm3-3b & 10.29 & 1.01 & -- & 10.21 & 45.82 \\
yi-1.5-9b & 12.47 & 5.88 & -- & 2.12 & 45.51 \\
yi-1.5-34b & 13.04 & 5.17 & -- & 2.52 & 58.94 \\
\midrule
\multicolumn{6}{l}{\textit{Medical LLMs}}\\
\midrule
doctor-r1-8b & 10.54 & 2.48 & -- & 4.25 & 61.07 \\
med42-8b & 10.51 & 1.70 & -- & 6.19 & 36.48 \\
ultramedical-8b & 10.58 & 2.00 & -- & 5.30 & 29.02 \\
m1-32b-1k & 10.38 & 7.18 & -- & 1.45 & 63.21 \\
m1-7b-23k & 10.51 & 2.41 & -- & 4.37 & 46.08 \\
\midrule
\multicolumn{6}{l}{\textit{HIPAA compliant API-based LLMs}}\\
\midrule
gpt-4.1-nano & 10.04 & 1.71 & 2.02 & 5.88 & 60.48 \\
gpt-4.1-mini & 10.14 & 3.01 & 7.47 & 3.37 & 66.79 \\
gpt-4.1 & 10.11 & 6.45 & 40.67 & 1.57 & 69.43 \\
gpt-5-nano & 10.15 & 3.28 & 1.70 & 3.10 & 57.80 \\
gpt-5-mini & 10.16 & 3.71 & 8.50 & 2.74 & 66.12 \\
gpt-5 & 10.14 & 4.02 & 42.37 & 2.52 & 69.06 \\
gpt-5.2 & 12.14 & 12.17 & 71.03 & 1.00 & 70.91 \\
\bottomrule
\end{tabular}}
\end{table*}

\subsection{Error Analysis}
\label{sec:error_analysis}

We analyze errors from all evaluated models and categorize them into three types: (i) prediction errors (Prediction Wrong), which reflect knowledge or reasoning failures under correct formatting; (ii) missing structured outputs (No JSON), where the model does not return a valid JSON object; and (iii) malformed structured outputs (Output Malformed), where the output is not parseable under the required schema. Table~\ref{tab:error_analysis_overview} summarizes the per-model error breakdown.

\begin{table*}[htbp]
\centering
\caption{Error breakdown on \modelname{}. We report overall accuracy (Accuracy), missing structured outputs (No JSON), malformed structured outputs (Output Malformed), and prediction errors under valid formatting (Prediction Wrong). All values are percentages (\%).}
\label{tab:error_analysis_overview}
\resizebox{0.85\linewidth}{!}{
\begin{tabular}{l|c|ccc}
\toprule
\textbf{Model} & \textbf{Accuracy (\%)$\uparrow$} & \textbf{No JSON (\%)$\downarrow$} & \textbf{Output Malformed (\%)$\downarrow$} & \textbf{Prediction Wrong (\%)$\downarrow$} \\
\midrule
\multicolumn{5}{l}{\textit{Open source general LLMs}}\\
\midrule
glm4-9b & 59.62 & 0.00 & 0.01 & 40.36 \\
glm4-32b & 66.12 & 0.00 & 0.02 & 33.86 \\
llama3-8b & 48.90 & 0.00 & 2.51 & 48.60 \\
llama3-70b & 63.35 & 0.00 & 0.02 & 36.63 \\
llama3.1-8b & 56.76 & 0.00 & 0.04 & 43.20 \\
llama3.2-3b & 49.85 & 0.00 & 0.27 & 49.88 \\
llama3.3-70b & 67.28 & 0.00 & 0.00 & 32.72 \\
mistral-7b & 38.23 & 0.01 & 18.12 & 43.63 \\
ministral-8b & 56.48 & 0.00 & 0.63 & 42.89 \\
mistral-small3-24b & 65.01 & 0.00 & 0.00 & 34.99 \\
qwen2.5-3b & 37.87 & 0.00 & 13.16 & 48.97 \\
qwen2.5-7b & 57.74 & 0.00 & 0.77 & 41.50 \\
qwen2.5-32b & 64.97 & 0.00 & 0.10 & 34.94 \\
qwen3-4b & 60.63 & 0.00 & 0.00 & 39.37 \\
qwen3-8b & 60.87 & 1.36 & 0.05 & 37.72 \\
qwen3-32b & 66.78 & 0.44 & 0.04 & 32.74 \\
smollm3-3b & 45.82 & 0.00 & 0.57 & 53.61 \\
yi-1.5-9b & 45.51 & 0.73 & 2.22 & 51.54 \\
yi-1.5-34b & 58.94 & 0.00 & 0.01 & 41.05 \\
\midrule
\multicolumn{5}{l}{\textit{Medical LLMs}}\\
\midrule
doctor-r1-8b & 61.07 & 0.70 & 0.04 & 38.19 \\
med42-8b & 36.48 & 0.20 & 25.28 & 38.04 \\
ultramedical-8b & 29.02 & 1.04 & 17.44 & 52.50 \\
m1-32b-1k & 63.21 & 0.00 & 0.44 & 36.34 \\
m1-7b-23k & 46.08 & 0.37 & 4.43 & 49.12 \\
\midrule
\multicolumn{5}{l}{\textit{HIPAA compliant API-based LLMs}}\\
\midrule
gpt-4.1-nano & 60.48 & 0.00 & 0.01 & 39.51 \\
gpt-4.1-mini & 66.79 & 0.00 & 0.00 & 33.21 \\
gpt-4.1 & 69.43 & 0.00 & 0.01 & 30.56 \\
gpt-5-nano & 57.80 & 0.00 & 0.01 & 42.18 \\
gpt-5-mini & 66.12 & 0.00 & 0.00 & 33.88 \\
gpt-5 & 69.06 & 0.00 & 0.00 & 30.94 \\
gpt-5.2 & 70.91 & 0.00 & 0.00 & 29.09 \\
\bottomrule
\end{tabular}
}
\end{table*}

Across HIPAA compliant API-based LLMs, format-related failures are essentially eliminated: No JSON is 0.00 and Output Malformed is 0.00--0.01 for gpt-4.1-nano, gpt-4.1-mini, gpt-4.1, gpt-5-nano, gpt-5-mini, gpt-5, and gpt-5.2. Consequently, residual errors are dominated by Prediction Wrong, which ranges from 29.09\% (gpt-5.2) to 42.18\% (gpt-5-nano), mirroring the accuracy ordering (70.91\% to 57.80\%). This indicates that, for high-capability API models, the evaluation pipeline is not bottlenecked by response formatting, and measured performance primarily reflects decision quality rather than output compliance.

Most open-source general LLMs show near-perfect compliance (No JSON at 0.00 and Output Malformed close to 0.00), but several models exhibit non-negligible formatting failures that can materially affect end-to-end reliability. For example, mistral-7b shows 18.12\% Output Malformed, and qwen2.5-3b shows 13.16\%, both coinciding with low accuracies of 38.23\% and 37.87\%, respectively. A smaller but notable No JSON rate appears in qwen3-8b (1.36\%), qwen3-32b (0.44\%), and yi-1.5-9b (0.73\%), indicating that even strong families can occasionally violate structured-output constraints. In contrast, top open-source performers combine high accuracy with clean outputs, such as llama3.3-70b (67.28\% accuracy, 0.00 No JSON, 0.00 Output Malformed) and mistral-small3-24b (65.01\%, 0.00, 0.00).

Compared with general-purpose models, medical LLMs show markedly higher rates of structured-output failures, suggesting weaker instruction following under the same JSON-constrained protocol. med42-8b has 25.28\% Output Malformed with 36.48\% accuracy, and ultramedical-8b has 17.44\% Output Malformed and the highest No JSON rate in the table at 1.04\%, with 29.02\% accuracy. Even the strongest medical model by accuracy, m1-32b-1k (63.21\%), exhibits 0.44\% Output Malformed, while doctor-r1-8b shows a non-trivial No JSON rate of 0.70\% despite 61.07\% accuracy. These results imply that, for medical LLMs, reliability issues arise from both decision errors and output-format instability.

\subsection{Breakdown Analysis of Medical LLMs versus Base Models}
\label{sec:medical_vs_base}

The main experiment in Section~\ref{sec:main_results} shows that medical-domain adaptation does not consistently improve performance on \modelname{}. To further analyze this phenomenon, we compare medical LLMs with their corresponding base models using question-level aligned evaluation averaged across 4C/5C/6C MCQ variants. Table~\ref{tab:medical_vs_base} summarizes the overall and task-specific performance differences.

\begin{table*}[htbp]
\centering
\caption{Comparison between medical LLMs and their corresponding base models on \modelname{}. We report overall accuracy and task-specific accuracy differences for diagnosis (Dx), treatment (Tx), and prognosis (Px). Overall results are reported as medical model / base model ($\Delta$), and task-specific columns report the absolute accuracy difference $\Delta$ in percentage points.}
\label{tab:medical_vs_base}
\resizebox{0.60\linewidth}{!}{
\begin{tabular}{l|cccc}
\toprule
\textbf{Pair}
& \textbf{Overall Acc (\%)}
& \textbf{Dx $\Delta$}
& \textbf{Tx $\Delta$}
& \textbf{Px $\Delta$} \\
\midrule
med42-8b vs llama3-8b
& 36.0 / 48.0 (-12.08)
& -10.91
& -17.83
& -8.48 \\

m1-7b-23k vs qwen2.5-7b
& 45.4 / 56.9 (-11.52)
& -17.78
& -8.94
& -8.37 \\

m1-32b-1k vs qwen2.5-32b
& 62.3 / 64.2 (-1.92)
& -3.53
& -1.09
& -1.24 \\
\bottomrule
\end{tabular}
}
\end{table*}

Three consistent patterns emerge across these comparisons. First, current medical-domain fine-tuning does not reliably improve grounded EHR reasoning performance. This observation further highlights the difficulty of EHR-grounded clinical decision-making, since most existing medical LLM adaptation pipelines are not specifically optimized for reasoning over longitudinal structured EHR contexts. Second, the largest performance degradations are generally observed on Dx tasks, which often require disentangling confounded disease--disease relations and comorbidity patterns. Third, larger medical models narrow the gap relative to their base models, but do not consistently reverse the overall trend.

At the same time, we observe several narrow topic-level exceptions where medical adaptation provides localized gains. For example, m1-32b-1k improves Px performance on topics such as \textit{Angioedema} (+31.60), \textit{Bradycardia} (+18.78), and \textit{Hyperkalemia} (+16.67). m1-7b-23k shows localized improvements on \textit{Heart failure} Tx (+14.58) and \textit{Obstructive Sleep Apnea} Px (+13.68). med42-8b also improves on several Dx topics, including \textit{Cold intolerance} (+14.71) and \textit{Osteoporosis, unspecified} (+14.58). These exceptions suggest that domain-specific tuning may help in narrow clinical niches, even though it does not consistently improve general EHR-grounded clinical decision-making performance.

Overall, these results suggest that strong performance on \modelname{} requires capabilities beyond biomedical terminology familiarity or medical text exposure alone. Models must reason over real longitudinal EHR contexts and resolve clinically confounded relations, including disease progression patterns and disease--treatment associations. Improving these capabilities may require training signals beyond conventional domain adaptation, such as large-scale clinical case supervision and decision-focused objectives. EHR-grounded resources such as \modelname{} may therefore provide a useful foundation for future clinical reasoning-oriented model development. Similar observations have also been reported in prior work~\cite{dorfner2025evaluating,xu2025medagentgym,dorfner2024biomedical}.

\subsection{Comparison with Embedding-based non-LLM Baselines}
\label{sec:retrieval_baselines}

To provide additional reference points beyond LLM-based evaluation, we compare \modelname{} with several embedding-based retrieval baselines under the same zero-shot QA setting. For each question, the model encodes the question together with each candidate option and selects the option with the highest cosine similarity. We evaluate these methods on the same 27{,}000 6C questions used in the main experiment in Section~\ref{sec:main_results}.

\begin{table*}[htbp]
\centering
\caption{Comparison between embedding-based retrieval baselines and LLMs on \modelname{}. We report overall accuracy and breakdowns by clinical decision task and data source. Dx, Tx, and Px denote diagnosis, treatment, and prognosis, respectively; MIII, MIV, and PRO denote MIMIC-III, MIMIC-IV, and PROMOTE, respectively. All values are percentages (\%).}
\label{tab:retrieval_baselines}
\resizebox{0.55\linewidth}{!}{
\begin{tabular}{l|ccccccc}
\toprule
\textbf{Model}
& \textbf{Overall}
& \textbf{Dx}
& \textbf{Tx}
& \textbf{Px}
& \textbf{MIII}
& \textbf{MIV}
& \textbf{PRO} \\
\midrule
SapBERT & 16.5 & 15.8 & 12.7 & 21.2 & 17.1 & 17.0 & 15.6 \\
SentenceTransformer & 27.0 & 30.3 & 19.0 & 31.7 & 29.0 & 28.0 & 23.9 \\
PubMedBERT & 32.8 & 36.9 & 26.2 & 35.3 & 35.2 & 32.8 & 30.3 \\
\midrule
llama3-8b & 43.8 & 39.6 & 58.4 & 33.4 & 45.1 & 43.4 & 42.9 \\
qwen3-8b & 55.3 & 52.2 & 70.4 & 43.3 & 56.1 & 57.1 & 52.7 \\
gpt-5.2 & 66.8 & 68.1 & 77.3 & 55.1 & 67.3 & 66.9 & 66.2 \\
\bottomrule
\end{tabular}
}
\end{table*}

Embedding-based retrieval baselines consistently underperform reasoning-capable LLMs across all evaluation settings. Even the strongest biomedical encoder baseline, PubMedBERT, achieves only 32.8\% overall accuracy, substantially below general-purpose LLMs such as llama3-8b (43.8\%) and qwen3-8b (55.3\%). The gap becomes even larger for the strongest API-based model, gpt-5.2 (66.8\%).

The largest performance differences are observed on treatment questions. PubMedBERT reaches only 26.2\% accuracy on Tx, whereas qwen3-8b and gpt-5.2 achieve 70.4\% and 77.3\%, respectively. This pattern suggests that many treatment questions in \modelname{} cannot be solved through simple semantic similarity or terminology matching alone. Instead, successful prediction often requires reasoning over clinically grounded relations between diagnoses, interventions, and longitudinal patient context.

Overall, these results support the design objective of \modelname{} as a benchmark for clinical reasoning rather than retrieval-oriented matching. Strong performance requires models to integrate biomedical knowledge with contextual inference over structured EHR-derived scenarios, which remains challenging for embedding-only retrieval approaches.

\subsection{Robustness to QA Generation LLM Choice}
\label{sec:source_model_bias}

In the main construction pipeline of \modelname{}, we use HuatuoGPT-o1-8B as the primary source LLM for generating QA templates and question instances. Although this choice provides a consistent generation protocol, it may raise a potential LLM bias concern. To examine this issue and strengthen the robustness and validity of \modelname{}, we conduct an additional source-model bias analysis.

Specifically, we regenerate held-out 4C subsets from the same 400 patients using three different medical LLMs as source generators: HuatuoGPT-o1-7B, HuatuoGPT-o1-8B, and m1-7b-23k. Each regenerated subset covers three data sources (MIII/MIV/PRO) and three clinical decision tasks (Dx/Tx/Px). We then evaluate six strong open-source LLMs on each regenerated subset. This design allows us to test whether the relative ordering of evaluated models remains stable when the QA generation model changes while the underlying patient set, task coverage, and evaluation protocol are held fixed. The results are summarized in Table~\ref{tab:source_model_bias}.

\begin{table*}[htbp]
\centering
\caption{Performance comparison by changing QA generation LLM choice on \modelname{}. Since HuatuoGPT-o1-8B is used as the primary LLM for \modelname{} generation in the main construction pipeline, we regenerate held-out 4C subsets from the same 400 patients using three QA generation LLMs and evaluate six open-source LLMs on each subset. All values are accuracy percentages (\%).}
\label{tab:source_model_bias}
\resizebox{0.55\linewidth}{!}{
\begin{tabular}{l|ccc}
\toprule
\textbf{Eval Model}
& \textbf{HuatuoGPT-o1-7B}
& \textbf{HuatuoGPT-o1-8B}
& \textbf{m1-7b-23k} \\
\midrule
llama3-70b & 61.1 & 69.1 & 61.4 \\
llama3-8b & 48.4 & 53.3 & 43.6 \\
qwen2.5-32b & 59.6 & 66.2 & 58.8 \\
qwen2.5-7b & 57.3 & 62.4 & 56.5 \\
qwen3-32b & 61.8 & 70.1 & 62.4 \\
qwen3-8b & 56.1 & 66.9 & 56.9 \\
\bottomrule
\end{tabular}
}
\end{table*}

The absolute accuracies vary across source generators, suggesting that different source LLMs can produce QA subsets with different difficulty levels. However, the relative model ordering remains highly stable. Kendall's W across the three source-generator settings is 0.937 ($p=0.015$), and the pairwise Spearman correlations are 0.829, 0.943, and 0.943. These results indicate that, although the source generator affects the absolute difficulty of the regenerated subset, the main comparative conclusions are robust to the choice of source LLM. Therefore, the observed model rankings in \modelname{} are unlikely to be an artifact of relying on HuatuoGPT-o1-8B as a single benchmark-construction model.

\subsection{Robustness to Context Event Size}
\label{sec:context_event_size}

We further evaluate whether the main conclusions are sensitive to the number of context events used to construct the question scenario. In the main construction pipeline of \modelname{}, each QA scenario is generated from a compact EHR context consisting of two context events together with the relation-subject entity. This design follows the benchmark objective of evaluating clinical decision-making under partial observation, rather than long-context retrieval over a large number of EHR events.

To assess whether this design choice affects the relative model comparison, we construct an aligned 4C subset with 300 templates across three clinical decision tasks (Dx/Tx/Px) and three data sources (MIII/MIV/PRO). We then vary the number of context events from 2 to 4 to 6 while keeping the templates, answer choices, evaluated models, and inference protocol fixed. The results are shown in Table~\ref{tab:context_event_size}.

\begin{table*}[htbp]
\centering
\caption{Performance comparison by changing context event size on \modelname{}. In the main construction pipeline, each QA scenario uses two context events together with the relation-subject entity. We evaluate an aligned 4C subset while varying the number of context events from 2 to 4 to 6. All values are accuracy percentages (\%).}
\label{tab:context_event_size}
\resizebox{0.4\linewidth}{!}{
\begin{tabular}{l|ccc}
\toprule
\textbf{Model}
& \textbf{2 Events}
& \textbf{4 Events}
& \textbf{6 Events} \\
\midrule
llama3-70b & 45.3 & 45.0 & 44.7 \\
llama3-8b & 36.3 & 36.1 & 35.5 \\
qwen2.5-32b & 44.7 & 44.1 & 44.6 \\
qwen2.5-7b & 41.4 & 41.4 & 38.7 \\
qwen3-32b & 46.9 & 46.9 & 46.7 \\
qwen3-8b & 40.5 & 40.4 & 37.9 \\
\bottomrule
\end{tabular}
}
\end{table*}

The relative model ordering remains the same across all three context-event settings, suggesting that the main conclusions are not driven by the specific use of two context events in the main benchmark. Larger 32B/70B models are comparatively robust when additional local context is provided, whereas smaller 7B/8B models show slightly greater sensitivity. For example, qwen3-32b remains nearly unchanged from 46.9\% with 2 events to 46.7\% with 6 events, while qwen3-8b decreases from 40.5\% to 37.9\%. Overall, these results indicate that increasing the amount of local EHR context does not materially change the model ranking pattern, further supporting the robustness of the main evaluation protocol.

\section{Additional Experiment: Testing Reasoning LLMs on \modelname{}}
\label{sec:reasoning_models}

Recent reasoning-oriented LLMs explicitly produce intermediate reasoning traces during inference, which substantially increases token usage and may interact with context-length constraints. To prevent these token-intensive behaviors from confounding the main benchmarking results and ensure a fair comparison, we report a separate analysis that characterizes accuracy-efficiency trade-offs under different reasoning-effort configurations. All experiments in this section use a maximum context length of 10{,}240 tokens. 

\begin{figure*}[htbp]
  \centering
  \includegraphics[width=\linewidth]{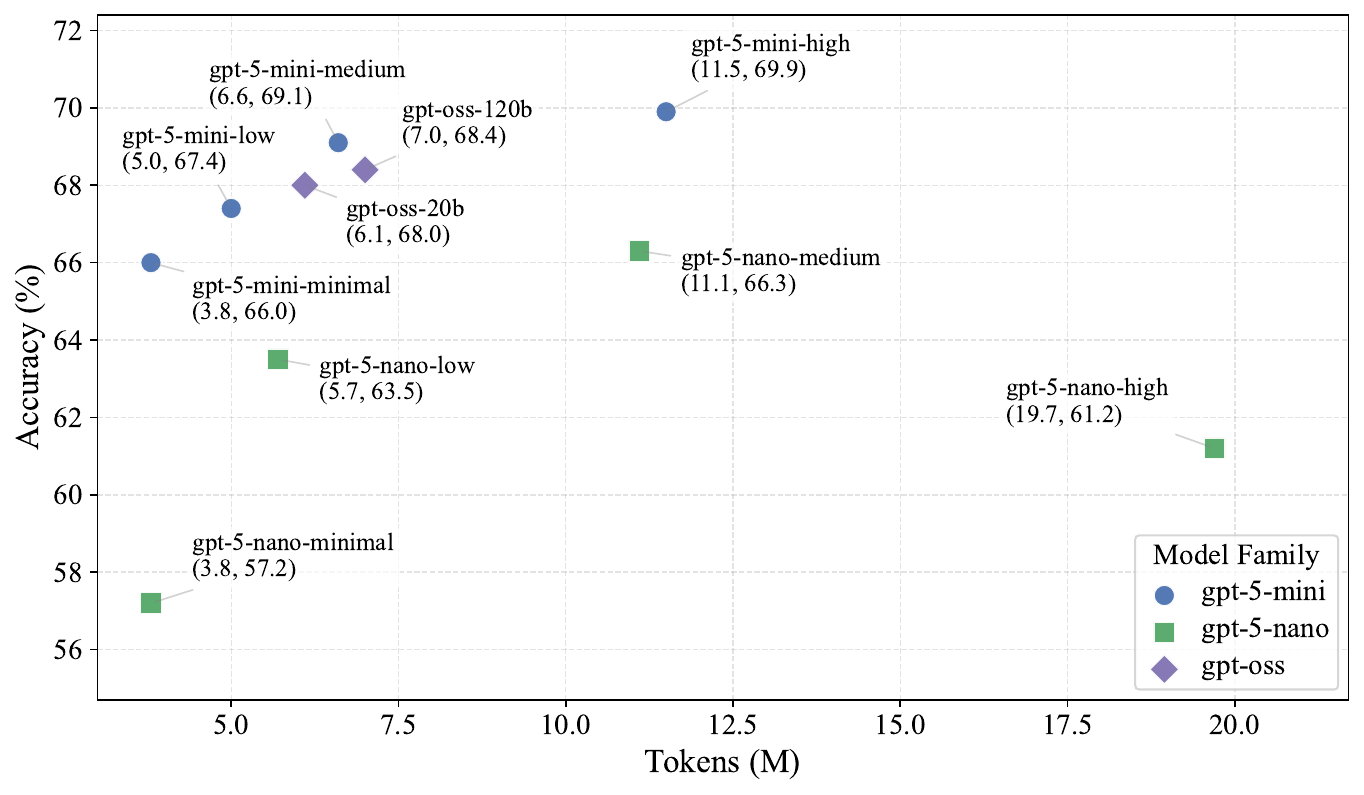}
  \caption{Overall performance and token cost of reasoning model configurations on \modelname{}. Each point corresponds to one configuration and is annotated with total token usage (in millions) and overall accuracy.}
  \label{fig:thinking_model}
\end{figure*}

We evaluate ten configurations in total. For gpt-5-nano and gpt-5-mini~\cite{singh2025gpt5}, we test four reasoning-effort settings (minimal/low/medium/high), where minimal matches the setting used in the main experiments. We additionally include gpt-oss-20b and gpt-oss-120b~\cite{openai2025gptoss}, which also enforce explicit reasoning output and therefore incur substantially higher token costs than the standard models evaluated in the main benchmark. To control evaluation cost while maintaining broad coverage, all reasoning models are served through the HIPAA-compliant Azure API, and evaluation is performed on a fixed subset of 1{,}000 QA items for each source--task--MCQ-type combination. This protocol yields $1{,}000 \times 3 \times 3 \times 3 = 27{,}000$ questions per configuration. Figure~\ref{fig:thinking_model} summarizes overall accuracy and total token usage across these configurations. Other settings are kept the same as the main experiment. 

Figure~\ref{fig:thinking_model} indicates that higher model capacity and greater reasoning effort generally correspond to higher accuracy and higher token cost, which is consistent with the expected scaling trends and supports the validity of the \modelname{} construction pipeline. At matched effort levels, gpt-5-mini outperforms gpt-5-nano, and gpt-oss-120b slightly exceeds gpt-oss-20b. Within gpt-5-mini, increasing effort from minimal to low and then to medium yields monotonic gains. A notable exception occurs for gpt-5-nano: the high-effort setting underperforms the medium setting despite consuming substantially more tokens (19.7M vs.\ 11.1M). Output inspection indicates that gpt-5-nano-high more frequently fails to return a final decision due to exceeding context or token limits (approximately 10\% of cases), which plausibly explains the accuracy degradation at the highest effort level.

The results also exhibit a consistent performance-cost trade-off, as higher reasoning effort incurs diminishing returns. For gpt-5-mini, moving from minimal (3.8, 66.0) to low (5.0, 67.4) and then to medium (6.6, 69.1) yields steady improvements, whereas the step from medium to high increases accuracy by only 0.8 points (69.1$\rightarrow$69.9) while increasing token usage by 4.9M (6.6$\rightarrow$11.5). This pattern suggests that medium-level reasoning can provide a more efficient operating point when token budgets are constrained.

\section{Additional Experiment: Testing LLMs with Multiple Versions of Questions from \modelname{}}
\label{sec:multi_version}

\begin{table*}[htbp]
\centering
\caption{Performance on multiple deterministic question versions in \modelname{}. Acc (\%) denotes mean accuracy; V-Std (pp) denotes the standard deviation of accuracy across versions of the same question; V-Cons. (\%) denotes the fraction of questions whose predicted option remains identical across versions of the same question. Overall aggregates results across 4/5/6-choice MCQs.}
\label{tab:multi_version_results}
\resizebox{0.95\linewidth}{!}{
\begin{tabular}{l|ccc|ccc|ccc|ccc}
\toprule
\multirow{2}{*}{\textbf{Model}} &
\multicolumn{3}{c|}{\textbf{Overall (\%)}} &
\multicolumn{3}{c|}{\textbf{4C (\%)}} &
\multicolumn{3}{c|}{\textbf{5C (\%)}} &
\multicolumn{3}{c}{\textbf{6C (\%)}} \\
& \textbf{Acc} & \textbf{V-Std} & \textbf{V-Cons.}
& \textbf{Acc} & \textbf{V-Std} & \textbf{V-Cons.}
& \textbf{Acc} & \textbf{V-Std} & \textbf{V-Cons.}
& \textbf{Acc} & \textbf{V-Std} & \textbf{V-Cons.} \\
\midrule
glm4-9b      & 58.63 & 2.03 & 81.62 & 64.11 & 1.78 & 84.73 & 58.28 & 2.04 & 81.62 & 53.49 & 2.17 & 78.50 \\
llama3-8b    & 46.82 & 3.01 & 65.18 & 53.74 & 2.96 & 71.11 & 45.95 & 3.16 & 64.53 & 40.76 & 2.92 & 59.89 \\
llama3.2-3b  & 48.90 & 2.27 & 70.80 & 54.83 & 2.06 & 75.36 & 48.10 & 2.36 & 70.34 & 43.77 & 2.33 & 66.70 \\
ministral-8b & 55.71 & 2.22 & 78.35 & 61.24 & 2.51 & 80.73 & 55.33 & 1.85 & 78.19 & 50.55 & 2.30 & 76.12 \\
qwen2.5-3b   & 36.88 & 3.22 & 60.29 & 46.16 & 3.34 & 67.66 & 36.07 & 2.60 & 59.62 & 28.41 & 3.58 & 53.59 \\
qwen2.5-7b   & 57.11 & 1.99 & 81.38 & 61.94 & 1.97 & 84.11 & 56.84 & 1.57 & 81.32 & 52.54 & 2.30 & 78.71 \\
qwen2.5-32b  & 64.32 & 1.64 & 88.36 & 69.29 & 1.67 & 90.31 & 63.86 & 1.52 & 88.21 & 59.80 & 1.72 & 86.55 \\
qwen3-4b     & 60.30 & 1.74 & 83.93 & 65.77 & 1.72 & 86.97 & 60.06 & 1.42 & 84.13 & 55.06 & 1.98 & 80.68 \\
qwen3-8b     & 60.90 & 1.91 & 84.45 & 66.57 & 1.92 & 87.30 & 60.51 & 1.44 & 84.37 & 55.61 & 2.21 & 81.68 \\
qwen3-32b    & 65.98 & 1.73 & 87.15 & 70.77 & 1.87 & 89.10 & 65.96 & 1.29 & 87.58 & 61.20 & 1.94 & 84.78 \\
\bottomrule
\end{tabular}
}
\end{table*}

To further evaluate robustness to multiple deterministic question versions in \modelname{}, we conduct an extended evaluation. Each version paraphrases the clinical context while preserving the same clinical meaning; meanwhile, answer options are systematically permuted so that each option, including the correct answer, appears in each position exactly once across versions. We select ten representative open-source LLMs spanning model scales to support a fair and comprehensive comparison: small models (llama3.2-3b, qwen2.5-3b, qwen3-4b), mid-sized models (glm4-9b, llama3-8b, ministral-8b, qwen2.5-7b, qwen3-8b), and large models (qwen2.5-32b, qwen3-32b). We evaluate the first 1{,}000 questions for 4-choice, 5-choice, and 6-choice MCQs, with 4/5/6 versions, respectively. In total, each model is evaluated on $1{,}000 \times 15 \times 3 \times 3 = 135{,}000$ questions. Other settings are kept the same as in the main experiment.

We evaluate models using three metrics: accuracy (Acc), variability across versions (V-Std), and prediction consistency across versions (V-Cons.).
Let $c\in\{4,5,6\}$ denote the choice size, and let $V_c=c$ denote the number of deterministic versions for $c$-choice questions (paraphrase + answer permutation).
For each base question $q$, the model produces one predicted option $\hat{y}_{q}^{(v)}\in\{A,\dots\}$ under version $v\in\{1,\dots,V_c\}$.

For each version $v$, define the version-level accuracy as
\begin{equation}
\mathrm{Acc}^{(v)} \;=\; \frac{1}{|\mathcal{Q}|}\sum_{q\in\mathcal{Q}} \mathbb{I}\!\left[\hat{y}_{q}^{(v)} = y_q\right],
\end{equation}
where $\mathcal{Q}$ is the evaluated set of base questions and $y_q$ is the gold answer. We report $\mathrm{Acc}$ as the mean of $\mathrm{Acc}^{(v)}$ over versions.

To quantify robustness to version perturbations, we measure the standard deviation of version-level accuracies:
\begin{equation}
\mathrm{V\text{-}Std} \;=\; \sqrt{\frac{1}{V_c}\sum_{v=1}^{V_c}\left(\mathrm{Acc}^{(v)} - \overline{\mathrm{Acc}}\right)^2},
\quad
\overline{\mathrm{Acc}} \;=\; \frac{1}{V_c}\sum_{v=1}^{V_c}\mathrm{Acc}^{(v)}.
\end{equation}
A smaller $\mathrm{V\text{-}Std}$ indicates more stable performance across paraphrase/permutation versions under the same choice size.

We further measure whether a model makes the \emph{same} prediction across versions for each base question.
Define a per-question consistency indicator:
\begin{equation}
\mathrm{Cons}(q) \;=\; \mathbb{I}\!\left[\hat{y}_{q}^{(1)}=\hat{y}_{q}^{(2)}=\cdots=\hat{y}_{q}^{(V_c)}\right].
\end{equation}
The overall consistency is then
\begin{equation}
\mathrm{V\text{-}Cons.} \;=\; \frac{1}{|\mathcal{Q}|}\sum_{q\in\mathcal{Q}} \mathrm{Cons}(q).
\end{equation}
Higher $\mathrm{V\text{-}Cons.}$ implies that predictions are less sensitive to version perturbations, complementing $\mathrm{V\text{-}Std}$, which measures accuracy fluctuation at the aggregate level.

The evaluation results are reported in Table \ref{tab:multi_version_results}. Across all settings, qwen3-32b achieves the highest overall accuracy at 65.98\%, followed by qwen2.5-32b at 64.32\%. The relative ordering among models is consistent with that in the main experiment, which further supports the correctness of the \modelname{} construction pipeline.

Moreover, both V-Std and V-Cons. indicate strong stability across versions, suggesting that evaluation on a single version in the main experiment is reasonable. For example, qwen2.5-32b attains the highest overall consistency (V-Cons.) of 88.36\% with the lowest overall variability (V-Std) of 1.64(pp), while qwen3-32b remains close in consistency (87.15) with a V-Std of 1.73. These results indicate that the evaluated LLMs are robust to deterministic paraphrasing and answer-option permutations, reducing the likelihood that the reported performance is driven by an arbitrary or randomly chosen question version rather than the underlying model capability for CDM.

Increasing the number of answer options consistently reduces accuracy and consistency across all models, and the degradation is substantially larger for weaker models. For the strongest models, the accuracy drops from 4C to 6C is approximately ten percentage points; for example, qwen3-32b decreases from 70.77\% (4C) to 61.20\% (6C), and qwen2.5-32b decreases from 69.29\% to 59.80\% (a 9.49-point drop). In contrast, qwen2.5-3b exhibits a much sharper decline from 46.16\% to 28.41\% (a 17.75-point drop). A similar pattern appears in V-Cons.: qwen2.5-32b drops modestly from 90.31 to 86.55, whereas qwen2.5-3b drops substantially from 67.66 to 53.59. These results suggest that increasing choice cardinality amplifies both difficulty and sensitivity to model capacity, particularly for reliability.

The results reveal a strong qualitative coupling between accuracy and stability: higher-accuracy models generally exhibit lower V-Std and higher V-Cons. For example, qwen2.5-32b and qwen3-32b jointly occupy the top tier in accuracy while maintaining high consistency (above 87) and low variability (at or below 1.73). Nevertheless, the comparison between these two models indicates a nuanced tradeoff: qwen3-32b yields the best overall accuracy (65.98\%), whereas qwen2.5-32b achieves slightly higher overall consistency (88.36 vs.\ 87.15) and the lowest overall V-Std (1.64). This separation between peak accuracy and peak reliability suggests that both metrics are necessary for characterizing clinically relevant robustness, especially under harder settings such as 6C, where both accuracy and consistency decline across all models.

\section{Additional Experiment: Testing LLMs with Extended Questions from \modelname{}}
\label{sec:full_eval}

\begin{table*}[htbp]
\centering
\caption{Additional evaluation on extended questions from \modelname{} across tasks, sources, and question types. We use abbreviations Dx/Tx/Px for diagnosis/treatment/prognosis decision task, MIII/MIV/PRO for MIMIC-III/MIMIC-IV/PROMOTE, and 4C/5C/6C for 4/5/6-choice MCQs. We additionally report evaluation cost in tokens (M) and time (h).}
\label{tab:full_eval}
\resizebox{\linewidth}{!}{
\begin{tabular}{l|c|ccc|ccc|ccc|cc}
\toprule
\multirow{2}{*}{\textbf{Model}}
&\textbf{Overall}
&\multicolumn{3}{c|}{\textbf{Task Acc.}}
&\multicolumn{3}{c|}{\textbf{Source Acc.}}
&\multicolumn{3}{c|}{\textbf{Type Acc.}}
&\multicolumn{2}{c}{\textbf{Cost}}\\
&\textbf{Acc (\%)$\uparrow$}
&\textbf{Dx (\%)}&\textbf{Tx (\%)}&\textbf{Px (\%)}
&\textbf{MIII (\%)}&\textbf{MIV (\%)}&\textbf{PRO (\%)}
&\textbf{4C (\%)}&\textbf{5C (\%)}&\textbf{6C (\%)}
&\textbf{Tokens (M)$\downarrow$}&\textbf{Time (h)$\downarrow$}\\
\midrule
glm4-9b       & 59.97 & 58.36 & 72.26 & 49.28 & 62.15 & 60.42 & 57.33 & 65.05 & 59.74 & 55.12 & 34.24 & 7.14 \\
llama3-8b     & 48.95 & 43.46 & 63.97 & 39.42 & 50.20 & 48.73 & 47.92 & 55.48 & 47.95 & 43.41 & 34.31 & 3.89 \\
ministral-8b  & 56.86 & 53.15 & 71.68 & 45.75 & 58.47 & 57.02 & 55.08 & 61.96 & 56.45 & 52.16 & 36.39 & 4.74 \\
qwen2.5-7b    & 58.16 & 56.69 & 71.56 & 46.22 & 59.85 & 58.86 & 55.77 & 62.79 & 57.91 & 53.78 & 34.43 & 4.53 \\
qwen3-4b      & 60.95 & 59.41 & 73.18 & 50.24 & 62.73 & 60.68 & 59.43 & 66.46 & 60.75 & 55.63 & 34.38 & 5.38 \\
qwen3-8b      & 61.28 & 58.36 & 74.36 & 51.12 & 63.05 & 61.19 & 59.59 & 67.14 & 60.96 & 55.73 & 34.64 & 7.69 \\
llama3.2-3b   & 49.66 & 43.07 & 64.75 & 41.16 & 50.34 & 50.15 & 48.48 & 55.33 & 48.94 & 44.70 & 34.38 & 3.15 \\
qwen2.5-32b   & 64.95 & 66.29 & 76.39 & 52.18 & 65.51 & 65.72 & 63.62 & 69.86 & 64.59 & 60.41 & 34.43 & 23.11 \\
qwen2.5-3b    & 38.05 & 35.37 & 47.80 & 30.97 & 39.45 & 38.99 & 35.71 & 47.67 & 36.69 & 29.79 & 34.48 & 4.17 \\
qwen3-32b     & 66.42 & 67.52 & 76.69 & 55.04 & 67.84 & 66.08 & 65.33 & 71.01 & 66.06 & 62.17 & 25.73 & 25.49 \\
\bottomrule
\end{tabular}
}
\end{table*}

In the main experiment, to ensure efficiency and a fair comparison across tasks, data sources, and question types, we evaluate a fixed subset consisting of the first 3{,}000 questions for each task--source--type combination. To further assess whether this subset-based protocol faithfully reflects model behavior at scale, we conduct an additional evaluation over the extended question set, which covers all extracted clinical relations in \modelname{}.
We select ten representative open-source LLMs spanning a wide range of model scales: small models (llama3.2-3b, qwen2.5-3b, qwen3-4b), mid-sized models (glm4-9b, llama3-8b, ministral-8b, qwen2.5-7b, qwen3-8b), and large models (qwen2.5-32b, qwen3-32b).
For each model, we evaluate all verified templates of each multiple-choice format (4-choice, 5-choice, and 6-choice), resulting in a total of 180{,}517 evaluated questions per model. All other settings are identical to those in the main experiment. We report the results in Table~\ref{tab:full_eval}.

Overall, the extended evaluation yields highly consistent conclusions with the main experiment. In particular, both the overall accuracy and the relative ranking of models closely match those observed under the subset protocol (only 0.15\% overall accuracy difference across all models), indicating that the subset-based results are not driven by sampling artifacts or a particular slice of questions. Across the ten representative LLMs spanning small to large scales, the same top-performing models remain at the top and the same weaker models remain at the bottom, with only minor fluctuations in absolute accuracy. This stability suggests that the first 3{,}000 questions per task--source--type provide sufficient coverage of the underlying relation and question distributions, and that model comparisons are robust to expanding the evaluation set. Consequently, the fixed-subset design offers a practical yet reliable proxy for extended-scale benchmarking, enabling efficient experimentation while preserving the key comparative conclusions about model capability.

Beyond overall performance, the same structural patterns persist across granular slices: treatment remains the easiest task for every evaluated model, whereas prognosis is consistently the most challenging, indicating a stable task-level difficulty imbalance rather than model-specific noise. Source-level differences are also small, suggesting limited sensitivity to data source choice under our evaluation setting. Finally, accuracy monotonically decreases as the number of answer options increases (MCQ-4 $>$ MCQ-5 $>$ MCQ-6), consistent with the main experiment. Taken together, these results further validate that the main experimental design captures the key performance trends and comparative conclusions of extended-scale evaluation on \modelname{}.

\section{Additional Experiment: Testing LLMs with Open-ended Questions from \modelname{}}
\label{sec:open_questions}

\begin{table*}[htbp]
\centering
\caption{Performance on Open-ended questions (OEQs) of \modelname{}. We report RC, ROUGE-1, ROUGE-L, and BERTScore (all in \%), together with token usage in millions  and runtime in hours.}
\label{tab:open_ended_results}
\resizebox{0.8\linewidth}{!}{
\begin{tabular}{l|cccc|cc}
\toprule
\textbf{Model} & \textbf{RC (\%)} & \textbf{ROUGE-1 (\%)} & \textbf{ROUGE-L (\%) }& \textbf{BERTScore (\%)} & \textbf{Tokens (M)} & \textbf{Time (h)} \\
\midrule
glm4-9b     & 28.32 & 22.98 & 20.46 & 42.39 & 1.21 & 1.13 \\
llama3-8b   & 26.93 & 28.69 & 24.03 & 45.82 & 1.32 & 1.38 \\
llama3.2-3b & 10.87 & 17.49 & 15.40 & 21.20 & 1.28 & 0.74 \\
ministral-8b& 39.86 & 28.92 & 25.60 & 47.50 & 1.27 & 1.13 \\
qwen2.5-3b  &  4.39 & 22.85 & 18.37 & 40.25 & 1.23 & 0.73 \\
qwen2.5-7b  & 36.09 & 27.40 & 24.57 & 46.32 & 1.25 & 0.91 \\
qwen2.5-32b & 68.24 & 34.05 & 30.82 & 56.25 & 1.29 & 3.79 \\
qwen3-4b    & 42.89 & 31.41 & 27.03 & 52.08 & 1.32 & 1.44 \\
qwen3-8b    & 61.86 & 31.41 & 27.83 & 52.17 & 1.31 & 1.77 \\
qwen3-32b   & 66.22 & 30.54 & 27.75 & 52.31 & 1.97 & 7.20 \\
\bottomrule
\end{tabular}
}
\end{table*}

To further evaluate LLMs on paraphrased open-ended questions (OEQs) in \modelname{}, we conduct an extended evaluation using ten representative open-source LLMs spanning different model scales to support a fair and comprehensive comparison, including small models (llama3.2-3b, qwen2.5-3b, qwen3-4b), mid-sized models (glm4-9b, llama3-8b, ministral-8b, qwen2.5-7b, qwen3-8b), and large models (qwen2.5-32b, qwen3-32b). Considering efficiency, we evaluate the first 1{,}000 OEQs for each source-task setting; in total, each model is evaluated on $1{,}000 \times 3 \times 3 = 9{,}000$ questions. Other settings are kept the same as in the main experiment.

We report four automatic metrics for OEQ evaluation, including RC, ROUGE-1, ROUGE-L, and BERTScore. Specifically, for each OEQ item $I_j=(S_j,Q_j,B_j)$ derived from a template $P_k$, the model produces a free-text answer $\hat{a}_j$, which is compared against the reference rationale $a_j$ stored in the template. RC measures whether $\hat{a}_j$ covers the target clinical relation $R_k=(x_k,r_k,y_k)$ by checking whether the answer recovers the intended entity (or clinically equivalent surface forms) $x_k$ under concept normalization. ROUGE-1 and ROUGE-L quantify lexical overlap between $\hat{a}_j$ and $a_j$, where ROUGE-1 emphasizes unigram-level overlap and ROUGE-L captures sequence-level similarity via the longest common subsequence. BERTScore measures semantic similarity between $\hat{a}_j$ and $a_j$ using contextual token embeddings (from the BERT model \textit{bert-base-uncased}) and soft matching, which makes it less sensitive to paraphrasing than ROUGE. For efficiency, we aggregate the total number of (prompt+completion) tokens consumed across all evaluated OEQs and the total wall-clock runtime to obtain Tokens (M) and Time (h), respectively. The results are presented in Table~\ref{tab:open_ended_results}.

The results in Table~\ref{tab:open_ended_results} show a clear scale-dependent trend that is consistent with the main experiments: larger models achieve substantially better OEQ quality across all metrics, while small models perform poorly. In particular, qwen2.5-32b achieves the strongest overall quality, reaching 68.24\% RC, 34.05\% ROUGE-1, 30.82\% ROUGE-L, and 56.25\% BERTScore, which outperforms all other evaluated models. Notably, qwen3-32b underperforms qwen2.5-32b across all reported metrics, indicating that model scaling alone does not guarantee superior OEQ performance across model families. Overall, these results align with the observations in the main experiments and further support that \modelname{} can reliably differentiate the open-ended clinical reasoning capabilities of LLMs at different scales.

OEQs also reveal a strong quality-efficiency trade-off. The fastest runtimes (0.73--0.74h) are achieved by llama3.2-3b and qwen2.5-3b, but both exhibit severe quality loss: llama3.2-3b drops to 10.87\% RC and 21.20\% BERTScore, while qwen2.5-3b is even lower in RC (4.39\%) despite a moderate BERTScore (40.25\%). These results indicate that speed alone does not guarantee usable OEQ performance and suggest a clear lower-capacity regime where models generate quickly but fail to meet accuracy-oriented criteria. In contrast, mid-sized models provide more reliable quality with modest cost (e.g., ministral-8b reaches 39.86\% RC with 1.13h runtime, and qwen3-8b reaches 61.86\% RC with 1.77h runtime). Finally, although qwen3-32b approaches the best RC (66.22\% versus 68.24\% for qwen2.5-32b), it incurs substantially higher cost (1.97M tokens and 7.20h), which makes it less attractive under runtime constraints. Overall, these findings suggest that OEQs are more demanding and benefit more from stronger models, and that model scale should be selected to balance quality and efficiency.

\section{LLMs Utilized in \modelname{}}
\label{sec:llm_modelcard}
In this paper, we leverage more than 30 representative LLMs released between 2023 and 2025. The set of evaluated LLMs is large and up-to-date, supporting meaningful conclusions about current LLM performance trends on \modelname{}. Their detailed descriptions and access links are provided below. 

\begin{itemize}[itemsep=1pt, topsep=0pt, parsep=0pt, partopsep=0pt]
    \item \textbf{glm4-9b}:
    GLM-4 instruction model with 9B parameters, released as a general-purpose bilingual (Chinese--English) LLM for conversational generation and instruction following.
    \newline \hspace*{1.5em}\textit{HuggingFace:} \href{https://huggingface.co/zai-org/GLM-4-9B-0414}{https://huggingface.co/zai-org/GLM-4-9B-0414}

    \item \textbf{glm4-32b}:
    Larger GLM-4 instruction model with 32B parameters, providing higher capacity than the 9B variant and typically used when stronger generation quality is desired under similar prompting.
    \newline \hspace*{1.5em}\textit{HuggingFace:} \href{https://huggingface.co/zai-org/GLM-4-32B-0414}{https://huggingface.co/zai-org/GLM-4-32B-0414}

    \item \textbf{llama3-8b}:
    \textsc{Llama 3} instruction model with 8B parameters, a general-purpose open-weight LLM commonly used as an efficient baseline for instruction following and text generation.
    \newline \hspace*{1.5em}\textit{HuggingFace:} \href{https://huggingface.co/meta-llama/Meta-Llama-3-8B}{https://huggingface.co/meta-llama/Meta-Llama-3-8B}

    \item \textbf{llama3-70b}:
    \textsc{Llama 3} instruction model with 70B parameters, a larger-capacity variant designed to improve performance on knowledge-intensive generation and complex instruction-following workloads.
    \newline \hspace*{1.5em}\textit{HuggingFace:} \href{https://huggingface.co/meta-llama/Meta-Llama-3-70B}{https://huggingface.co/meta-llama/Meta-Llama-3-70B}

    \item \textbf{llama3.1-8b}:
    \textsc{Llama 3.1} instruction model with 8B parameters, an updated release in the Llama family that is used as a drop-in general-purpose model under the same prompting interface.
    \newline \hspace*{1.5em}\textit{HuggingFace:} \href{https://huggingface.co/meta-llama/Meta-Llama-3.1-8B}{https://huggingface.co/meta-llama/Meta-Llama-3.1-8B}

    \item \textbf{llama3.2-3b}:
    \textsc{Llama 3.2} instruction model with 3B parameters, a lightweight variant intended for low-latency or resource-constrained inference while retaining basic instruction-following capabilities.
    \newline \hspace*{1.5em}\textit{HuggingFace:} \href{https://huggingface.co/meta-llama/Llama-3.2-3B}{https://huggingface.co/meta-llama/Llama-3.2-3B}

    \item \textbf{llama3.3-70b}:
    \textsc{Llama 3.3} instruction model with 70B parameters, a later Llama release that maintains the same open-weight instruction interface, and is evaluated here as a high-capacity general-purpose model.
    \newline \hspace*{1.5em}\textit{HuggingFace:} \href{https://huggingface.co/meta-llama/Llama-3.3-70B-Instruct}{https://huggingface.co/meta-llama/Llama-3.3-70B-Instruct}

    \item \textbf{mistral-7b}:
    \textsc{Mistral} instruction model with 7B parameters, widely used as a compact general-purpose baseline that offers strong practical throughput under open-weight deployment.
    \newline \hspace*{1.5em}\textit{HuggingFace:} \href{https://huggingface.co/mistralai/Mistral-7B-Instruct-v0.2}{https://huggingface.co/mistralai/Mistral-7B-Instruct-v0.2}

    \item \textbf{ministral-8b}:
    \textsc{Ministral} instruction model with 8B parameters from the Mistral family, evaluated as a mid-sized open-weight model emphasizing practical instruction-following performance.
    \newline \hspace*{1.5em}\textit{HuggingFace:} \href{https://huggingface.co/mistralai/Ministral-3-8B-Instruct-2512}{https://huggingface.co/mistralai/Ministral-3-8B-Instruct-2512}

    \item \textbf{mistral-small3-24b}:
    \textsc{Mistral Small} instruction model with 24B parameters, offering a larger open-weight option than the 7B/8B variants and used when higher capacity is beneficial.
    \newline \hspace*{1.5em}\textit{HuggingFace:} \href{https://huggingface.co/mistralai/Mistral-Small-24B-Instruct-2501}{https://huggingface.co/mistralai/Mistral-Small-24B-Instruct-2501}

    \item \textbf{qwen2.5-3b}:
    \textsc{Qwen2.5} model with 3B parameters, a small multilingual checkpoint commonly used for lightweight inference and as a compact baseline within the Qwen family.
    \newline \hspace*{1.5em}\textit{HuggingFace:} \href{https://huggingface.co/Qwen/Qwen2.5-3B}{https://huggingface.co/Qwen/Qwen2.5-3B}

    \item \textbf{qwen2.5-7b}:
    \textsc{Qwen2.5} model with 7B parameters, a mid-sized multilingual model that supports instruction-style prompting and serves as a standard open-weight baseline.
    \newline \hspace*{1.5em}\textit{HuggingFace:} \href{https://huggingface.co/Qwen/Qwen2.5-7B}{https://huggingface.co/Qwen/Qwen2.5-7B}

    \item \textbf{qwen2.5-32b}:
    \textsc{Qwen2.5} model with 32B parameters, a higher-capacity multilingual checkpoint typically used for improved response quality and more complex generation tasks relative to smaller Qwen variants.
    \newline \hspace*{1.5em}\textit{HuggingFace:} \href{https://huggingface.co/Qwen/Qwen2.5-32B}{https://huggingface.co/Qwen/Qwen2.5-32B}

    \item \textbf{qwen3-4b}:
    \textsc{Qwen3} model with 4B parameters, evaluated as a newer-generation multilingual model in the Qwen series under standard instruction prompting.
    \newline \hspace*{1.5em}\textit{HuggingFace:} \href{https://huggingface.co/Qwen/Qwen3-4B}{https://huggingface.co/Qwen/Qwen3-4B}

    \item \textbf{qwen3-8b}:
    \textsc{Qwen3} model with 8B parameters, evaluated as a mid-sized Qwen3 checkpoint that balances capacity and efficiency for multilingual instruction-style generation.
    \newline \hspace*{1.5em}\textit{HuggingFace:} \href{https://huggingface.co/Qwen/Qwen3-8B}{https://huggingface.co/Qwen/Qwen3-8B}

    \item \textbf{qwen3-32b}:
    \textsc{Qwen3} model with 32B parameters, evaluated as a large Qwen3 checkpoint representing a higher-capacity multilingual baseline under the same prompting and decoding setup.
    \newline \hspace*{1.5em}\textit{HuggingFace:} \href{https://huggingface.co/Qwen/Qwen3-32B}{https://huggingface.co/Qwen/Qwen3-32B}

    \item \textbf{smollm3-3b}:
    \textsc{SmolLM3} model with 3B parameters, a lightweight open-weight model used to study low-resource performance and efficiency under the same evaluation protocol.
    \newline \hspace*{1.5em}\textit{HuggingFace:} \href{https://huggingface.co/HuggingFaceTB/SmolLM3-3B}{https://huggingface.co/HuggingFaceTB/SmolLM3-3B}

    \item \textbf{yi-1.5-9b}:
    \textsc{Yi-1.5} bilingual (Chinese--English) model with 9B parameters, used as an additional open-weight general-purpose baseline with strong Chinese/English coverage.
    \newline \hspace*{1.5em}\textit{HuggingFace:} \href{https://huggingface.co/01-ai/Yi-1.5-9B}{https://huggingface.co/01-ai/Yi-1.5-9B}

    \item \textbf{yi-1.5-34b}:
    \textsc{Yi-1.5} model with 34B parameters, a larger-capacity Yi checkpoint included to compare scaling behavior under the same evaluation pipeline.
    \newline \hspace*{1.5em}\textit{HuggingFace:} \href{https://huggingface.co/01-ai/Yi-1.5-34B}{https://huggingface.co/01-ai/Yi-1.5-34B}

    \item \textbf{doctor-r1-8b}:
    \textsc{Doctor-R1} medical model released as a domain-focused checkpoint intended for clinical reasoning and instruction-style medical generation, fine-tuned on the \textsc{Qwen3-8B} model.
    \newline \hspace*{1.5em}\textit{HuggingFace:} \href{https://huggingface.co/unicornftk/Doctor-R1}{https://huggingface.co/unicornftk/Doctor-R1}

    \item \textbf{med42-8b}:
    \textsc{Med42} clinical model (8B) fine-tuned on the \textsc{Llama3-8B} model for medical and biomedical language understanding and instruction following.
    \newline \hspace*{1.5em}\textit{HuggingFace:} \href{https://huggingface.co/m42-health/Llama3-Med42-8B}{https://huggingface.co/m42-health/Llama3-Med42-8B}
    
    \item \textbf{ultramedical-8b}:
    \textsc{UltraMedical} instruction-tuned medical model (8B) based on the \textsc{Llama3-8B} model, designed for medical QA-style prompting and clinical instruction following.
    \newline \hspace*{1.5em}\textit{HuggingFace:} \href{https://huggingface.co/TsinghuaC3I/Llama-3-8B-UltraMedical}{https://huggingface.co/TsinghuaC3I/Llama-3-8B-UltraMedical}

    \item \textbf{m1-7b-23k}:
    \textsc{m1} long-context medical model (7B; 23K variant) based on \textsc{Qwen2.5-7B}, included to study the impact of long-context capacity in clinical-style prompting.
    \newline \hspace*{1.5em}\textit{HuggingFace:} \href{https://huggingface.co/UCSC-VLAA/m1-7B-23K}{https://huggingface.co/UCSC-VLAA/m1-7B-23K}

    \item \textbf{m1-32b-1k}:
    \textsc{m1} long-context medical model (32B; 1K variant) based on \textsc{Qwen2.5-32B}, included to represent a higher-capacity medical checkpoint with an extended context interface.
    \newline \hspace*{1.5em}\textit{HuggingFace:} \href{https://huggingface.co/UCSC-VLAA/m1-32B-1K}{https://huggingface.co/UCSC-VLAA/m1-32B-1K}

    \item \textbf{huatuogpt-o1-8b}:
    \textsc{HuatuoGPT-o1} medical reasoning model (8B) fine-tuned on \textsc{Llama3-8B}, included as a medical-domain checkpoint with instruction-style interfaces and clinically oriented training.
    \newline \hspace*{1.5em}\textit{HuggingFace:} \href{https://huggingface.co/FreedomIntelligence/HuatuoGPT-o1-8B}{https://huggingface.co/FreedomIntelligence/HuatuoGPT-o1-8B}

    \item \textbf{gpt-oss-20b}:
    Open-weight GPT-OSS model with 20B parameters, included as an additional open-weight baseline with a GPT-style architecture and publicly released weights.
    \newline \hspace*{1.5em}\textit{HuggingFace:} \href{https://huggingface.co/openai/gpt-oss-20b}{https://huggingface.co/openai/gpt-oss-20b}

    \item \textbf{gpt-oss-120b}:
    Open-weight GPT-OSS model (120B) used in our evaluation setup as a large-capacity open-weight baseline.
    \newline \hspace*{1.5em}\textit{Azure OpenAI:} \href{https://learn.microsoft.com/en-us/azure/ai-services/openai/}{https://learn.microsoft.com/en-us/azure/ai-services/openai/}

    \item \textbf{gpt-4.1-nano}:
    Proprietary GPT-4.1 family model accessed via Azure OpenAI (deployment: \textit{gpt-4.1-nano}), included as a low-latency API model under the same prompting and evaluation protocol.

    \item \textbf{gpt-4.1-mini}:
    Proprietary GPT-4.1 family model accessed via Azure OpenAI (deployment: \textit{gpt-4.1-mini}), included as a cost-efficient API model for instruction-style generation in our evaluation setting.
    
    \item \textbf{gpt-4.1}:
    Proprietary GPT-4.1 family model accessed via Azure OpenAI (deployment: \textit{gpt-4.1}), included as a higher-capacity API model for strong general-purpose instruction following and generation.

    \item \textbf{gpt-5-nano}:
    Proprietary GPT-5 family model accessed via Azure OpenAI (deployment: \textit{gpt-5-nano}), included as a compact API model in the latest available GPT series within our subscription at evaluation time.

    \item \textbf{gpt-5-mini}:
    Proprietary GPT-5 family model accessed via Azure OpenAI (deployment: \textit{gpt-5-mini}), included as a mid-sized API model representing the same GPT series under a higher-capacity configuration than \textit{gpt-5-nano}.

    \item \textbf{gpt-5}:
    Proprietary GPT-5 family model accessed via Azure OpenAI (deployment: \textit{gpt-5-chat}), included as a large-sized API model representing the same GPT series under a higher-capacity configuration than \textit{gpt-5-nano}.

    \item \textbf{gpt-5.2}:
    Proprietary GPT-5 family model accessed via Azure OpenAI (deployment: \textit{gpt-5.2-chat}), included as a large-sized API model representing the same GPT series under a higher-capacity configuration than \textit{gpt-5-nano}.

\end{itemize}

\section{Limitations}
\label{sec:appendix_limitations}

While \modelname{} enables large-scale, reliable EHR-grounded evaluation of LLMs for clinical decision making, it has limitations. First, its construction uses only structured diagnoses, prescriptions, and procedures, excluding informative modalities (e.g., demographics, vital signs, laboratory tests, and imaging). Second, to make KB verification feasible and limit leakage between scenario context and the queried relation, each template uses a small, fixed context window. We focus on encounter-level settings as they are more reliable and grounded, whereas long-range cross-visit relations are weaker and harder to validate. Accordingly, our prognosis task is framed as next-encounter risk prediction rather than calibrated time-to-event forecasting, reflecting uncertain real-world visit timing. Third, although \modelname{} contains 960{,}067 questions, full-set benchmarking is omitted because inference cost and runtime would be prohibitive for many models, making comparisons impractical. We therefore evaluate a capped subset per data source and task for feasible, fair comparisons across diverse models. Finally, KB support trades recall for precision by favoring relations covered by resources (e.g., SemMedDB and UMLS-linked concepts), potentially under-representing rare, emerging, institution-specific, or context-dependent practices. Future work can extend \modelname{} by adding modalities, relaxing fixed-context assumptions with leakage-aware controls, supporting reliable multi-visit reasoning, and broadening verification coverage through KBs and evidence sources.

\section{Prompt Templates}
We summarize the prompts used for relation extraction (Table~\ref{tab:prompt_template_relation_extraction}), template completion (Table~\ref{tab:prompt_template_mcq_gen}), QA generation for MCQ (Table~\ref{tab:prompt_template_mcq_paraphrase}) and OEQ (Table~\ref{tab:prompt_template_reason_only}), and evaluation for MCQs and OEQs (Tables~\ref{tab:prompt_template_mcq_eval} and \ref{tab:prompt_template_open_eval}, respectively).

\begin{table*}[htbp]
\centering
\caption{Prompt template for task-grounded clinical relation extraction, used in Stage~1 of template generation.}
\label{tab:prompt_template_relation_extraction}
\resizebox{\linewidth}{!}{%
\begin{minipage}{0.99\linewidth}

\begin{tcolorbox}[
  colback=white,
  colframe=black!60,
  arc=1.2mm,
  boxrule=0.6pt,
  left=6pt,right=6pt,top=6pt,bottom=6pt
]
\footnotesize

\textbf{(A) Common Instruction Skeleton}

\vspace{2pt}
\begin{tabularx}{\linewidth}{@{}lX@{}}
\toprule
\textbf{Section} & \textbf{Template Content (polished \& condensed)} \\
\midrule

\textbf{ROLE} &
You are a clinical relation extractor and a \emph{strict JSON generator}. \\

\textbf{GOAL} &
Extract a bounded set of \textbf{clinically plausible relations} under the task-specific definitions below and output \textbf{exactly one} JSON object.
These relations will later be used to construct MCQ items:
\texttt{entity\_1} serves as evidence in the question context, and \texttt{entity\_2} serves as the correct option.
You do \textbf{not} generate questions here, but you must select relations that are usable for QA construction (for example, informative targets, feasible distractors, and minimal leakage). \\

\textbf{INPUT} &
You will receive structured EHR content listing clinical events for one or two visits.
The task-specific block defines:
(i) allowable \texttt{entity\_1/entity\_2} pools, (ii) visit scope (intra-visit versus cross-visit), and (iii) allowed relation labels. \\

\textbf{ABSOLUTE OUTPUT RULES} &
\begin{itemize}[leftmargin=*, itemsep=1pt, topsep=2pt]
\item Output \textbf{exactly one} valid JSON object enclosed in \textbf{one} \texttt{json} code block; output \textbf{nothing} else.
\item Top-level keys must include \texttt{"raw\_relations"} and \texttt{"context\_events"}; both must be lists.
\item Each relation item must contain \textbf{exactly} the keys:
\texttt{"entity\_1"}, \texttt{"relation"}, \texttt{"entity\_2"}, \texttt{"rationale"}.
\item Type constraints:
\texttt{entity\_1/entity\_2/relation} are non-empty strings; \texttt{relation} must be from the task-defined label set;
\texttt{rationale} is \textbf{exactly one English sentence} (no line breaks).
\item \texttt{context\_events} must contain a fixed number of \textbf{plain strings} (no objects) and must obey strict disjointness from all relation endpoints (see the procedure below).
\item Never output any medical codes; use human-readable event names only. No placeholders, no comments. Stop after the closing code fence.
\end{itemize} \\

\textbf{Context event string extraction (if present)} &
Use the following priority rules to form each \texttt{context\_events} string:
\begin{itemize}[leftmargin=*, itemsep=1pt, topsep=2pt]
\item Diagnosis: use \texttt{event["description"]} only.
\item Drug/prescription: use \texttt{event["drug"]} or \texttt{event["description"]}.
\item Procedure: use \texttt{event["description"]}.
\end{itemize}
Each string must be descriptive and human-readable (no timestamps and no codes). \\

\textbf{STEP-BY-STEP PROCEDURE} &
\begin{enumerate}[leftmargin=*, itemsep=1pt, topsep=2pt]
\item \textbf{Build candidate pools}: derive allowable \texttt{entity\_1} and \texttt{entity\_2} strictly from the task scope and the input events.
\item \textbf{Propose candidate relations}: form directed pairs \texttt{entity\_1 $\rightarrow$ entity\_2} with an allowed label.
Enforce: no self-loop; verbatim names only; remove duplicates (same endpoints and meaning).
Write a one-sentence rationale for each candidate.
\item \textbf{Select final \texttt{raw\_relations}}: prioritize clinical relevance, informativeness (avoid overly trivial or overly common endpoints),
and downstream MCQ usability (diversity, low leakage risk, and clean distractor feasibility). Ensure uniqueness and task-scope compliance.
\item \textbf{Select \texttt{context\_events}} from remaining events not used as any relation endpoint, subject to:
\begin{itemize}[leftmargin=*, itemsep=1pt, topsep=2pt]
\item \textbf{No-overlap rule}: no string overlap with any \texttt{entity\_1/entity\_2} (case-insensitive substring match).
\item \textbf{Stronger unrelatedness rule}: additionally exclude synonyms, abbreviations, spelling variants, and obvious parent/child concepts of any relation endpoint, and avoid same-topic restatements.
\item Do not reuse the same event object that instantiated any relation endpoint.
\end{itemize}
\item \textbf{Emit JSON only}: output exactly one \texttt{json} code block matching the schema.
\item \textbf{Silent self-check}: the schema is exact; the bounded relation count is satisfied; no codes; no duplicates; each rationale is one English sentence; context-event disjointness holds.
\end{enumerate} \\
\bottomrule
\end{tabularx}

\vspace{6pt}
\textbf{(B) Strict Output Schema (must match exactly)}

\vspace{2pt}
\begin{tcolorbox}[
  colback=black!2,
  colframe=black!40,
  arc=1.0mm,
  boxrule=0.4pt,
  left=6pt,right=6pt,top=4pt,bottom=4pt
]
\footnotesize
\texttt{json}\\
\texttt{\{}\\
\texttt{\ \ "raw\_relations":[}\\
\texttt{\ \ \ \ \{"entity\_1":"...", "relation":"...", "entity\_2":"...", "rationale":"..."\},}\\
\texttt{\ \ \ \ \ldots}\\
\texttt{\ \ ],}\\
\texttt{\ \ "context\_events":["...", "..."]}\\
\texttt{\}}\\
\end{tcolorbox}

\vspace{6pt}
\textbf{(C) Task-Specific Block (choose exactly one)}

\vspace{2pt}
\begin{tabularx}{\linewidth}{@{}lX@{}}
\toprule
\textbf{Task} & \textbf{Requirements} \\
\midrule

\textbf{Prognosis (Cross-visit)} &
Extract relations from \textbf{prior-visit} events to \textbf{next-visit diagnoses/outcomes}.
\texttt{entity\_1} must come from the prior visit; \texttt{entity\_2} must be a diagnosis/outcome present in the next visit.
\texttt{context\_events} must be chosen \textbf{only} from the prior visit.
Allowed labels: \texttt{cause}, \texttt{affect}, \texttt{associate\_with}.
Prefer relations reflecting progression/complications, treatment effects, or latent conditions that become explicit later. \\

\textbf{Diagnosis (Same-visit)} &
Extract diagnosis--diagnosis relations \textbf{within the same visit}.
Both \texttt{entity\_1} and \texttt{entity\_2} must be diagnoses from the current visit.
\texttt{context\_events} must be diagnoses from the current visit (no drugs/procedures).
Allowed labels: \texttt{cause}, \texttt{affect}, \texttt{associate\_with}.
Prefer complication links, shared mechanisms, or strong comorbidity patterns that support identification of an additional diagnosis in the same visit. \\

\textbf{Treatment (Same-visit)} &
Extract treatment-to-diagnosis relations \textbf{within the same visit}.
\texttt{entity\_1} must be a prescription or procedure event; \texttt{entity\_2} must be a diagnosis from the same visit.
\texttt{context\_events} must be diagnoses (no drugs/procedures).
Allowed labels: \texttt{drug\_treat}, \texttt{procedure\_treat}.
Prefer guideline-consistent therapies and procedures that directly manage or treat the diagnosis. \\
\bottomrule
\end{tabularx}

\end{tcolorbox}

\end{minipage}%
}
\end{table*}

\begin{table*}[t]
\centering
\caption{Prompt template for template completion, used in Stage~3 of template generation.}
\label{tab:prompt_template_mcq_gen}
\resizebox{\linewidth}{!}{%
\begin{minipage}{0.99\linewidth}

\begin{tcolorbox}[
  colback=white,
  colframe=black!60,
  arc=1.2mm,
  boxrule=0.6pt,
  left=6pt,right=6pt,top=6pt,bottom=6pt
]
\small
\textbf{DISTRACTOR\_NUMBER}: \texttt{10 (TEN)}

\vspace{4pt}
\textbf{(A) Common Instruction Skeleton}

\vspace{2pt}
\begin{tabularx}{\linewidth}{@{}lX@{}}
\toprule
\textbf{Section} & \textbf{Template Content (polished \& condensed)} \\
\midrule
\textbf{ROLE} &
You are a \emph{strict JSON generator} for clinical QA. You must produce \textbf{exactly one} valid JSON object and nothing else. \\

\textbf{GOAL} &
Generate \textbf{one clinically meaningful MCQ} for \textbf{one identified relation}, with \textbf{one correct answer} and distractors, grounded \textbf{only} in:
(i) the provided structured EHR context (task-defined scope), and
(ii) the provided relation object (with \texttt{relation\_id}, rationale, and KB verification evidence). \\

\textbf{INPUT} &
You will receive:
(1) \textbf{IDENTIFIED RELATION}: \texttt{relation\_id}, \texttt{entity\_1}, \texttt{entity\_2}, \texttt{relation}, \texttt{rationale}, \texttt{semmed\_verification}, and definitions for \texttt{entity\_1/entity\_2}. \newline
(2) \textbf{FORBIDDEN TERMS}: real events from the patient record; they \textbf{must not appear} in distractors (case-insensitive substring match). \\

\textbf{ABSOLUTE OUTPUT RULES} &
\begin{itemize}[leftmargin=*, itemsep=1pt, topsep=2pt]
\item Output \textbf{exactly one} JSON object enclosed in \textbf{one} \texttt{json} code block.
\item Output \textbf{nothing} outside the code block; \textbf{no} placeholders (for example, \texttt{<...>} or \texttt{\{\{...\}\}}) and \textbf{no} comments.
\item The top-level key must be \texttt{"base\_questions"} and its value must be a list.
\item Each question item must follow the schema \textbf{exactly} (keys and types).
\item Stop immediately after closing the code block.
\end{itemize} \\

\textbf{STEP-BY-STEP PROCESS} &
\begin{enumerate}[leftmargin=*, itemsep=1pt, topsep=2pt]
\item \textbf{Read inputs}: interpret \texttt{entity\_1} as the observed event and \texttt{entity\_2} as the target consequence/associated outcome per the task definition; use the rationale and KB evidence as grounding.
\item \textbf{Build the MCQ}: the stem must reflect the relation and the task scope (same-visit versus cross-visit).
\item \textbf{Set answer/topic}: \texttt{answer} must exactly equal the correct option string (task-defined; typically \texttt{entity\_2}, except for the treatment task); \texttt{topic} is a concise target label (default: \texttt{entity\_2}).
\item \textbf{Generate distractors}:
each distractor should be plausible in general but \textbf{not supported or favored} by the sample context and relation evidence, and it must not leak forbidden terms.
Prefer three types: \texttt{reversed} (wrong direction), \texttt{contradicted} (negated by evidence), and \texttt{unrelated} (not implied).
\item \textbf{Self-validate (mandatory)}:
valid JSON; the schema is exact; \#distractors = 10; no forbidden-term leakage; \textbf{non-nesting} (no synonyms, abbreviations, or parent-child granularity overlaps among choices).
\end{enumerate} \\
\bottomrule
\end{tabularx}

\vspace{6pt}
\textbf{(B) Strict Output Schema (must match exactly)}

\vspace{2pt}
\begin{tcolorbox}[
  colback=black!2,
  colframe=black!40,
  arc=1.0mm,
  boxrule=0.4pt,
  left=6pt,right=6pt,top=4pt,bottom=4pt
]
\footnotesize
\texttt{json}\\
\texttt{\{}\\
\texttt{\ \ "base\_questions":[}\\
\texttt{\ \ \ \ \{}\\
\texttt{\ \ \ \ \ \ "answer":"...",}\\
\texttt{\ \ \ \ \ \ "topic":"...",}\\
\texttt{\ \ \ \ \ \ "distractors":[\{"entity\_name":"...", "type":"reversed"\}, \ldots ]}\\
\texttt{\ \ \ \ \}}\\
\texttt{\ \ ]}\\
\texttt{\}}\\
\end{tcolorbox}

\vspace{6pt}
\textbf{(C) Task-Specific Block (choose exactly one)}

\vspace{2pt}
\begin{tabularx}{\linewidth}{@{}lX@{}}
\toprule
\textbf{Task} & \textbf{Requirements} \\
\midrule
\textbf{Prognosis (Cross-visit)} &
Predict a \textbf{next-visit diagnosis/outcome}. Relation: prior-visit \texttt{entity\_1} $\rightarrow$ next-visit \texttt{entity\_2}. \newline
Correct choice: \textbf{\texttt{entity\_2}}. \newline
All distractors: \textbf{diagnoses/outcomes only} (no drugs/procedures). \newline
Stem intent example: ``Given prior history, which diagnosis/outcome is most likely at the next visit?'' \\

\textbf{Diagnosis (Same-visit)} &
Infer an \textbf{additional diagnosis in the same visit}. Relation: same-visit \texttt{entity\_1} $\rightarrow$ same-visit \texttt{entity\_2}. \newline
Correct choice: \textbf{\texttt{entity\_2}}. \newline
All distractors: \textbf{diagnoses only}; must not appear in forbidden terms; enforce non-nesting aggressively. \\

\textbf{Treatment (Same-visit)} &
Select a \textbf{treatment/drug/procedure in the same visit}. Relation: treatment \texttt{entity\_1} $\rightarrow$ diagnosis \texttt{entity\_2}. \newline
Correct choice: \textbf{\texttt{entity\_1}}. \newline
All distractors: \textbf{treatments only}; no brand/generic duplicates, abbreviations, or formulation near-duplicates; no forbidden leakage. \newline
Topic default: typically \texttt{entity\_2} (indication/target diagnosis). \\
\bottomrule
\end{tabularx}

\end{tcolorbox}

\end{minipage}%
}
\end{table*}

\begin{table*}[t]
\centering
\caption{Prompt template for MCQ QA generation.}
\label{tab:prompt_template_mcq_paraphrase}
\resizebox{\linewidth}{!}{%
\begin{minipage}{0.99\linewidth}

\begin{tcolorbox}[
  colback=white,
  colframe=black!60,
  arc=1.2mm,
  boxrule=0.6pt,
  left=6pt,right=6pt,top=6pt,bottom=6pt
]
\small

\textbf{(A) Common Paraphrasing Instruction Skeleton}

\vspace{2pt}
\begin{tabularx}{\linewidth}{@{}lX@{}}
\toprule
\textbf{Section} & \textbf{Template Content (polished \& condensed)} \\
\midrule

\textbf{ROLE} &
You are a \emph{strict JSON generator} for \textbf{paraphrasing} clinical MCQ question stems. \\

\textbf{GOAL} &
Given an input clinical \texttt{context} and an ask-only \texttt{question}, generate a fixed set of paraphrased versions of the \textbf{same} question.
This is a \textbf{surface-level paraphrase} task:
\begin{itemize}[leftmargin=*, itemsep=1pt, topsep=2pt]
\item Do not invent new content or omit any event.
\item Do not change clinical intent.
\item Do not introduce emphasis, causality, or interpretation.
\end{itemize} \\

\textbf{INPUT} &
You will receive:
\begin{itemize}[leftmargin=*, itemsep=1pt, topsep=2pt]
\item \texttt{context}: one sentence (or two short sentences) that neutrally summarizes all clinical events.
\item \texttt{question} (ask-only): asks what may occur or what will happen (task-defined) and \textbf{must not mention any specific event}.
\end{itemize} \\

\textbf{ABSOLUTE OUTPUT RULES} &
\begin{itemize}[leftmargin=*, itemsep=1pt, topsep=2pt]
\item Output \textbf{exactly one} valid JSON object enclosed in \textbf{one} \texttt{json} code block; output \textbf{nothing} else.
\item The top-level key must be \texttt{"question\_versions"} and its value must be a list with a fixed length.
\item Each item must be a JSON object with \textbf{exactly} the required keys, and \texttt{version} must form a complete consecutive index set (each appears exactly once).
\item Stop immediately after closing the code fence.
\end{itemize} \\

\textbf{PARAPHRASE RULES (apply to every version)} &
\begin{itemize}[leftmargin=*, itemsep=1pt, topsep=2pt]
\item \textbf{Neutrality}: treat all events equally; no highlighting (for example, ``notably'' or ``especially'') and no causal language.
\item \textbf{Anti-abstraction (critical)}: explicitly name each concrete event from the input; do not replace events with high-level summaries
(for example, ``medical course'', ``complex presentation'', or ``respiratory challenges'').
\item \textbf{Linguistic constraints}: fluent clinical English; similar length across versions; prefer two sentences (at most three short sentences);
ensure meaningful syntactic and lexical variation while staying close to the original (high lexical overlap).
\item \textbf{No interpretation}: do not add qualifiers (for example, ``severe'', ``suggesting'', or ``consistent with'') and do not create mechanisms or causal chains.
\end{itemize} \\

\textbf{STEP-BY-STEP (mandatory, silent)} &
\begin{enumerate}[leftmargin=*, itemsep=1pt, topsep=2pt]
\item Read input \texttt{context} and \texttt{question}.
\item Produce the required number of versions by paraphrasing the \texttt{context} while preserving neutrality and full event coverage, and keep the \texttt{question} ask-only (no event leakage).
\item Self-check silently: the count is correct; the intent is unchanged; the schema is valid; the output contains a single JSON code block only.
\end{enumerate} \\
\bottomrule
\end{tabularx}

\vspace{6pt}
\textbf{(B) Strict Output Schema (must match exactly)}

\vspace{2pt}
\begin{tcolorbox}[
  colback=black!2,
  colframe=black!40,
  arc=1.0mm,
  boxrule=0.4pt,
  left=6pt,right=6pt,top=4pt,bottom=4pt
]
\footnotesize
\texttt{json}\\
\texttt{\{}\\
\texttt{\ \ "question\_versions":[}\\
\texttt{\ \ \ \ \{"version":1, "context":"...", "question":"..."\},}\\
\texttt{\ \ \ \ \ldots}\\
\texttt{\ \ ]}\\
\texttt{\}}\\
\end{tcolorbox}

\vspace{6pt}
\textbf{(C) Task-Specific Block (choose exactly one)}

\vspace{2pt}
\begin{tabularx}{\linewidth}{@{}lX@{}}
\toprule
\textbf{Task} & \textbf{Context framing constraint (paraphrasable; logic fixed)} \\
\midrule
\textbf{Prognosis (Prior $\rightarrow$ Next)} &
The \texttt{context} must be framed as a \textbf{prior-visit} summary and must neutrally cover all events from \texttt{sample\_context} with equal emphasis.
Recommended logic: ``At the prior visit, the patient's history included \texttt{<all events>}.''
Do not mention any next-visit outcome in the question. \\

\textbf{Diagnosis (Same-visit)} &
The \texttt{context} must be framed as a \textbf{current-visit} summary and must neutrally cover all events from \texttt{sample\_context}.
Recommended logic: ``At the current visit, the patient's diagnoses included \texttt{<all events>}.''
The question remains ask-only and event-free. \\

\textbf{Treatment (Same-visit)} &
The \texttt{context} must be framed as a \textbf{current-visit} summary, neutrally covering all events from \texttt{sample\_context} with equal emphasis.
Recommended logic: ``At the current visit, the patient's diagnoses included \texttt{<all events>}.''
Do not introduce treatment rationale or prioritization. \\
\bottomrule
\end{tabularx}

\end{tcolorbox}

\end{minipage}%
}
\end{table*}

\begin{table*}[t]
\centering
\caption{Prompt template for OEQ generation.}
\label{tab:prompt_template_reason_only}
\resizebox{\linewidth}{!}{%
\begin{minipage}{0.99\linewidth}

\begin{tcolorbox}[
  colback=white,
  colframe=black!60,
  arc=1.2mm,
  boxrule=0.6pt,
  left=6pt,right=6pt,top=6pt,bottom=6pt
]
\small

\textbf{(A) Common Reason-Only Instruction Skeleton}

\vspace{2pt}
\begin{tabularx}{\linewidth}{@{}lX@{}}
\toprule
\textbf{Section} & \textbf{Template Content (polished \& condensed)} \\
\midrule

\textbf{ROLE} &
You are a clinical summarizer. Produce a concise reason grounded \textbf{only} in the provided verified relation fields. \\

\textbf{INPUT} &
You will receive one JSON object containing:
\texttt{entity\_1}, \texttt{entity\_2}, \texttt{relation}, \texttt{rationale},
\texttt{entity\_1\_definition}, \texttt{entity\_2\_definition}. \\

\textbf{FORBIDDEN} &
\begin{itemize}[leftmargin=*, itemsep=1pt, topsep=2pt]
\item Do not mention any knowledge source or database name (for example, ``SemMed'', ``evidence'', citations, or similar).
\item Do not introduce new medical facts beyond the provided fields.
\item Do not explicitly reference that you were given ``definitions'' or ``rationale''.
\end{itemize} \\

\textbf{STYLE} &
\begin{itemize}[leftmargin=*, itemsep=1pt, topsep=2pt]
\item Neutral, concise clinical English.
\item Length: \textbf{2--4 sentences}.
\end{itemize} \\

\textbf{OUTPUT (STRICT)} &
Output \textbf{exactly one} JSON object enclosed in \textbf{one} \texttt{json} code block; output \textbf{nothing} else.
The JSON schema must be:
\texttt{\{"reason":"..."\}}. \\
\bottomrule
\end{tabularx}

\vspace{6pt}
\textbf{(B) Strict Output Schema (must match exactly)}

\vspace{2pt}
\begin{tcolorbox}[
  colback=black!2,
  colframe=black!40,
  arc=1.0mm,
  boxrule=0.4pt,
  left=6pt,right=6pt,top=4pt,bottom=4pt
]
\footnotesize
\texttt{json}\\
\texttt{\{"reason":"..."\}}\\
\end{tcolorbox}

\vspace{6pt}
\textbf{(C) Task-Specific Block (choose exactly one)}

\vspace{2pt}
\begin{tabularx}{\linewidth}{@{}lX@{}}
\toprule
\textbf{Task} & \textbf{Guidance (time framing fixed; do not reinterpret)} \\
\midrule

\textbf{Prognosis (Prior $\rightarrow$ Next)} &
Explain why \texttt{entity\_1} supports \texttt{entity\_2} in a cross-visit setting:
prior-visit \texttt{entity\_1} precedes and is linked to next-visit \texttt{entity\_2}.
Use only the provided relation fields and keep the temporal framing unchanged. \\

\textbf{Diagnosis (Same-visit)} &
Explain why \texttt{entity\_1} supports \texttt{entity\_2} within a single visit:
same-visit \texttt{entity\_1} is linked to same-visit \texttt{entity\_2}.
Use only the provided relation fields and keep the visit framing unchanged. \\

\textbf{Prescription (Same-visit)} &
Explain why \texttt{entity\_2} supports a treatment decision involving \texttt{entity\_1}:
diagnoses (\texttt{entity\_2}) motivate selection of a treatment (\texttt{entity\_1}) in the same visit.
Use only the provided relation fields and keep it neutral and grounded. \\
\bottomrule
\end{tabularx}

\end{tcolorbox}

\end{minipage}%
}
\end{table*}

\begin{table*}[t]
\centering
\caption{Prompt template for evaluating MCQs.}
\label{tab:prompt_template_mcq_eval}
\resizebox{\linewidth}{!}{%
\begin{minipage}{0.99\linewidth}

\begin{tcolorbox}[
  colback=white,
  colframe=black!60,
  arc=1.2mm,
  boxrule=0.6pt,
  left=6pt,right=6pt,top=6pt,bottom=6pt
]
\small

\textbf{(A) Common Evaluation Instruction Skeleton}

\vspace{2pt}
\begin{tabularx}{\linewidth}{@{}lX@{}}
\toprule
\textbf{Section} & \textbf{Template Content (polished \& condensed)} \\
\midrule

\textbf{INSTRUCTION} &
You will be given multiple independent multiple-choice medical questions.
For \textbf{each} question, select the \textbf{single best answer} and return the result under the strict JSON output rules below.
Do \textbf{not} provide explanations, reasoning, or any text outside the JSON output. \\

\textbf{OUTPUT REQUIREMENTS (STRICT)} &
\begin{itemize}[leftmargin=*, itemsep=1pt, topsep=2pt]
\item Output \textbf{exactly one} JSON object enclosed in \textbf{one} \texttt{json} code block; output \textbf{nothing} else.
\item The JSON object must contain the top-level key \texttt{"answers"}.
\item \texttt{"answers"} must be a list of \textbf{single capital letters}.
\item The $i$-th letter corresponds to the $i$-th question.
\item Each letter must be one of the options shown for that question.
\end{itemize} \\

\textbf{SELF-CHECK (silent)} &
Before outputting, silently verify:
valid JSON; exactly one code block; English-only output; the answer count equals the question count;
each answer is a single valid capital-letter option. \\
\bottomrule
\end{tabularx}

\vspace{6pt}
\textbf{(B) Required Output Schema (must match exactly)}

\vspace{2pt}
\begin{tcolorbox}[
  colback=black!2,
  colframe=black!40,
  arc=1.0mm,
  boxrule=0.4pt,
  left=6pt,right=6pt,top=4pt,bottom=4pt
]
\footnotesize
\texttt{json}\\
\texttt{\{}\\
\texttt{\ \ "answers":[}\\
\texttt{\ \ \ \ "A", "C", "B", \ldots}\\
\texttt{\ \ ]}\\
\texttt{\}}\\
\end{tcolorbox}

\end{tcolorbox}

\end{minipage}%
}
\end{table*}

\begin{table*}[t]
\centering
\caption{Prompt template for evaluating OEQs.}
\label{tab:prompt_template_open_eval}
\resizebox{\linewidth}{!}{%
\begin{minipage}{0.99\linewidth}

\begin{tcolorbox}[
  colback=white,
  colframe=black!60,
  arc=1.2mm,
  boxrule=0.6pt,
  left=6pt,right=6pt,top=6pt,bottom=6pt
]
\small

\textbf{(A) Common Evaluation Instruction Skeleton}

\vspace{2pt}
\begin{tabularx}{\linewidth}{@{}lX@{}}
\toprule
\textbf{Section} & \textbf{Template Content (polished \& condensed)} \\
\midrule

\textbf{INSTRUCTION} &
You will be given multiple independent clinical open-ended questions.
For each question, produce two outputs:
\texttt{answer} (a short event phrase) and \texttt{reason} (a concise explanation). \\

\textbf{CRITICAL RULES} &
\begin{itemize}[leftmargin=*, itemsep=1pt, topsep=2pt]
\item Do not invent new facts; use only what the question implies.
\item \texttt{answer} must be a \textbf{short phrase} (not a sentence; no extra words).
\item \texttt{reason} must be \textbf{1--3 sentences}.
\item Do not include literal labels such as ``Answer:'' or ``Reason:'' inside the strings.
\item Do not add any extra keys beyond the required schema.
\end{itemize} \\

\textbf{OUTPUT REQUIREMENTS (STRICT)} &
\begin{itemize}[leftmargin=*, itemsep=1pt, topsep=2pt]
\item Output \textbf{exactly one} valid JSON object enclosed in \textbf{one} \texttt{json} code block; output \textbf{nothing} else.
\item The JSON object must contain \textbf{exactly} two top-level keys:
\texttt{"answers"} (list of strings) and \texttt{"reasons"} (list of strings).
\item The $i$-th answer and reason correspond to question $i$.
\item The two lists must have the same length and equal the number of questions.
\item Stop immediately after closing the code fence.
\end{itemize} \\

\textbf{SELF-CHECK (silent)} &
Before outputting, silently verify:
one code block; valid JSON; only keys \texttt{"answers"} and \texttt{"reasons"};
lengths match question count; each answer is a short phrase without sentence-ending punctuation;
each reason is 1--3 sentences; no text outside the JSON block. \\
\bottomrule
\end{tabularx}

\vspace{6pt}
\textbf{(B) Required Output Schema (must match exactly)}

\vspace{2pt}
\begin{tcolorbox}[
  colback=black!2,
  colframe=black!40,
  arc=1.0mm,
  boxrule=0.4pt,
  left=6pt,right=6pt,top=4pt,bottom=4pt
]
\footnotesize
\texttt{json}\\
\texttt{\{}\\
\texttt{\ \ "answers":["...","...","..."],}\\
\texttt{\ \ "reasons":["...","...","..."]}\\
\texttt{\}}\\
\end{tcolorbox}

\end{tcolorbox}

\end{minipage}%
}
\end{table*}

\end{document}